\documentclass[11pt]{article}

\usepackage[final]{acl}

\usepackage{times}
\usepackage{latexsym}

\usepackage[T1]{fontenc}
\usepackage[utf8]{inputenc}

\usepackage{microtype}

\usepackage{inconsolata}

\usepackage{graphicx}
\usepackage{adjustbox}
\usepackage{amsmath}
\usepackage[boxed,ruled,linesnumbered]{algorithm2e}
\usepackage{booktabs} 
\usepackage{array}
\newcolumntype{L}[1]{>{\raggedright\arraybackslash}p{#1}}
\newcolumntype{C}[1]{>{\centering\arraybackslash}p{#1}}
\usepackage{longtable}
\usepackage{pdflscape} % {lscape}
\usepackage{threeparttable}

\usepackage{colortbl} % colors for the table
\usepackage{multirow} % for the table
\usepackage{cleveref}
\usepackage{ulem}

\newcommand{\Comment}[1]{}

\newcommand{\dan}[1]{}
\newcommand{\zk}[1]{}
\newcommand{\ca}[1]{}

\title{Linguistic and Argument Diversity in Synthetic Data for Function-Calling Agents}
\author{Dan Greenstein\thanks{\raggedright Work done while the author was an intern at MorganaLabs.}\\
  Technion, Haifa, Israel\\
  {\small\texttt{sdngreen@campus.technion.ac.il}} \And
  Chen Amiraz\\
  MorganaLabs, Haifa, Israel\\
  {\small\texttt{chen@morganalabs.ai}}\\ \AND
  Zohar Karnin\\
  MorganaLabs, Haifa, Israel\\
  {\small\texttt{zkarnin@morganalabs.ai}}\\ \And
  Oren Somekh\\
  MorganaLabs, Haifa, Israel\\
  {\small\texttt{oren.somekh@morganalabs.ai}}}

\begin{document}

\maketitle

\begin{abstract}
The construction of function calling agents has emerged as a promising avenue for extending model capabilities. A major challenge for this task is obtaining high quality diverse data for training. Prior work emphasizes diversity in functions, invocation patterns, and interaction turns, yet linguistic diversity of requests and coverage of arguments (e.g., \texttt{city\_name}, \texttt{stock\_ticker}) remain underexplored. We propose a method that generates synthetic datasets via optimizing general-purpose diversity metrics across both queries and arguments, without relying on hand-crafted rules or taxonomies, making it robust to different usecases. We demonstrate the effectiveness of our technique  via both intrinsic and extrinsic testing, comparing it to SoTA data generation methods. 

First, we show that while maintaining the same level of correctness (i.e., examples are realistic) our method significantly improves the resulting benchmark diversity. Next, when used as a training set, the model resulting from fine-tuning on our dataset exhibits superior performance compared to analogous models fine-tuned on baseline data generation methods. In particular, our fine-tuned model achieve an $7.4\%$ increase in accuracy on the BFCL benchmark compared to the baseline counterparts.
\end{abstract}

\section{Introduction}
Function-calling agents play a central role in modern LLM-based systems enabling interaction with external APIs, tools, and services through structured invocations \citep{schick2023toolformer,zhou2023webarena,qu2025tool}. A major challenge in this area is the limited availability of high-quality training and evaluation data. Prior work has addressed this gap by introducing manually curated benchmarks \citep{patilberkeley, wang2024toolflow} as well as automated methods for generating synthetic training data \citep{liu2024apigen, liu2409toolace}.

As in other areas of machine learning, diversity in training data is critical for robust generalization \citep{wang2024survey}. Namely, greater dataset diversity improves model generalization to unseen inputs \cite{yu2022can, gao2020pile, liu2023can, jung2025prismatic, ba2024does}.

Prior work on datasets for function-calling agents had recognized the importance of diversity, but to the best of our knowledge, had primarily focused on a limited set of dimensions: (1) the invoked functions, emphasizing both broad coverage and diversity across domains; (2) the invocation structure (e.g., single-function versus multi-function calls); and (3) single-turn versus multi-turn interactions. Our work is motivated by the observation that two additional high-level dimensions of diversity remain underexplored.

The first concerns the shape of the request, namely the linguistic form of the user utterance. While linguistic diversity has been actively studied and encouraged in other domains \citep{wang2024survey, filice2025generating}, we are not aware of prior work that explicitly measures or directly optimizes for such diversity in the context of function-calling datasets. The second dimension is argument diversity, i.e., the diversity of parameter values. In many cases, arguments admit a wide range of valid values, yet generated datasets tend to concentrate on a small set of frequent, head values.
As an illustrative example, consider a stock-market-related function library with a \texttt{stock\_ticker} argument. Without a dedicated diversification procedure, 
the data is likely to contain only
popular tickers such as \texttt{AAPL}. Under such conditions, model performance on less common tickers remains unclear. In fact, in this particular example, we conjecture that an LLM may successfully map free-text mentions to popular tickers, while failing on less common ones without access to a dedicated ticker lookup or search tool.

To further motivate these diversity dimensions, consider an industrial evaluation setting in which one aims to assess a task-specific agent, e.g., a calendar assistant or travel agent. Such agents are typically limited to a small, fixed set of functions; therefore, constructing a meaningful test set that captures a broad range of realistic use cases requires diversity beyond simply increasing the number of available functions.

In this work we introduce a technique that promotes diversity across both language and arguments. Our core contribution is a novel component that generates a diverse collection of items. This component does not rely on hand-written rules or predefined taxonomies of diversity, such as explicitly enumerating user personas \cite{ge2024scaling} or manually specifying linguistic properties to encourage variation \cite{filice2025generating}. Instead, it automatically generates items by optimizing simple, general-purpose diversity metrics. A key advantage of this approach is its robustness to new domains and settings, as by definition, it adapts to them automatically. In contrast, a fixed persona pool, a manually defined criteria or prompts inducing diversity cannot adapt easily. Any such general purpose collection should be manually adapted to specific topics.
We apply this component to generate diverse collections of argument values (e.g., \texttt{city\_name}, \texttt{stock\_ticker}) as well as diverse user queries, focusing on variation in query shape, including semantic, lexical, and syntactic properties.

To evaluate the quality of our techniques we compare it against (i) SoTA recently published dataset generation methods for function calling (ToolAce~\cite{liu2409toolace} and APIGen~\cite{liu2024apigen}), (ii) simple post-hoc LLM augmentations aimed at increasing linguistic or argument diversity---LLM paraphrase and entity swapping, defined in \S\ref{sec:Analysis}---and (iii) those augmentations applied to each release. %\os{this was is a bit unclear. What do you mean by that?} \zk{rephrased, please check if better}. \ca{I strongly dislike the word "strawmen". suggestion: na\"ive augmentation methods aimed to increase either linguistic or parameter diversity} Our evaluation is both intrinsic and extrinsic. 

For the intrinsic evaluation, we used both our method and baseline techniques to generate datasets from a shared pool of functions. While achieving comparable levels of data correctness (i.e., the fraction of realistic and valid examples), our generated dataset achieves higher diversity on the large majority of established diversity metrics (metrics we did not optimize for), including against augmented variants of the baselines. Interestingly, these findings extend beyond synthetic baselines; when compared to manually curated datasets, we observe similar trends. In particular, the BFCL benchmark~\cite{patilberkeley} exhibits markedly lower diversity than our collection.

For the extrinsic evaluation we use our generated collection to fine-tune a language model for tool invocation and compare it against models fine-tuned on equally sized datasets produced by baseline data-generation methods, namely ToolAce and APIGen. We conducted two experiments demonstrating that the substantially increased linguistic and argument diversity introduced by our approach improves out-of-distribution (OOD) performance. In the first experiment, we evaluated compared each pair of models on the third dataset (e.g.\ our model and the model trained on the ToolAce dataset were both evaluated on the APIGen dataset). Our model consistently achieved superior OOD performance, with an improvement of approximately $4\%$ in accuracy. In the second experiment, we evaluated all three models on the BFCL benchmark, and observed the same trend. Specifically, our model outperformed its counterparts by approximately $7\%$ accuracy points.

Summarizing, our contributions are (1) a novel and flexible technique for generating diverse data instances; (2) a tailored application of this technique for generating synthetic function-calling data; (3) a thorough analysis of the proposed method, highlighting the value of increasing linguistic and argument diversity in function-calling datasets.

\section{Related Work}
\label{sec:Related Work}
Function calling has become a central capability for large language models, enabling interaction with external APIs, tools, and services through structured invocations \citep{schick2023toolformer,zhou2023webarena,qu2025tool}. 
The standard setting is to determine, given a user request, which function should be called and with what arguments (see \citet{schick2023toolformer, patil2024gorilla} and the survey of \citet{qu2025tool}). A natural extension is single-turn interaction, where a request may result in one, multiple, or no function calls. This setting was standardized by \citet{patil2024gorilla} through APIBench, a benchmark grounded in real API documentation for evaluating function invocation accuracy. Subsequently, the Berkeley Function-Calling Leaderboard (BFCL) emerged as a widely adopted benchmark for single-turn function calling under controlled conditions \citep{patilberkeley}, evaluating single-function, multi-function, and no-function-call scenarios.
Beyond single-turn interactions, several works study multi-turn tool-use settings, including data generation \citep{wang2024toolflow, yin2025magnet}, training \citep{yin2025magnet, polyakov2025toolreflection}, and evaluation \citep{yao2024tau, barres2025tau}. Our focus is on the single-turn setting, where we show that increasing diversity remains an important challenge even without the additional complexity of multi-turn interactions.

\vspace{-0.1in}
As the supervised data requirements of large language models continue to grow, manual data collection and curation become increasingly expensive and difficult to scale. Consequently, synthetic data generation has become a widely adopted approach, using language models to generate supervised examples conditioned on task descriptions or formal specifications \citep{wang2024survey, tan2024large, liu2024best}.  

Within tool-use research, prior work has explored instruction tuning and model training \citep{qin2023toolllm, tang2023toolalpaca, wang2023mint, wang2024mtu, zeng2025toolace}, as well as structured and multi-step tool use \citep{yin2025magnet, zhou2025toolgrad, polyakov2025toolreflection}. A common pattern is to define a collection of APIs or functions from multiple domains and automatically generate natural-language queries paired with structured tool invocations, as in \citet{patil2024gorilla}. Subsequent work studied challenges such as API normalization and cleaning, synthetic API generation \citep{liu2409toolace, qu2025exploration}, and the construction of multi-tool interactions through structured sampling of related functions \citep{yin2025magnet}. Systems such as ToolFlow additionally emphasize sequential and parallel tool usage across turns \citep{wang2024toolflow}.

A similar line of work focuses specifically on constructing large-scale function-calling datasets over realistic APIs. APIGen \citep{liu2024apigen} generates natural-language queries and corresponding API calls from executable API documentation, primarily targeting single and parallel function calls. In contrast, ToolACE \citep{liu2409toolace} constructs synthetic API specifications and generates a broader range of interaction patterns, including non-invocation cases and sequential function calls. These works are the closest to our paper, as their main contribution is a synthetic dataset generation method over a realistic set of APIs. The key difference is our emphasis on linguistic and argument diversity. 

See Appendix~\ref{app: Measuring Diversity - Related work} for a survey of diversity metrics used in related work. 
%Complementary to such \emph{de novo} generation approaches, one can instead post-process existing datasets (e.g., through paraphrasing or entity substitution) to increase diversity; in our experiments we compare against both styles of baselines.
% \ca{I suggest to emphasize more that these are the most relevant and closely related works, that they also generate synthetic benchmarks, but without specifically addressing diversity, and that we use them as baselines} \zk{added a sentence, please review.}

\section{Our Methods}
\label{sec:Our Methods}
\begin{figure}[t]
    \centering
    \includegraphics[width=\linewidth]{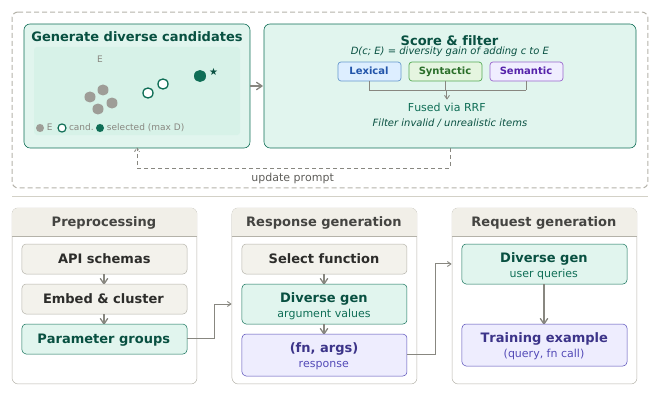}
    \caption{Overview of synthetic function-calling example generation for a single call. Top: diverse generation; bottom: data-generation pipeline.}
    \label{fig:method pipeline}
\end{figure}

Next, we detail each component of our synthetic data-generation method (\Cref{fig:method pipeline}).

\subsection{Diverse Generation} \label{sec:diverse gen alg}
Our core component is a greedy procedure for diverse generation of data instances (or items). 
As a running example, consider the task of generating requests for queries related to stock ticker values.

The procedure takes four inputs: (1) instructions $\cal I$ specifying what constitutes a valid item; (2) a set of existing examples $E$; (3) a diversity contribution metric $D$; and (4) a filter $F$ that removes invalid items.
For example, $\cal I$ may be ``\texttt{Generate a query about the price of a stock ticker}'', $E$ may contain ``\texttt{How much is Apple stock worth?}'' and ``\texttt{Did Amazon's price change in the last month?}'', $D$ may be a unique n-gram count or the Vendi score \cite{friedman2022vendi}, and $F$ may reject invalid items, e.g., off-topic or mismatched queries.
The procedure operates in rounds. Each round produces several candidates, e.g., ``\texttt{How much is Intel stock today?}''. At the end of a round we keep the top $b$ candidates according to $D$ among those that pass $F$, add them to $E$, and repeat until "enough" items are collected.
 
In round $1$ we ask the LLM to generate candidates different from those in $E$. If $E$ is too "large", we pass a subsample. We build a prompt from $\cal I$ and $E$, e.g., ``\texttt{Generate a query about the price of a stock ticker. Make sure it is structured differently than $\ldots$}''. Formally, $C_1 = \text{Gen}_{\text{cand}}({\cal I}, E)$ is the candidate set of round 1.
For rounds $r>1$, we score each item in $C_{r-1}$ with $F$ and $D$, then query an LLM for updated instructions from the scored candidates. Formally, ${\cal I}_r = \text{Gen}_{\text{inst}}({\cal I}, C_{r-1}, D(C_{r-1}), F(C_{r-1}))$, where we abuse notation to let $D(C) = \{D(c) \mid c \in C\}$, and likewise for $F$. The new instructions steer generation away from items that $F$ rejects and toward higher $D$. We then set $C_r = \text{Gen}_{\text{cand}}({\cal I}_r, E)$. The details of this procedure are provided in Appendix \ref{sec:detailed_method}, and the prompts are given in Appendix \ref{sec:prompts_appendix}.

Note that $D$ assigns a score to an item. Diversity metrics are usually defined on sets; given a set-level metric $\tilde{D}$, the score for candidate $c$ given existing items $E$ is $D(c; E) = \tilde{D}(E \cup \{c\}) - \tilde{D}(E)$, i.e., the marginal contribution of $c$. When $E$ is clear from context we write $D(c)$.
This process is the only stage that requires non-constant time per example; computing $\tilde{D}$ for the $n^{th}$ data instance takes time at least linear in $n$. Our experiments use corpora of roughly 2K examples, where that cost remained comparable to an LLM call. For much larger datasets, subsampling $E$ when evaluating $\tilde{D}$ would limit the effective cost per new item. %\ca{I suggest we make this a footnote, it interrupts the flow and takes up space. Another option is to defer this discussion to the appendix.}\dan{Not sure to which degree we can - this is a specific ask by the AC.} \zk{TODO: Move the runtime comment to the end of this subsection - this way flow doesn't break.}

\subsection{Generating Function Calling Interactions} \label{sec: method single func}
We review here the process of generating interactions resulting in a single function call. In Section~\ref{sec: method extensions} we discuss extensions, detailing the full process.  

Our process has three main parts (\Cref{fig:method pipeline}). (1) {\em Preprocessing} clusters semantically similar parameters across functions, such as city names and file paths, to support later diversity; (2) {\em response generation} selects the invoked function and its argument values; and (3) {\em request generation} produces the user query that matches that call. We generate the response before the request because this order yields better agreement between the call and the query.

\noindent {\bf Preprocessing} We apply an embedding model\footnote{We used the all-MiniLM-L6-v2 model available in Hugging Face.} to embed each parameter, e.g., \texttt{stock\_query/ticker}, \texttt{stock\_query/datetime}, or \texttt{stock\_purchase/ticker}. 
Formally, for each parameter $p$ we compute an embedding $E(p)$ using its name and description. We use cosine similarity to identify similar parameters, and cluster together all parameters with $\text{cos}(E(p_1), E(p_2)) \geq 0.6$.
A detailed discussion of the cutoff choice and the distribution of similarity scores is in Appendix \ref{sec:param_embed_threshold_calib}.

\noindent {\bf Response generation} We first choose the function for the response. For a single function call, we choose a function at random\footnote{To ensure sufficient representation of different parameter types, the sampling process was not strictly uniform. Additional details are provided in Appendices~\ref{sec:api_pools} and \ref{sec:api_sampling}.}. We then apply the diverse generation method of \S\ref{sec:diverse gen alg} to choose values for its arguments, e.g., \texttt{stock\_query/ticker}.

For each argument, we build instructions ${\cal I}$, such as ``\texttt{Provide a stock ticker to be queried}'' that specify the function, parameter and any values already chosen for other arguments. This context helps avoid incompatible combinations, e.g., an end time earlier than a start time. We also supply prior examples $E$, i.e., values previously chosen for the same semantic parameter type, identified during preprocessing, are passed into diverse generation.

To save resources, we use only a single diversity round and run the filter once after all arguments are generated, rather than after each argument individually\footnote{In case of a failure we retried up to three times.}. The diversity metric and filter are detailed in Appendix \ref{sec:parameter_val_generation}.

\noindent {\bf Request generation} 
Given the required response, we apply the diverse generation method of \S\ref{sec:diverse gen alg} to generate a matching request, using the full multi-round procedure rather than the single round used for response generation.
The instructions $\cal I$ include the function definition and sought response, e.g., ``\texttt{Generate a request looking for the value of AAPL in Apr 30, 2022}''. The previous examples $E$ are previously generated requests.

We evaluate each candidate with several diversity metrics capturing lexical, semantic, and syntactic diversity. For each metric, we compute the marginal contribution $D(c; E)$ from \S\ref{sec:diverse gen alg}, i.e., the increase in that metric when candidate $c$ is added to $E$. Because the metrics use incomparable scales, we rank candidates separately under each one rather than merging raw scores. We then fuse these per-metric rankings with Reciprocal Rank Fusion (RRF)\footnote{We use $k=60$, a standard default for the RRF parameter.} and use the fused rank as the diversity score. Filtering is done via a tailored prompt aimed to avoid both unrealistic requests and requests not matching the intended response. We provide the full details for the process in Appendix \ref{sec:query_generation}.
\S\ref{sec:Analysis} ablates this multi-round setting against a single-round request-generation baseline.

\subsection{Beyond single function calls} \label{sec: method extensions}
We now move to explain how to (1) extend the process to zero or multiple function calls; and (2) assemble a candidate API list for each example, including \emph{challenging distractors}--related invocation candidates that should not be selected (see Appendix \ref{sec:distractors_retrieval}).

\noindent {\bf Multiple functions}. We follow ToolFlow \cite{wang2024toolflow} to select the APIs invoked in each multi-call example. In preprocessing we build an API similarity graph: \emph{parallel} edges connect APIs with similar inputs, e.g., \texttt{get\_weather} and \texttt{search\_hotel}; \emph{sequential} edges connect APIs whose outputs can feed the other API's inputs. We sample the target APIs via a biased random walk on this graph (for further details, see Appendix \ref{sec:function_graph}). For sequential examples, the walk order determines execution order, since later APIs may consume outputs of earlier calls; for parallel examples, no API requires another's output to execute, so the calls may run in any order.

For parameter values, we run the diverse generation procedure above on \emph{one} API per example, i.e., the terminal API in a sequential chain, or one API chosen at random in a parallel set.
All remaining parameters are filled by a single LLM call conditioned on the schemas and values already chosen, so shared fields such as dates, locations, and identifiers, stay mutually consistent.
We also tried applying diverse generation to every API in multi-call examples; although diversity metrics increased, implausible joint settings became too frequent, so we restrict diversity optimization to one API per example.

\noindent {\bf Zero function calls.} 
We also generate examples with no function call. Completely vague requests, such as \textit{How are you?} or \textit{What can you do?}, are unrelated to any API; we only adjust the instructions and do not select a target function. For requests with incomplete information, such as \textit{Book a flight leaving Paris tomorrow}, we still build the full ground-truth function call but deliberately omit one or more required arguments from the user query, i.e., we choose a function with at least one required parameter, generate the call as in the single-function case, randomly withhold selected required parameters from the request, and record the omission in the instructions. We detail the full procedure in Appendix \ref{sec:query_generation}.

\noindent {\bf Choosing distractors.}
Each training example comprises the user request, the ground-truth function call(s), and a candidate list of APIs the model may invoke (including the target API(s)).
A \emph{distractor} is an API in that list that should not be invoked for the request.
Prior work often samples distractors uniformly from the API pool or from the same domain as the target; we instead aim to reflect deployment, where incorrect tools are frequently related to the query, and prefer challenging distractors over obviously irrelevant negatives.

After the request and ground-truth response are fixed, we embed the query and API descriptions and retrieve the top 20 candidates by similarity.
A strong LLM judge (Amazon's Nova-pro-v1:0) removes candidates that could plausibly answer the request on their own, including near-duplicates of the target, so the annotated call remains the intended solution.
Among the remaining non-target APIs, we keep those with retrieval scores above an approximate elbow in the sorted similarity curve, dropping weakly related easy negatives; if no elbow is found, we retain the top candidate.
We provide full details in Appendix \ref{sec:distractors_retrieval}.

\section{Analysis}
\label{sec:Analysis}
In this section, we analyze the diversification benefits of our approach in comparison to recently published methods for function-calling data generation. We evaluate diversity using multiple metrics that capture both linguistic variation and argument diversity. In addition, we assess data correctness to verify that increased diversity does not come at the cost of noisy or low-quality data.

Our primary comparison is against two SoTA synthetic generation pipelines, ToolAce~\cite{liu2409toolace} and APIGen~\cite{liu2024apigen}, evaluated on their publicly released corpora of approximately 20K and 60K examples, respectively. We chose these baselines because they obtain SoTA performance on public benchmarks and emphasize single-turn interactions.

Since these releases were not designed to maximize linguistic or argument diversity, we also compare against post-hoc \emph{LLM paraphrase} and \emph{entity swapping} augmentations of each release (prompt specifications in Appendix~\ref{sec:prompts-augmentation-baselines}). Augmented corpora are not refiltered through our generation-time judge, so these comparisons are conservative.

For intrinsic diversity, we compare our dataset against the strongest post-hoc augmentation of each baseline release per metric (see below). We report entity-swapped ToolAce and base APIGen for argument diversity, and paraphrased ToolAce and base APIGen for linguistic diversity; full results for all augmentation variants appear in Appendices~\ref{app:Argument values comparison tables}, \ref{app:Linguistic diversity}.

Throughout the analysis, we use a dataset of 1{,}793 examples generated by our procedure. Our goal is to assess whether diversity-oriented generation improves linguistic and argument diversity without sacrificing correctness; to reduce costs, we therefore did not generate data at the scale of the baseline methods, as we are not attempting to train a SoTA function-calling LLM.
During generation, argument values are optimized using \texttt{Cluster Entropy}, which measures the entropy of cluster assignments. Linguistic diversity is optimized using a set of lexical, syntactic, and semantic metrics: \texttt{Type–Token Ratio (TTR)} and \texttt{Compression Ratio} (lexical); \texttt{Variance of FKGL} (lexical/syntactic); \texttt{Parse Tree Entropy} and \texttt{Variance of Text Length} (syntactic); and \texttt{Paraphrase Variety}, \texttt{Chamfer Distance Score}, and \texttt{Vendi Score} (semantic).

We generate four types of examples: \texttt{Single}, \texttt{Parallel}, \texttt{Sequential}, and \texttt{None}, corresponding to requests with a ground-truth response involving a single function call, multiple functions executed in parallel, multiple functions executed sequentially, and zero function calls, respectively.

\subsection{Argument Diversity}
We evaluate argument diversity using two metrics: \textit{Normalized Compression Distance} (NCD), which is based on compression efficiency, and \textit{Cluster Entropy} (CE). Since our generation process explicitly optimizes for CE, NCD serves as a complementary metric. Both metrics take as input a set of argument values of the same type (e.g., city names) and output a diversity score.

In addition to the reference datasets, we include a simple na\"ive baseline obtained by repeatedly prompting an LLM to generate argument values with a temperature of 1.0. Although this baseline is intentionally weak, it provides a useful point of reference for interpreting the range of the diversity metrics.

For a fair comparison, we identify argument types that appear at least 20 times in each dataset and then select types shared between datasets. Since few types appear in all three datasets, we compare our dataset separately against ToolAce and APIGen. For each comparison, we select a uniformly random set of 20 items from each (dataset, argument type) pair, and compute the diversity metrics for that collection. We report the averge metric values, taken over all shared argument types as the overall argument diversity score of a dataset. Table~\ref{tab:combined_averages} reports the average metric values across these groups. We estimate variability using bootstrap standard deviation, computed by repeatedly evaluating each metric on random 80\% subsamples.

Since ToolAce and APIGen did not optimize for argument diversity, we augmented them with entity swapping (E.S.): an LLM proposes four alternative argument values per example and we retain one at random. Entity swapping reduced APIGen's argument diversity, so we report the base release; ToolAce improved slightly and we report the entity-swapped corpus. For further details, see Appendix~\ref{app:Argument values comparison tables}.

% \begin{table}[h]
% \centering
% \small
% \begin{tabular}{@{}lcccc@{}}
% \toprule
% \textbf{Comparison} & \textbf{Ours} & \textbf{ToolAce} & \textbf{APIGen} & \textbf{Temp 1.0} \\
% \midrule
% O/T, NCD & \textbf{0.548} & 0.483 & N/A & 0.371 \\
%          & \textbf{(0.007)} & (0.014) &  &(0.020) \\
% O/T, CE & \textbf{\underline{4.322}}  & 3.396  & N/A & 1.846  \\
%         & \textbf{\underline{(0.000)}} & (0.155) &    & (0.180) \\
% O/A, NCD & \textbf{0.549} & N/A & 0.486& 0.417 \\
%          & \textbf{(0.010)} &   & (0.013) & (0.016) \\
% O/A, CE & \textbf{\underline{4.322}} & N/A & 3.843 & 2.346\\
%         & \textbf{\underline{(0.000)}} &   & (0.101) & (0.153) \\
% \bottomrule
% \end{tabular}

\begin{table}[h]
\centering
\small
\resizebox{\columnwidth}{!}{%
\begin{tabular}{@{}lcccc@{}}
\toprule
\textbf{Comparison} & \textbf{Ours} & \textbf{ToolAce (E.S.)} & \textbf{APIGen} & \textbf{Temp 1.0} \\
\midrule
O/T, NCD & $\mathbf{0.55_{0.01}}$ & $0.50_{0.01}$ & N/A & $0.37_{0.02}$ \\
O/T, CE  & $\mathbf{4.32_{0.01}}$ & $3.44_{0.15}$ & N/A & $1.85_{0.18}$ \\
% O/T, CE  & $\mathbf{\underline{4.32_{0.01}}}$ & $3.44_{0.15}$ & N/A & $1.85_{0.18}$ \\
O/A, NCD & $\mathbf{0.55_{0.01}}$ & N/A & $0.49_{0.01}$ & $0.42_{0.02}$ \\
O/A, CE  & $\mathbf{4.32_{0.01}}$ & N/A & $3.84_{0.10}$ & $2.35_{0.15}$ \\
% O/A, CE  & $\mathbf{\underline{4.32_{0.01}}}$ & N/A & $3.84_{0.10}$ & $2.35_{0.15}$ \\
\bottomrule
\end{tabular}}
\caption{Argument diversity metrics of our method compared against ToolAce (O/T) or APIGen (O/A). \textbf{Bold values} indicate statistically significant superiority. 
% \underline{Underlined values} indicate the maximum value for CE over 20 elements, $\log_2(20) = 4.32$.
}
% Average parameter diversity metrics comparing our method against the strongest baseline per comparison and a na\"ive sampler. `O/T' denotes Ours vs.\ ToolAce (entity swap, E.S.)\ and the na\"ive baseline; `O/A' denotes Ours vs.\ the original APIGen release and the na\"ive baseline. 
% For APIGen, entity swapping reduces argument diversity relative to the original release, so the original is used. NCD and CE denote NCD Diversity and Cluster Entropy.
% Values show mean and bootstrap standard deviation (in subscript). \textbf{Bold values} indicate statistically significant superiority. \underline{Underlined values} indicate the maximum value for CE over 20 elements, $\log_2(20) = 4.32$. The ToolAce entity-swapped corpus is not refiltered through our generation-time judge; see Appendix~\ref{app:Argument values comparison tables} for per-group scores.}
\label{tab:combined_averages}
\end{table}
%\vspace{-0.1in}

Our method consistently achieves higher diversity than both baselines reported in \Cref{tab:combined_averages}. Notably, the improvement in NCD Diversity over ToolAce (E.S.)\ is comparable in magnitude to the gap between ToolAce (E.S.)\ and the na\"ive baseline. We confirmed that the diversity gains are not due to incorrect examples by manually inspecting the generated argument values.

\begin{figure}[htbp]
    \centering
    \includegraphics[width=0.2\textwidth]{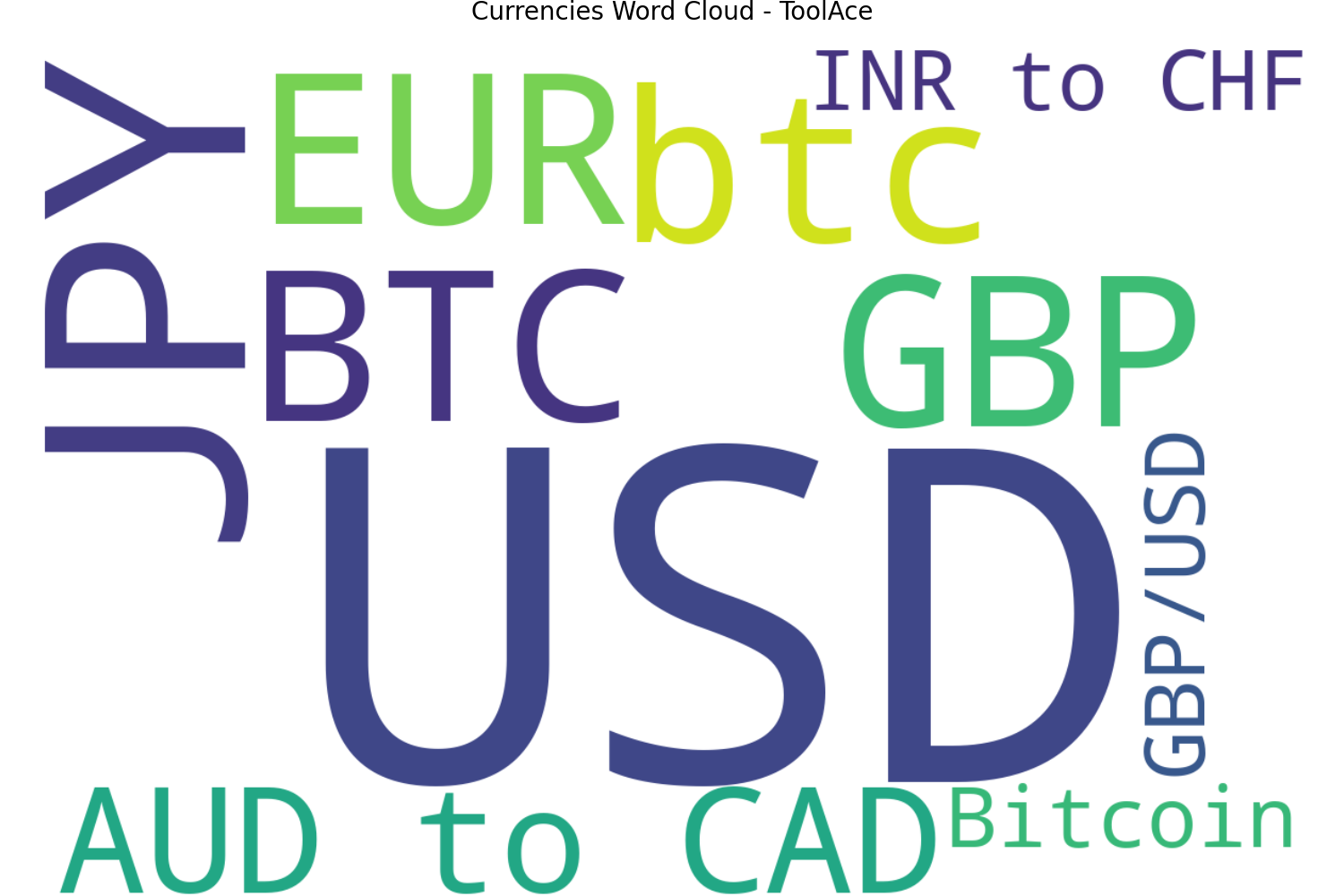}
    \hfill
    \includegraphics[width=0.2\textwidth]{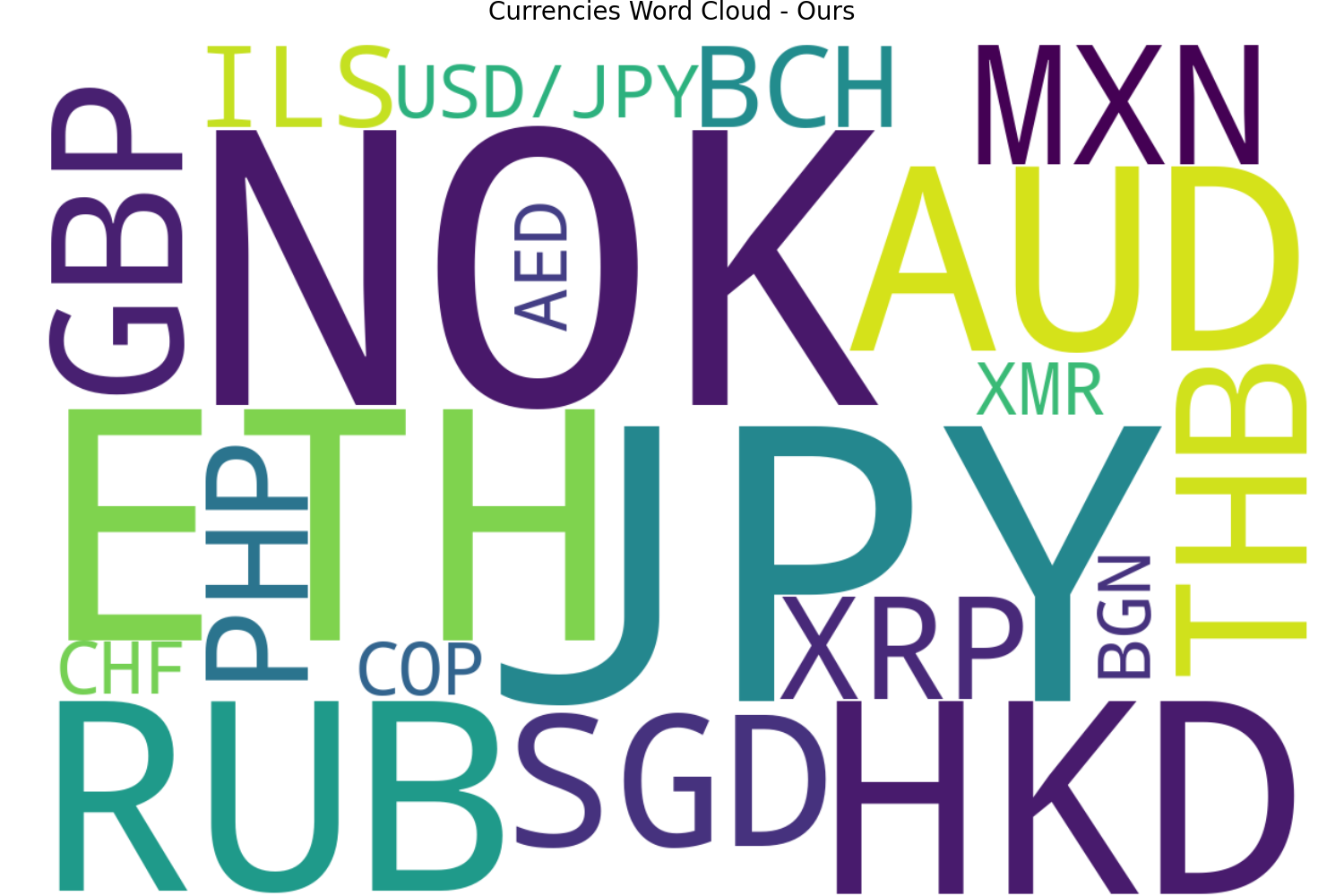}
    \caption{Currency word-cloud for ToolAce (left) and our method (right).}
    \label{fig:currency word-clouds}
    \vspace{-0.5cm}
\end{figure}

For a qualitative illustration, \Cref{fig:currency word-clouds} shows word clouds for a currency argument from our dataset and ToolAce. In ToolAce, the distribution is heavily dominated by USD, whereas our method yields a more balanced distribution across currencies.

% Per-group NCD and cluster-entropy scores, including entity-swapped releases, are in Appendix~\ref{app:Argument values comparison tables}.

\subsection{Linguistic Diversity}
To evaluate linguistic diversity, we compare user requests from our dataset against size-matched samples from ToolAce and APIGen. For ToolAce, we sampled an equal number of examples from each query type (\texttt{single}, \texttt{parallel}, \texttt{sequential}, and \texttt{none}). Since APIGen does not include \texttt{sequential} queries, we proportionally sampled from the remaining query types. 

To improve the linguistic diversity of the baselines, we applied LLM paraphrase: an LLM proposes four rephrasings of each user query that preserve the ground-truth tool call, and we retain one at random. Paraphrasing improved ToolAce slightly while APIGen was unchanged or slightly worse; we therefore report paraphrased ToolAce and base APIGen below. 

% As the representative augmented baselines we use LLM-paraphrase ToolAce and entity-swapped APIGen, the stronger augmentation variant for each release (see opening of \S\ref{sec:Analysis}); augmented corpora are not refiltered through our generation-time judge.
%
We additionally evaluate linguistic diversity on the BFCL benchmark, focusing on its non-live\footnote{The non-live subset refers to the original, expert-curated examples from BFCL V1~\cite{patilberkeley}, consisting of static, researcher-designed function-documentation and query pairs.} Python function-calling subset, which includes 1{,}240 examples spanning \texttt{single}, \texttt{parallel} and \texttt{zero} queries.
% (as detailed in \S\ref{sec: method extensions}).

Beyond the metrics explicitly optimized during generation (see \S\ref{sec:Our Methods}), we also evaluate \texttt{Simpson’s Index} (lexical) \cite{simpson1949measurement}, \texttt{Tree Edit Distance} (syntactic), and \texttt{Cluster Entropy} and \texttt{Semantic Spread} (semantic).
% ; formal definitions appear in Appendix~\ref{app:Linguistic diversity}.
As with argument diversity, we estimate variability via bootstrapping. To ensure that increased diversity does not introduce noise, we assess correctness following the methodology of \citet{iskander2024quality}. Specifically, we manually annotate 100 randomly sampled queries from each dataset after removing dataset identifiers. The results indicate comparable correctness across our dataset, ToolAce, and APIGen.
% (Appendix~\ref{app:Manual evaluation}).

Table~\ref{tbl:diversity_scores_toolace_apigen} summarizes the linguistic diversity results against the strongest augmented baselines for the metrics we did not optimize for. Our dataset achieves higher diversity on three of the four held-out metrics with statistically significant margins. The exception is Cluster Entropy, where ToolAce (Para.)\ edges our dataset by a small margin ($10.80$ vs.\ $10.77$). For APIGen, the entity-swapped variant is approximately on par with the base release on most metrics, so the comparison is only marginally more conservative than using the original release. The same trend, our dataset leading on the majority of dimensions, holds for BFCL. The detailed metric definitions and experiments are available in Appendix~\ref{app:Linguistic diversity}.

\begin{table}[htbp]
\centering
\small
\resizebox{\columnwidth}{!}{%
\begin{tabular}{@{}lccc@{}}
\toprule
\textbf{Metric} & \textbf{Ours} & \textbf{ToolAce (Para.)} & \textbf{APIGen}\\
\textbf{(Type)} & \textbf{(STD)} & \textbf{(STD)} & \textbf{(STD)}\\
\midrule
Simpson's Index & \textbf{0.9901} & 0.9892 & 0.9859\\
(Lexical) & (\textbf{0.0001}) & (0.0001) & (0.0001)\\
Tree Edit Distance & \textbf{23.3919} & 21.2785 & 19.9884 \\
(Syntactic) & \textbf{(0.1368)} & (0.1603) & (0.1386) \\
Cluster Entropy & 10.7715 & \textbf{10.8037} & 10.6748 \\
(Semantic) & (0.0047) & (\textbf{0.0014}) & (0.0094) \\
Semantic Spread  & \textbf{0.9707} & 0.9542 & 0.9517 \\
(Semantic) & \textbf{(0.0003)}  & (0.0005) & (0.0004) \\
\bottomrule
\end{tabular}}
\caption{Linguistic diversity scores. Bold values mark the highest score in a row that is statistically significant with 95\% confidence.}

% Linguistic diversity scores (non-optimized metrics) comparing our dataset against the strongest augmentation variant per baseline: ToolAce with LLM paraphrase and APIGen with entity swap (E.S.). Bootstrap standard deviations from 100 subsamples of 80\% of data. Bold values mark the highest score in a row that is statistically significant at $\alpha = 0.05$ under $|\mu_1 - \mu_2| > 1.96\sqrt{\sigma_1^2 + \sigma_2^2}$. Augmented corpora are not refiltered through our generation-time judge. The full seven-way comparison (all four augmented variants) is in Appendix~\ref{app:Linguistic diversity}.}
\label{tbl:diversity_scores_toolace_apigen}
\end{table}

We further ablate the iterative diversity-optimization procedure used during \emph{user request} generation (\S\ref{sec:diverse gen alg}), comparing our multi-round setting to a one-round baseline on two corpora of 90 queries each. Multi-round optimization yields statistically significant gains on most generation-time metrics and on several held-out metrics under the bootstrap criterion we use elsewhere. This motivates our choice of using multiple rather than single round for increasing linguistic diversity (Detailed results in Appendix~\ref{app:ablation_linguistic_rounds}).

% ; compression ratio diversity, Simpson's index, and tree edit distance do not differ significantly at $\alpha = 0.05$ between the two settings. Parameter-value generation uses a single round for cost reasons (\S\ref{sec: method single func}) (Appendix~\ref{app:ablation_linguistic_rounds}).

In addition to the quantitative analysis, we provide qualitative examples illustrating linguistic diversity: multiple queries for the same ground-truth response that vary in formality, length, contextual depth, and user personas (see Appendix~\ref{app:Qualitative table}).

\section{Application: LLM Fine-tuning}
\label{sec:LLM Fine-tuning}
\begin{table*}[ht]
\centering
\small
\begin{tabular}{lcc|lcc|lcc}
\toprule
\multicolumn{3}{c}{\textbf{Evaluated on APIGen}} & \multicolumn{3}{c}{\textbf{Evaluated on ToolAce}} & \multicolumn{3}{c}{\textbf{Evaluated on Our Dataset}} \\
\cmidrule(r){1-3} \cmidrule(lr){4-6} \cmidrule(l){7-9}
\textbf{Query Type} & \textbf{Ours} & \textbf{ToolAce} & \textbf{Query Type} & \textbf{Ours} & \textbf{APIGen} & \textbf{Query Type} & \textbf{ToolAce} & \textbf{APIGen} \\
\midrule
Single     & 89.6          & 89.1          & Single     & \textbf{84.4} & 71.7          & Single     & \textbf{65.9} & 51.4 \\
Parallel   & \textbf{85.8} & 70.7          & Parallel   & 61.3          & \textbf{70.0} & Parallel   & 14.2          & \textbf{36.9} \\
Sequential & —             & —             & Sequential & 44.0          & 42.6          & Sequential & \textbf{39.8} & 27.0 \\
None       & 31.2          & 28.5          & None       & 54.3          & 56.3          & None       & 40.5          & \textbf{69.4} \\
\midrule
All        & \textbf{73.2} & 69.0          & All        & \textbf{67.0} & 63.2          & All        & 46.8          & 48.8 \\
\bottomrule
\end{tabular}
\caption{LLM-judge accuracy comparison across three evaluation environments (APIGen, ToolAce, and Our Dataset) partitioned by query type. Bold values indicate statistically significant differences ($p < 0.05$) based on McNemar's test. }
\label{tab:combined_three_datasets_full}
\end{table*}
\subsection{Setup}
Next, we applied our generated examples for improving the function-calling abilities of LLMs. 
Towards this goal, we used the same four datasets from \S\ref{sec:Analysis}: our generated dataset, the subsampled ToolAce and APIGen datasets, and the BFCL test set. We do not fine-tune on the LLM-augmented ToolAce/APIGen variants from \S\ref{sec:Analysis}; those corpora are included only in the intrinsic diversity study.
We fine-tuned Llama-3.1-8B-Instruct \citep{grattafiori2024llama3herdmodels} using LoRA \citep{hu2022lora} for each of the three training sets described above, resulting in one model per training set.
We then evaluated each model on the three remaining out-of-distribution (OOD) datasets on which it was not trained, to assess its generalization ability. 
To compare performance over ToolAce, APIGen, and our dataset, we used Amazon’s Nova-pro-v1:0, a large instruction-tuned language model, as an LLM-as-a-judge to assign a binary correctness label to each model response relative to the ground-truth. For evaluation over BFCL, we used their provided AST metric, which also assigns a binary score per example. In both cases, we report mean performance over the test sets and use McNemar’s test to assess whether pairwise comparisons yield statistically significant differences. We consider a model to be significantly better when the resulting p-value is below $0.05$. For BFCL, which involves multiple comparisons, we apply a Holm–Bonferroni–corrected McNemar’s test with a significance threshold of $p < 0.05$.

\subsection{Results}

\begin{table}[ht]
\centering
\resizebox{\columnwidth}{!}{
\begin{tabular}{lcccccccc}
\toprule
\textbf{Category} & \textbf{Ours} & \textbf{ToolAce} & \textbf{APIGen} & \textbf{No fine-tuning} \\
\midrule
SIMPLE\_PYTHON & 82.0 & 78.3 & 61.5 & 50.0 \\
PARALLEL & 82.0 & 77.5 & 72.5 & 48.5 \\
MULTIPLE & \textbf{91.0} & 79.0 & 82.0 & 53.0 \\
PARALLEL\_MULTIPLE & 80.0 & 55.0 & 83.5 & 34.5 \\
IRRELEVANCE & 92.9 & 95.4 & 97.5 & 1.3 \\
\midrule
ALL & \textbf{85.2} & 77.8 & 77.1 & 38.3 \\
\bottomrule
\end{tabular}
}
\caption{AST accuracy on BFCL by category for models fine-tuned on our dataset, ToolAce, and APIGen, plus Llama-3.1-8B-Instruct without dataset-specific fine-tuning. Bold indicates statistically significant differences ($p < 0.05$) under McNemar’s test with Holm–Bonferroni correction.}
\label{tab:bfcl_benchmark}
\end{table}

\Cref{tab:combined_three_datasets_full} presents the OOD results over the test sets for the three compared methods, i.e., APIGen, ToolAce, and our method, respectively. In other words, for each dataset, its corresponding table compares the performance of models fine-tuned using data generated by the two other methods. 

As shown in the left and center panes of \Cref{tab:combined_three_datasets_full}, our fine-tuned model achieves statistically significant improvements over the other OOD fine-tuned models, with a lift of $4.2$ accuracy points over the ToolAce model on the APIGen test set and a $3.8$ accuracy points lift over the APIGen model on the ToolAce test set. 
%When diving into the partition into query types, the advantage of our model is especially pronounced for the more challenging multiple-function invocations over both test sets, where our models achieves lifts of $15-30\%$ compared to the other models.

The right pane of \Cref{tab:combined_three_datasets_full} shows a comparison of the ToolAce and APIGen models on our dataset. A key takeaway from this table is that our dataset is much more challenging in comparison to either of the other datasets, with accuracy of only $46.8\%$ and $48.8\%$ by the models fine-tuned on ToolAce and APIGen, respectively (compared to $>63\%$ elsewhere). This lower cross-dataset accuracy suggests our benchmark is substantially harder; as models continue to improve, such difficulty helps avoid premature saturation and keeps evaluation informative.
% \os{do we have an explanation for that? Is it because the increased diversity? Is thia a good thing?}\dan{I don't think we can attribute it specifically to the increased diversity. At least one competing explanation would be the increased difficulty of the distractors, and there are probably other explanations as well. I do think its a good thing to have a more challenging dataset, in general, so long as its "correct" - and we do say that it is correct, since we've done (a) correctness checks specifically, and (b) our extrinsic performance on APIGen, ToolAce and BFCL is decent - that is not as likely to occur if the dataset is "incorrect".}

Finally, we test all three fine-tuned models on the BFCL (non-live) benchmark. As a fourth baseline, we also evaluate Llama-3.1-8B-Instruct without dataset-specific LoRA fine-tuning, to provide a reference point before task adaptation.
Table \ref{tab:bfcl_benchmark} shows that the model fine-tuned on our training data significantly outperforms both the ToolAce and the APIGen models ($85.2\%$ vs. $77.8\%$ and $77.1\%$ respectively) under a Holm–Bonferroni–corrected McNemar test.

Fine-tuning hyperparameters and training-library details are in Appendix \ref{sec:fine_tuning_details}. The LLM-as-a-judge evaluation prompt is in Appendix \ref{sec:prompts_llm_judge_eval}.

\section{Conclusions and Future Work}
\label{sec:Conclusions}

In this work, we identified a weakness in current data generation methods for function-calling agents, namely the lack of both linguistic and argument diversity. We proposed a data generation technique, based on a novel component generating diverse values, and evaluated its effectiveness in the scenario of general purpose functions. Our evaluations demonstrate the utility of our methods, both directly through higher diversity scores on the large majority of standard metrics (including against augmented baseline variants), and via its downstream impact, showing superior OOD  performance for models fine-tuned on our data (roughly $4\%-7\%$ accuracy improvement).

We believe that our core component of generating diverse items is interesting in its own right, and in future works plan to evaluate it in additional settings.

\section*{Limitations}
% Our work has several limitations. First, portions of the manuscript were polished using AI-assisted text tools, which may introduce subtle stylistic biases; however, all core content, experiments, and claims remain fully human-designed and verified. Second, we provide only a high-level description of the human annotation procedure and omit step-by-step instructions or screenshots. While this limits exact reproducibility, the annotations are used solely to corroborate our claim that increased diversity does not degrade dataset quality and are not central to our contribution. Third, our experiments are conducted on selected function libraries in English, such as stock-market, travel, and calendar domains, and the approach’s effectiveness on other languages or multilingual settings remains untested. Fourth, we did not invest in optimizing our synthetic data generation algorithm for parallel function calls, which may affect efficiency and coverage in settings with complex multi-function sequences. Finally, our diversity evaluation relies on standard semantic, lexical, and syntactic metrics, which may not capture all aspects of meaningful variation. We mitigate this by complementing intrinsic metrics with extrinsic evaluation via LLM fine-tuning and out-of-distribution testing, but further work could explore additional metrics, multi-lingual settings, and human-centered evaluation.

% \begin{itemize}
%     \item multi-turn
%     \item 
% \end{itemize}

Our work provides linguistically diverse queries; however, our analysis is restricted to English. Extending the approach to additional languages, particularly low-resource ones, would likely require further effort. Moreover, our generation process relies on strong LLMs to filter invalid examples, making it more costly than approaches that depend solely on smaller-scale models. Regarding the structure of the data, while prior work has demonstrated the generation of multi-turn interactions with function-calling agents, our study is limited to single-turn interactions. We intentionally focus on this setting to avoid confounding factors introduced by multi-turn designs, which are orthogonal to our core contribution of increasing diversity. Nevertheless, extending our approach to multi-turn interactions is a natural and promising direction for future work. Lastly, due to the cost of high-quality LLMs, we avoided generating very large datasets (i.e., $20$K-$60$K examples as opposed to roughly $2$K) comparable to that of previous works. As a result we do not have concrete empirical evidence that our methods retain their advantage at that scale. Nevertheless, recent work suggests that the diversity gains we observe are meaningful in their own right, since diverse datasets can be more useful than substantially larger but less diverse alternatives (see, e.g.,~\cite{nguyen2025we} and references therein). %This being said, multiple recent papers (see, e.g.,~\cite{nguyen2025we} and references therein) demonstrate the utility of diverse datasets even compared to less diverse but larger counterparts. \ca{suggestion: }

\bibliography{paper}

\clearpage
\newpage

\appendix

\section{Detailed Method}\label{sec:detailed_method}

This section provides a comprehensive explanation of our data generation pipeline, which consists of five main parts: (i) preprocessing, (ii) function selection, (iii) argument value generation, (iv) query generation, and (v) distractor selection. 

We begin by describing the preprocessing part, that is divided into 3 steps in Appendices~\ref{sec:parameter_grouping}, \ref{sec:api_pools}, and \ref{sec:function_graph}.

\subsection{Preprocessing: Parameter Grouping}\label{sec:parameter_grouping}
We cluster function parameters with similar semantic meanings across different APIs to ensure systematic diversity in parameter value generation. For example, we group location-related parameters such as \textit{destination} from \texttt{book\_hotel()} with \textit{city} from \texttt{get\_weather\_forecast()}, and temporal parameters such as \textit{check\_in\_date} from \texttt{book\_hotel()} with \textit{departure\_date} from \texttt{book\_flight()}. Other semantic clusters include identifiers (user IDs, session tokens), numerical ranges (prices, counts, ratings), and format specifications (file types, units of measurement).

% This grouping serves a critical purpose: by tracking and generating diverse values at the parameter group level rather than for individual parameters in isolation, we ensure that when our dataset contains multiple queries with semantically equivalent parameters (e.g., across different location-based APIs), the generated values collectively cover a diverse range rather than repeatedly using the same values. This approach reduces the total number of dataset entries required to achieve comprehensive parameter value diversity.

We perform parameter grouping once as a preprocessing step and reuse the results during subsequent parameter generation. The procedure begins by extracting parameter metadata from each API, including names, types, descriptions, and enumerable values where applicable. Parameters are categorized into four types: NUMERICAL, STRING, ENUM, and OTHER. 
Note that the OTHER category typically has a rich structure, e.g.,\ a list or dictionary.
Within each category, we cluster semantically similar parameters by generating a descriptive sentence for each parameter, then using embedding-based similarity to create clusters. Algorithm~\ref{alg:param_grouping} details the complete procedure.

\ULforem
\normalem
\begin{algorithm}
\caption{Parameter Grouping via Embedding Similarity}\label{alg:param_grouping}
\KwIn{Set of parameters $P$ with metadata}
\KwOut{Grouped parameters $G = \{G_1, G_2, \ldots, G_k\}$}

$\text{embeddings} \leftarrow \emptyset$\;

\ForEach{parameter $p \in P$}{
    Construct description: ``The \textit{p.name} parameter is a \textit{p.type} that \textit{p.description}''\;
    
    \If{$p.type = \text{ENUM}$}{
        Append: `` and must be one of: \textit{p.enum\_values}''\;
    }
    
    $e_p \leftarrow \text{encode}(\text{description})$ using all-MiniLM-L6-v2\;
    $\text{embeddings} \leftarrow \text{embeddings} \cup \{e_p\}$\;
}

$\text{ungrouped} \leftarrow P$, $G \leftarrow \emptyset$\;

\ForEach{parameter $p \in P$}{
    \If{$p \in \text{ungrouped}$}{
        $G_{\text{new}} \leftarrow \{p\}$\;
        $\text{ungrouped} \leftarrow \text{ungrouped} \setminus \{p\}$\;
        
        \ForEach{parameter $q \in \text{ungrouped}$}{
            \If{$\cos(e_p, e_q) \geq 0.6$}{
                $G_{\text{new}} \leftarrow G_{\text{new}} \cup \{q\}$\;
                $\text{ungrouped} \leftarrow \text{ungrouped} \setminus \{q\}$\;
            }
        }
        
        $G \leftarrow G \cup \{G_{\text{new}}\}$\;
    }
}

\Return{$G$}
\end{algorithm}
\ULforem

\subsubsection{Cosine threshold calibration}\label{sec:param_embed_threshold_calib}
The edge threshold $\tau = 0.6$ in Algorithm~\ref{alg:param_grouping} was chosen heuristically and then checked empirically rather than from a closed-form criterion. We inspected the resulting clusters for two properties: groups were large enough to be useful for diversity tracking across APIs, and we did not see obvious false positives where clearly distinct parameter semantics were merged (e.g., a free-text \textit{city name} field grouped with a constrained \textit{airport code} field).

To calibrate $\tau$ against the bulk of pairwise similarities, we computed cosine similarity for every unordered pair of parameters in our schema set, yielding $98{,}880{,}412$ pairs. Summary statistics of the resulting distribution are as follows: mean $0.3442$, median $0.3349$, $25$th percentile $0.2753$, $75$th percentile $0.4005$, standard deviation $0.1046$. The empirical distribution is approximately Gaussian. The value $\tau = 0.6$ corresponds to percentile rank $97.895$, i.e., only about $2.1\%$ of all pairs exceed the threshold, so graph edges are sparse and the cutoff is strict relative to a random pair of parameters (whose similarity is typically near the mean above).

We also experimented with agglomerative clustering on the same embeddings (scikit-learn's \texttt{AgglomerativeClustering}); linkage and distance choices that produced comparable edge density gave similar groupings in spot checks, so we kept the simpler connected-threshold construction for reproducibility and speed.

\subsection{Preprocessing: API Pools}\label{sec:api_pools}
For each dataset example, we must select which API(s) to target. Our approach supports five execution types: SINGLE (one API call), PARALLEL (multiple API calls that can be performed in parallel), SEQUENTIAL (chained API calls where outputs feed into inputs), MISSING\_PARAMS (API call with intentionally missing required parameters), and NONE (no API should be called). To facilitate sampling of APIs for SINGLE and MISSING\_PARAMS queries, we construct three API pools with different sampling characteristics.

\textbf{Pool Construction:}
\begin{itemize}
    \item \textit{General pool:} Contains all available APIs for uniform random sampling.
    \item \textit{Focused pool:} Contains APIs with parameters belonging to large parameter groups (randomly selected, with weights proportional to group size), prioritizing APIs with commonly used parameter patterns.
    \item \textit{Other pool:} Contains APIs with parameters of type OTHER that don't fit standard type categories.
\end{itemize}

This three-pool design increases the probability of generating multiple values for a single parameter type (via the focused pool) and selecting APIs with non-standard parameters (via the other pool), both of which contribute to dataset diversity and difficulty.

\subsection{Preprocessing: API Similarity Graph}\label{sec:function_graph}
The generation of PARALLEL and SEQUENTIAL queries requires sampling multiple related functions. To ensure that sampled functions are semantically related and produce natural queries, we follow \citet{wang2024toolflow} and construct a graph where vertices represent APIs and edges connect APIs with sufficiently similar parameters.
An edge between two APIs is formed if the similarity between a pair of their parameters exceeds a threshold. We distinguish between two edge types: P-P edges connect APIs with similar input parameters, while P-R edges connect APIs where an input parameter of one API is similar to the output parameter of another. This distinction enables us to bias sampling toward edges that are more appropriate for each query type: P-P edges for PARALLEL queries, where APIs operate on similar inputs, and P-R edges for SEQUENTIAL queries, where the output of one API serves as input to another.

\subsection{Function Sampling}\label{sec:api_sampling}
Using the API pools and similarity graph constructed in preprocessing, we sample APIs for our queries according to their execution type.

For SINGLE and MISSING\_PARAMS queries, we uniformly sample one of the three API pools, then uniformly sample an API from the selected pool.

For PARALLEL and SEQUENTIAL queries, we perform a biased random walk on the API similarity graph. We first sample the walk length $\ell$ by drawing $N \sim \text{Poisson}(\lambda=0.75)$ and setting $\ell = N+2$. During the walk, P-P edges receive higher selection probability for PARALLEL queries (ensuring multiple independent but related APIs), while P-R edges receive higher probability for SEQUENTIAL queries (ensuring compatible input-output chains). For SEQUENTIAL queries, the order of sampled APIs determines their execution order in the ground truth answer.

For PARALLEL and SEQUENTIAL queries, we apply additional rejection sampling steps to ensure quality. In particular, we generate a description string from each API's schema and verify that the average pairwise similarity between sampled APIs exceeds a threshold. Note that this differs from the graph construction, where edges are based on individual parameter similarities; here we measure similarity between complete API descriptions. Additionally, for SEQUENTIAL queries, we use an LLM judge to verify that the input and output parameters of consecutively sampled APIs are compatible. 

\subsection{Parameter Value Generation}\label{sec:parameter_val_generation}
After sampling APIs, we generate ground truth parameter values. For each API, we generate appropriate parameter values while ensuring that the collection of values across our dataset exhibits diversity within each semantic parameter group identified in Appendix \ref{sec:parameter_grouping}.

% This target-first approach serves two purposes: (1) generating a natural language query to match a predetermined API call and parameter set is significantly more reliable than extracting API calls and parameters from natural language queries, and (2) by controlling parameter value generation directly, we can systematically ensure diversity across parameter groups rather than relying on diverse queries to naturally produce diverse parameter coverage.

We generate parameter values sequentially within each API call to ensure valid parameter combinations. Functions often have implicit constraints on legal parameter combinations (e.g., a check-out date must follow the check-in date), so we provide the LLM with previously generated parameters for the current API call as context when generating each subsequent parameter. This sequential generation allows the LLM to maintain consistency and validity across all parameters within a single API call.

\subsubsection{When to Apply Diversity Optimization}
Not all parameters should be optimized for diversity. Within a single API, we seek diversity among parameters only to the extent that it does not create contradictions. For instance, in a hotel booking API, the number of guests and number of rooms must be compatible—requesting 10 guests with only 1 room would be unrealistic. Similarly, a flight search should not combine a ``business class'' seat preference with a ``budget airline'' filter.

For queries requiring multiple APIs (PARALLEL and SEQUENTIAL), we optimize diversity for only one API, then generate parameters for the remaining APIs to match the existing parameter set. For PARALLEL queries, we randomly select which API receives diverse parameter generation. For SEQUENTIAL queries, we always apply diversity optimization to the last API in the chain, as its parameters are not constrained by downstream API requirements. For these non-optimized APIs, we prompt the LLM to generate cohesive parameter values that are contextually appropriate given the existing parameters and API schemas from other APIs in the query, rather than maximizing diversity.

\subsubsection{Generation Strategy by Parameter Type}

The generation strategy varies by parameter type:
\begin{itemize}
    \item \textbf{ENUM parameters:} Sample uniformly from predefined valid values.
    \item \textbf{OTHER parameters:} Use the LLM to generate a single appropriate value considering existing parameter context and group diversity.
    \item \textbf{NUMERICAL and STRING parameters with diversity optimization:} Generate multiple candidates and select the one that maximizes diversity within the parameter's semantic group (detailed below).
    \item \textbf{NUMERICAL and STRING parameters without diversity optimization:} Use the LLM to generate a single cohesive value given existing parameters and API context.
\end{itemize}

\subsubsection{Diversity-Optimized Generation Procedure}
For NUMERICAL and STRING parameters where diversity optimization is enabled, we employ a ``virtual augmentation'' strategy: the LLM internally generates 25 candidate values, from which we randomly select 5 to return as the candidate set. The remaining 20 values are combined with existing group values to form an augmented set for diversity evaluation. This approach increases the likelihood of selecting tail values even when the actual parameter group is small. For example, without virtual augmentation, sampling just two cities might yield ``New York City'' and ``London''—values that are technically different but not diverse in a broader sense. Virtual augmentation makes it more likely to select less common values like ``Nairobi'' early in the sampling process, rather than requiring many samples before tail values emerge.

Algorithm~\ref{alg:diverse_param_val_gen} details the complete parameter generation procedure when diversity is required, including validation to ensure type correctness, constraint satisfaction, and logical consistency with the API schema.

\ULforem
\normalem
\begin{algorithm}[h]
\caption{Diverse Parameter Value Generation}\label{alg:diverse_param_val_gen}
\KwIn{Parameter set $P_a$, groups $G$, API $a$, diversity measure $\mathcal{A}$}
\KwOut{Generated parameter values $Y_{\text{current}}$}
\smaller
$Y_{\text{current}} \leftarrow \emptyset$\;
\ForEach{parameter $p \in P_a$}{
    Retrieve parameter group $g \in G$ where $p \in g$\;
    Retrieve existing values $V_g$ generated for parameters in group $g$\;
    
    \If{$p.type = \text{ENUM}$}{
        $y_p \leftarrow$ sample uniformly from $p.\text{enum\_values}$\;
    }
    \ElseIf{$p.type = \text{OTHER}$}{
        $y_p \leftarrow$ LLM-generate using context:\;
        \hspace{2em} $\{p.\text{name}, p.\text{description}, a.\text{name},$ \\\hspace{2em}$a.\text{description}, Y_{\text{current}}, V_g\}$\;
    }
    \Else{
        Generate 25 candidates $Z = \left\{z_1,\ldots,z_{25}\right\}$ using LLM with context:\;
        \hspace{2em} $\{p.\text{name}, p.\text{description}, p.\text{exact\_type},$ \\ \hspace{2em}$a.\text{name}, a.\text{description}, Y_{\text{current}}, V_g\}$\;
        Randomly select 5 candidates: $X \subset Z$, $\left|X\right| = 5$\;
        Create augmented parameter group: $\bar{V}_g \leftarrow V_g \cup (Z \setminus X)$\;
        
        $y_p \leftarrow \arg\max_{x \in X} \mathcal{A}(\bar{V}_g \cup \{x\}, p.\text{type})$\;
    }
    
    $Y_{\text{current}} \leftarrow Y_{\text{current}} \cup \{y_p\}$\;
}
\textbf{Validate} complete parameter set $Y_{\text{current}}$ using LLM validator\;
\If{validation fails}{
    Track failure information and retry with adjusted parameters\;
}
\Else{
    Update $V_g \leftarrow V_g \cup \{y_p\}$ for each $y_p \in Y_{\text{current}}$ and its matching group $g$\;
}
\Return{$Y_{\text{current}}$}
\end{algorithm}
\ULforem

\subsubsection{Diversity Evaluator}

For NUMERICAL and STRING parameters, we employ a clustering-based entropy metric as our diversity evaluator, defined in Algorithm~\ref{alg:clustering_based_entropy}.

\ULforem
\normalem
\begin{algorithm}[h]
\caption{Parameter Diversity Evaluator}\label{alg:clustering_based_entropy}
\KwIn{Parameter group values $V = \{v_1, \ldots, v_{|V|}\}$, parameter type $t$}
\KwOut{Diversity score (entropy of cluster distribution)}
\If{$t = \text{NUMERICAL}$}{
    Apply DBSCAN on $V$ with $\epsilon = 0.5$, min\_samples $= 2$, Euclidean distance\;
}
\Else{
    Embed $V$ using all-MiniLM-L6-v2\;
    Apply DBSCAN on embeddings with $\epsilon = 0.1$, min\_samples $= 2$, cosine distance\;
}
Obtain DBSCAN clusters $\mathcal{C} = \{C_1, C_2, \ldots, C_k\}$ with sizes $\{c_1, c_2, \ldots, c_k\}$\;
Compute cluster probabilities: $p_i = \frac{c_i}{|V|}$ for $i = 1, \ldots, k$\;
Compute entropy: $H = -\sum_{i=1}^k p_i \log(p_i)$\;
\Return{$H$}
\end{algorithm}
\ULforem

The diversity evaluator uses DBSCAN clustering to identify natural groupings within parameter values, then computes the entropy of the resulting cluster size distribution. Higher entropy indicates more balanced cluster sizes, suggesting greater diversity. By selecting the candidate that maximizes this entropy, we ensure that each new parameter value contributes meaningfully to the overall diversity of its semantic group rather than duplicating existing patterns.

\subsubsection{Execution Type-Specific Considerations}

While the core parameter generation procedure (Algorithm~\ref{alg:diverse_param_val_gen}) applies across all execution types, certain types require additional coordination logic:

\textbf{PARALLEL Execution:} APIs are processed sequentially with cross-API awareness. We enhance each API's description with information about co-occurring APIs and encourage parameter value reuse when logically appropriate (e.g., same location, date, or identifier across a hotel booking and restaurant reservation). Generated parameters are added to diversity trackers only after all APIs in the parallel set pass validation, ensuring the complete parallel call is coherent.

\textbf{SEQUENTIAL Execution:} We generate parameters backward from the last API to prioritize chain cohesion. For each API in reverse order, we first generate its return value from the return schema, then generate its parameters with awareness of both this return value and the downstream API parameters. This ensures parameters for earlier APIs naturally produce return values compatible with later API inputs, maintaining proper data flow throughout the chain.

\textbf{MISSING\_PARAMS Execution:} We use a binomial distribution to determine how many required parameters to intentionally omit (minimum of one). We then apply the standard generation procedure to all required parameters except those selected for omission, plus optionally sampled optional parameters. Omitted parameters are marked with a special sentinel value that the validator recognizes as intentionally absent.

\subsection{Query Generation}\label{sec:query_generation}

After generating (API(s), parameters) tuples, we create natural language user requests that correspond to each tuple. Our diversity metrics at this stage focus on the varied ways users might express the same underlying intent. 
% Since each query has a predetermined target API invocation, semantic diversity is necessarily constrained—each query must request the specific action and parameter values assigned to it. Within this constraint, we optimize for diversity in query tone, brevity, sentence structure complexity, context, and other linguistic attributes that affect how the intent is expressed. Our goal is to make models trained on our dataset robust to the diverse ways an identical API invocation request could be phrased, ensuring the model learns to map varied natural language expressions to the same underlying API call.

% We generate diverse, high-quality queries through an iterative refinement process. 
For each (API(s), parameters) tuple, we conduct 5 rounds of generation, producing 5 candidate queries per round. Each candidate is evaluated by an LLM judge to verify validity, and valid queries are assessed for their impact on overall dataset diversity. The LLM judge results and diversity scores from previous rounds serve as context for subsequent generation attempts, creating a feedback loop that progressively improves both correctness and diversity.

Algorithm~\ref{alg:query_generator} details the complete procedure.

\ULforem
\normalem
\begin{algorithm}[h]
\caption{Iterative Query Generation with Diversity Optimization}\label{alg:query_generator}
\KwIn{(API, parameters) tuple $(a, p)$, existing dataset $D$, diversity estimator $\mathcal{A}()$}
\KwOut{Generated query $q$ or $\emptyset$}
Sample $\min\{10, |D|\}$ reference queries $Q_{\text{ref}}$ from $D$\;
$\text{prev\_attempts} \leftarrow \emptyset$\;
$\text{prev\_judge\_results} \leftarrow \emptyset$\;
$\text{prev\_rank} \leftarrow \emptyset$\;
$\text{best\_query} \leftarrow \emptyset$\;
$Q_{valid} \leftarrow \emptyset$\;
\For{$t = 1$ \textbf{to} $5$}{
    Generate 5 query candidates $Q_t = \{q_1, q_2, q_3, q_4, q_5\}$ using LLM with context:\;
    $\{Q_{\text{ref}}, \text{prev\_attempts}, \text{prev\_judge\_results},$\\ $ \text{prev\_rank}, a, p\}$\;
    
    $\text{judge\_results}_t \leftarrow$ LLM judge evaluation of $Q_t$\;
    $\text{prev\_attempts} \leftarrow \text{prev\_attempts} \cup Q_t$\;
    $\text{prev\_judge\_results} \leftarrow \text{prev\_judge\_results} \cup \text{judge\_results}_t$\;
    
    \ForEach{query $q \in Q_t$}{
        \If{$q$ is judged valid}{
            $Q_{valid} \leftarrow Q_{valid} \cup \left\{q\right\}$
        }
    }
    
    $prev\_rank = \mathcal{A}(D, Q_{valid})$
            
    $best\_query \leftarrow \text{top query according to $prev\_rank$}$
}
\If{$\text{best\_query} \neq \emptyset$}{
    \Return{$\text{best\_query}$}
}
\Else{
    \Return{$\emptyset$}
}
\end{algorithm}
\ULforem

\subsubsection{Judge Validation}

The LLM judge evaluates whether each candidate query reasonably justifies calling the target API(s) with the specified parameters. The judge provides step-by-step reasoning and a binary decision (YES/NO). Validation criteria vary by execution type:

\begin{itemize}
    \item \textbf{SINGLE:} Judge verifies the query naturally leads to calling the target API with the specified parameters.
    \item \textbf{PARALLEL:} Judge verifies all APIs are necessary to fully answer the query and all parameters can be inferred from the query text alone.
    \item \textbf{SEQUENTIAL:} Judge verifies the chain is logical and parameters can be inferred from the query and/or previous API outputs.
    \item \textbf{MISSING\_PARAMS:} Judge verifies the query would justify the API call but lacks information for the missing required parameters.
    \item \textbf{NONE:} No verification is required.
\end{itemize}

For computational efficiency, we may use batch judging to evaluate multiple queries in a single LLM call.

\subsubsection{Diversity Evaluation}

For queries passing judge validation, we compute multiple diversity metrics in parallel using multiprocessing. Each metric measures a different aspect of linguistic variety:

\begin{itemize}
    \item \textbf{Lexical diversity:} Type-Token Ratio (TTR), compression ratio diversity.
    \item \textbf{Syntactic diversity:} Parse tree entropy, variance in query length.
    \item \textbf{Semantic diversity:} Paraphrase variety (average embedding similarity), Chamfer distance score, Vendi score.
    \item \textbf{Readability diversity:} Variance in Flesch-Kincaid grade level (which captures both lexical and syntactic characteristics through word and sentence length patterns).
\end{itemize}

Each metric is computed by temporarily adding the candidate query to the dataset and measuring the resulting diversity. Since these metrics operate on different scales, we combine them using Reciprocal Rank Fusion (RRF): we rank candidates according to each metric separately, then fuse the rankings to produce a final combined ranking. This approach allows metrics to contribute equally regardless of their absolute scale differences.

Algorithm~\ref{alg:dataset_diversity_estimator} details the complete diversity estimation procedure.

\ULforem
\normalem
\begin{algorithm}[h]
\caption{Dataset Diversity Estimation}\label{alg:dataset_diversity_estimator}
\KwIn{Dataset $D$, candidates set $Q = \left\{q_1,\ldots,q_k\right\}$}
\KwOut{Dataset diversity score}
\For{$i = 1$ \textbf{to} $k$}{
    $D_i \leftarrow D \cup \left\{q_i\right\}$
    Compute the following diversity metrics for $D_i$:
    
    \hspace{2em} • Type-Token Ratio (TTR)\;
    \hspace{2em} • Compression ratio diversity\;
    \hspace{2em} • Paraphrase variety (average embedding similarity)\;
    \hspace{2em} • Parse tree entropy\;
    \hspace{2em} • Chamfer distance score\;
    \hspace{2em} • Variance of Flesch-Kincaid Grade Level\;
    \hspace{2em} • Variance of query length in tokens\;
    \hspace{2em} • Vendi score\;
}
Rank candidates by their marginal addition to each metric separately\;
Apply Reciprocal Rank Fusion (RRF) with $k=60$ to combine rankings\;
\Return{combined RRF score}
\end{algorithm}
\ULforem

\subsubsection{Iterative Feedback}

Between generation iterations, we construct contextual feedback that includes:
\begin{itemize}
    \item Diversity score rankings of previous attempts.
    \item Judge reasoning for failed attempts.
    \item Strategic guidance for the next generation round, generated by an LLM given the diversity score ranking and judge reasoning for failed attempts.
\end{itemize}

The feedback prioritizes validation issues if the failure rate is high, or diversity optimization if most queries pass validation. This adaptive feedback mechanism allows the generation process to self-correct and progressively improve quality.

By combining Algorithms~\ref{alg:query_generator} and~\ref{alg:dataset_diversity_estimator}, we generate query sequences that satisfy two key requirements: (1) queries match the target (API(s), parameters) tuple as validated by the LLM judge, and (2) each query generation maximizes the diversity of the resulting dataset according to our multi-metric evaluation.

\noindent\textbf{Quality-control filtering.}\label{sec:pipeline_qc_filtering}
During development we added validation and filtering steps incrementally after observing that raw generations at several intermediate stages frequently failed either automatic checks or informal manual review---typically on the order of 20\%--50\% of items at a given stage before tightening prompts and thresholds (the exact rate depends on the API family and the stage). These components are not optional polish: without judges and filters analogous to those in \Cref{alg:query_generator} and distractor construction in Appendix \ref{sec:distractors_retrieval}, the corpus mixes inconsistent query--call pairs and implausible distractors that would dominate a na\"ive diversity objective. Appendix~\ref{app:ablation_linguistic_rounds} summarizes how this engineering rationale complements the quantitative ablation of multi-round request diversity.

\subsection{Distractor API Retrieval and Query Filtering}\label{sec:distractors_retrieval}
In this final step, we select distractor APIs for each query to create realistic training examples where the model must distinguish between valid and invalid API calls.
% The goal is to include APIs that are semantically similar to the query but should not be invoked, forcing the model to learn fine-grained distinctions between relevant and irrelevant APIs.

The distractor selection process varies by execution type to ensure appropriate difficulty and validity of the final training examples. We describe the common pipeline first, then detail execution-specific variations.

\subsubsection{Common Pipeline}

All execution types follow this core pipeline, with variations detailed in subsequent subsections:

\begin{enumerate}
    \item \textbf{Candidate Retrieval:} Use semantic similarity retrieval (embedding-based) to find APIs similar to the query. Each retrieved API receives a similarity score from the retriever.
    
    \item \textbf{Plausibility Assessment:} Use an LLM-based judge to rate how plausible each candidate API is as an answer to the query on a scale of 1--5, where higher ratings indicate the API could plausibly answer the query.
    
    \item \textbf{Plausibility Filtering:} Remove candidates that could plausibly answer the query, retaining only those that represent poor matches and serve as effective distractors.
    
    \item \textbf{Elbow Method Filtering:} Sort remaining candidates by retrieval similarity score in descending order, use gradient-based elbow detection to identify the natural cutoff point where similarity scores drop significantly, and include all candidates up to the elbow point while ensuring a minimum number of distractors. Target API(s) are excluded from this filtering but always included in the output.
    
    \item \textbf{Shuffle:} Randomly shuffle the final candidate list to avoid positional bias during training.
\end{enumerate}

\subsubsection{Execution Type-Specific Procedures}

\textbf{SINGLE Execution Type.} For single API calls, we ensure the target API is included during retrieval. After plausibility assessment, we remove non-target candidates with plausibility ratings $\ge 3$, retaining only ratings of 1--2.

\textbf{NONE Execution Type.} For negative examples where no API should be called, we retrieve candidates without ensuring any specific target is included. The plausibility filtering threshold remains at rating $\le 2$ to ensure all included candidates are clearly not relevant to the query, creating a proper negative example.

\textbf{MISSING\_PARAMS Execution Type.} For queries with missing required parameters, we retrieve candidates ensuring the target API (with missing required parameters) is included. We apply stricter plausibility filtering with threshold $\le 1$ instead of $\le 2$ to exclude APIs that might be valid alternatives even with missing parameters, ensuring only clearly invalid/distractor APIs are included alongside the target.

\textbf{PARALLEL Execution Type.} For parallel API calls, we apply plausibility filtering (threshold $\le 2$) with a minimum candidate count for elbow method filtering of at least twice the number of target APIs. After the common pipeline, we validate the parallel invocation: use an LLM to construct alternative API invocations by combining candidates with target APIs, and determine if these alternatives would correctly answer the query. If a valid alternative exists that differs from the target, we handle two cases: (1) if the alternative includes non-target candidates, remove these candidates from the distractor set to prevent ambiguity; (2) otherwise, the alternative represents a strict subset of the current target APIs, indicating redundancy---remove the redundant APIs from the target. We update the execution type accordingly: if the updated target contains only one API, change to SINGLE; if no APIs remain, mark the query as invalid.

\textbf{SEQUENTIAL Execution Type.} For sequential API chains, the procedure mirrors PARALLEL with additional constraints. The minimum candidate count for elbow method filtering is twice the number of target APIs. After the common pipeline, when validating sequential chains, we construct alternatives from candidates and check that chains form valid data flow where outputs of earlier APIs serve as inputs to later APIs. If a valid alternative exists, we update the candidate and target API list following the same logic as PARALLEL.

%\onecolumn
% \newpage
\section{Analysis - Complementary material}\label{app:Analysis}

This appendix provides detailed supporting material for the diversity analysis presented in \S\ref{sec:Analysis}. We organize this material into six main components: A brief review of related work on diversity metrics (Appendix \ref{app: Measuring Diversity - Related work}), qualitative examples demonstrating linguistic variation in generated queries (Appendix \ref{app:Qualitative table}), quantitative comparisons of argument value diversity against baseline methods including LLM entity-swapping augmentations of the released ToolAce and APIGen corpora (Appendix \ref{app:Argument values comparison tables}), comprehensive linguistic diversity metrics including LLM paraphrase and entity-swapped variants (Appendix \ref{app:Linguistic diversity}), ablations and pipeline quality-control discussion (Appendix \ref{app:ablation_linguistic_rounds}), and prompts for those augmentation baselines (Appendix \ref{sec:prompts-augmentation-baselines}). A final subsection records the BFCL category errors from the fine-tuning evaluation in \S\ref{sec:LLM Fine-tuning}.

\subsection{Measuring Diversity - Related work}
\label{app: Measuring Diversity - Related work}
A tangent line of prior work has introduced a range of quantitative metrics for characterizing diversity in generated text, spanning lexical, syntactic, semantic, and structural dimensions. Lexical diversity is commonly measured using type--token ratio (TTR) and its moving-average variant (MATTR), as well as n-gram uniqueness and corpus-level compression ratios, which capture surface variation and redundancy \citep{jarvis2013capturing, li2015diversity, riaz2025metasynth}. Syntactic diversity has been studied using metrics derived from parse structures, including entropy over syntactic trees and average tree edit distance between sentences \citep{sidorov2015computing, moscoso2025measuring}. Semantic diversity is often measured using embedding-based metrics, such as average cosine distance between samples or nearest-neighbor variants, sometimes referred to as Chamfer-style distance scores in recent work \citep{tevet2020evaluating, riaz2025metasynth}. Related work has also considered sentence-level complexity and dataset-level dispersion, using measures such as sentence length, Flesch--Kincaid grade level, model-based complexity judgments, and kernel-based scores like the Vendi score \citep{lorge1939predicting, feng2010comparison, flesch1948new, kincaid1975derivation, trott2024measuring, friedman2022vendi}. 
In our work we select a subset of these metrics to guide our generation process, and reserve the rest to obtain a more fair diversity metric for our data.

\subsection{Qualitative table}\label{app:Qualitative table}
% qualitative table
To illustrate the linguistic diversity achieved by our generation approach, Table~\ref{tbl:salary_query_diversity} presents five distinct queries that all target the same API function (getSalaryBenchmark) with identical parameters (job\_role: "Software Engineer", location: "Melbourne"). Despite this shared ground truth, the queries exhibit substantial variation across multiple dimensions: formality ranges from casual ("Hey, what's the deal with tech salaries in Melb?") to highly professional language; length varies from 10 to 42 words; contextual depth spans from minimal context to rich scenarios embedding the query within specific use cases; and user personas range from casual inquirers to domain experts such as career advisors and city planners. This qualitative example demonstrates how our diversity optimization produces queries that maintain correctness while covering a broad range of natural language variations that an LLM might encounter in real-world applications.
\subsection{Argument diversity}\label{app:Argument values comparison tables}

In this section, we provide detailed comparisons of argument values generated by our method against those from ToolAce and APIGen, as well as a na\"ive baseline that simply samples from an LLM at temperature 1.0. We additionally report scores for an \emph{entity-swapping} augmentation baseline applied independently to each release: starting from each original example, an LLM proposes four variants that replace named entities in the user query and, when the ground-truth call exposes parameters, updates aligned parameter values in the assistant turn; one variant is chosen uniformly at random (see Appendix \ref{sec:prompts-augmentation-baselines} for the prompt specification). These augmented datasets are not rerun through our generation-time judge, so they may admit looser query--call consistency than our pipeline enforces. For each comparison, we first identified argument types that appear at least 20 times in both datasets using our automatic semantic clustering algorithm for parameter grouping (see Appendix \ref{sec:parameter_grouping}). This clustering approach groups parameters based on semantic similarity rather than explicit type annotations, which explains some of the apparent heterogeneity in parameter values -- for instance, a parameter cluster labeled ``city'' may include both city names and zip codes if the underlying APIs accept either format, and ``location'' parameters may encompass both geographic coordinates and city names depending on the API specifications.

Our comparison methodology works as follows: we randomly sampled 20 parameter values from each identified argument type in the baseline dataset (ToolAce or APIGen), extracting not only the parameter values themselves but also the specific API functions they were called with and all other parameter values used in those function calls. We then applied the parameter value generation component of our algorithm to generate new values for the same APIs with all other parameters held fixed, ensuring a controlled comparison where only the target parameter varies. This approach allows us to directly assess whether our diversity optimization produces more varied values than the baseline methods when generating data for identical API contexts.

We evaluate diversity using two complementary metrics: NCD Diversity, which measures diversity based on compression efficiency, and Cluster Entropy, which our generation process explicitly optimizes. For the comparison with ToolAce, we identified six parameter groups: Currency, Username (Instagram), Location, Year, Email, and Username (TikTok). For APIGen, we identified six groups: Domain, Year, Location/Geographic, Currency, Email, and Username (social media). Tables~\ref{tab:toolace parameter_diversity} and~\ref{tbl:apigen parameter_diversity} present quantitative diversity scores for each parameter group, showing both metrics with bootstrap standard deviations computed by repeatedly sampling 80\% of the parameter values. The tables add a column for the entity-swapped release alongside base ToolAce or APIGen, our method, and the na\"ive baseline. The average diversity scores across all parameter groups (reported in Table~\ref{tab:combined_averages} of the main paper, which uses entity-swapped ToolAce and base APIGen together with ours and the na\"ive sampler) demonstrate that our method consistently achieves higher diversity than both base releases and their entity-swapped variants.

We also provide detailed tables showing the actual parameter values for each group. For the ToolAce comparison, Tables~\ref{tbl:toolace city}--\ref{tbl:toolace name_tiktok} present samples across six argument types: city, currency, year, email, Instagram username, and TikTok username. For the APIGen comparison, Tables~\ref{tbl:apigen domain}--\ref{tbl:apigen username} show samples for location/geographic, currency, year, email, domain, and social media username. Each table displays 20 samples for a specific argument type, with three columns comparing values from the baseline dataset (ToolAce or APIGen), our method, and the na\"ive temperature-based sampling approach. The diversity differences are immediately apparent: baseline datasets often exhibit heavy repetition (e.g., ``USD'' appearing 11 times in ToolAce's currency samples, visible in Table~\ref{tbl:toolace currency}), while our method produces more balanced distributions across the value space. The na\"ive baseline typically shows even more severe repetition, confirming that explicit diversity optimization is essential for generating high-quality training data. These concrete examples illustrate how the abstract diversity metrics translate into practical differences in the generated data.
\subsection{Linguistic diversity}\label{app:Linguistic diversity}

Having examined argument diversity, we now turn to linguistic diversity in the queries themselves. This section provides comprehensive metrics for evaluating how varied the natural language requests are across our dataset compared to ToolAce, APIGen, their LLM-paraphrase and entity-swapped augmentations, and the BFCL benchmark. We begin with a brief overview of the diversity metrics used, followed by detailed results tables and a manual correctness evaluation to ensure that increased diversity does not compromise data quality.

\subsubsection{Diversity metrics overview}

There are multiple established approaches to measuring diversity in generated text. Lexical diversity captures vocabulary richness and can be measured through metrics such as Type-Token Ratio (TTR) and moving average TTR (MATTR), which quantify the proportion of unique words to total words, as well as N-gram uniqueness, which measures how many distinct sequences of words appear in the corpus. Simpson's Index, borrowed from biodiversity research, measures the probability that two samples from a population belong to the same species. In our context, we treat the population as all words used in the query corpus, so Simpson's Index quantifies the probability that two words selected uniformly at random from the concatenated corpus will be identical—higher values indicate that most selections yield different words, reflecting greater lexical diversity.

Syntactic diversity examines structural variation in sentences and can be estimated using metrics applied to parse trees, including entropy of tree structure and Tree Edit Distance, which quantifies how many operations are needed to transform one parse tree into another and is averaged across all pairs of sentences to measure overall syntactic variation.

The compression ratio leverages a fundamental principle: diverse data is harder to compress than repetitive data. Compression algorithms eliminate redundancy by encoding repeated patterns once and referencing them multiple times, achieving significant size reduction in corpora with similar elements. A diverse corpus with few repetitions offers little compression opportunity, resulting in output close to the original size. The compression ratio (compressed size / original size) quantifies this: ratios near 1.0 indicate high diversity, while ratios near 0 indicate low diversity with high repetition.

Semantic diversity evaluates meaning-level variation and can be assessed through several complementary metrics. Paraphrase variety measures the average cosine distance between sentence embeddings, with larger distances indicating more semantically distinct queries. The Chamfer Distance Score provides a related measure: for each query, it computes the minimum distance to all other queries, then averages this quantity across the dataset, effectively measuring how isolated each query is in the semantic space. Semantic Spread takes a different approach by embedding all queries, computing the centroid of these embeddings, and then calculating the average distance from each query to the centroid -- similar in spirit to a standard deviation but using the first moment rather than the second, providing a measure of how dispersed the queries are around their semantic center. Cluster Entropy applies the same clustering-based algorithm we use for argument diversity (\Cref{alg:clustering_based_entropy}), but operates on query embeddings rather than parameter values. Specifically, we embed all queries using all-MiniLM-L6-v2, apply DBSCAN clustering with cosine distance and $\epsilon = 0.3$ (rather than the $\epsilon = 0.5$ used for argument values, reflecting the different nature of sentence embeddings), treat DBSCAN outliers as singleton clusters, and compute the entropy of the resulting cluster distribution -- higher entropy indicates that queries are spread across many distinct semantic clusters rather than concentrating in a few. Finally, the Vendi score provides a unified semantic diversity measure by computing the entropy of eigenvalues from a kernel matrix constructed over the dataset embeddings.

Sentence complexity represents another measurable dimension, captured through sentence length and the Flesch-Kincaid Grade Level readability score. High variance in the sentence complexity metrics demonstrates diversity in this dimension.

\subsubsection{Comparative results and analysis}

We compute all diversity metrics for queries sampled from our dataset, the base ToolAce and APIGen releases, four augmented corpora (LLM paraphrase and entity swap applied to each release), and BFCL. Table~\ref{tbl:diversity_scores_toolace_apigen_full} lists every corpus at 1{,}793 queries per column: ours, ToolAce, APIGen, ToolAce (paraphrase), ToolAce (entity swap), APIGen (paraphrase), and APIGen (entity swap). Bold entries mark the highest score in a row when that lead is statistically significant at $\alpha = 0.05$ against \emph{all} other columns under the criterion stated in the table caption. Table~\ref{tbl:diversity_scores_bfcl} compares our method (subsampled to 1{,}240 queries) against the BFCL benchmark. Each table is divided into two sections: the upper section contains metrics that were explicitly optimized during our generation process (TTR, Compression Ratio Diversity, Var(FKGL), Parse Tree Entropy, Var(Length), Paraphrase Variety, Chamfer Distance Score, and Vendi Score), while the lower section presents metrics that were not optimized (Simpson's Index, Tree Edit Distance, Cluster Entropy, and Semantic Spread). These non-optimized metrics provide the strongest evidence that our diversity improvements generalize beyond the specific objectives we targeted -- it is unsurprising that we excel on optimized metrics, but achieving superior diversity on independent metrics demonstrates that our approach produces broadly diverse data rather than simply overfitting to particular measures. Bootstrap standard deviations are computed from 100 subsamples of 80\% of the data.

Across the seven-way ToolAce/APIGen comparison, our dataset achieves the highest scores on most lexical, syntactic, and semantic metrics, often with statistically significant margins. The paraphrased ToolAce variant attains slightly higher Parse Tree Entropy and Cluster Entropy than ours; ToolAce (base or augmented) remains highest on Var(Length), again driven by rare very long requests rather than broad length diversity. The BFCL comparison follows the same qualitative pattern as in the main paper. Augmented corpora are not filtered by our dataset-level judge, so diversity advantages for our method are conservative relative to those baselines.

\subsubsection{Ablations and pipeline quality control}\label{app:ablation_linguistic_rounds}

Two complementary aspects of our pipeline most strongly shape intrinsic query diversity in our setup: \emph{quality-control filtering} (keeping examples trainable and consistent with their calls) and \emph{iterative diversity optimization} for user requests.

\smallskip\noindent\textbf{Filtering.} We added validation and filtering stages incrementally while scaling generation, after observing that non-trivial fractions of raw outputs at individual stages were unusable (order-of-magnitude, roughly 20\%--50\% depending on the stage and API family) until judges and filters were tightened. Concrete mechanisms include per-candidate LLM judging during query generation (Appendix \ref{sec:query_generation}), plausibility and retrieval-based filtering when attaching distractor APIs (Appendix \ref{sec:distractors_retrieval}), and analogous checks elsewhere in the detailed pipeline (Appendix \ref{sec:pipeline_qc_filtering}). We do not report a full counterfactual ablation that disables these filters, since that setting yields mostly invalid or misleading training pairs; instead, we document them as engineering components motivated by measured failure rates during development.

\smallskip\noindent\textbf{Multi-round request diversity.} The greedy multi-round procedure in \S\ref{sec:diverse gen alg} is applied to \emph{request} generation with the fused linguistic metrics described in \S\ref{sec:Our Methods} and detailed in Appendix \ref{sec:query_generation}. Parameter-value generation, by contrast, uses a single diversity round for cost reasons (\S\ref{sec: method single func}); Table~\ref{tbl:ablation_linguistic_diversity_multi_vs_single_round} therefore isolates the effect of iterative rounds on \emph{queries}, not on argument sampling. We compare (i) a stratified subsample of 90 queries from our full corpus, generated with our standard multi-round configuration, against (ii) a freshly generated corpus of 90 queries produced with the same pipeline except that request diversity optimization uses a single round. We evaluate the same lexical, syntactic, and semantic metrics as in \Cref{tbl:diversity_scores_toolace_apigen_full}, with bootstrap standard deviations from 100 subsamples of 80\% of the data and the same pairwise significance rule as in that table's caption. Multi-round optimization improves most metrics that are actively optimized during generation and several held-out metrics at $\alpha = 0.05$; compression ratio diversity, Simpson's index, and tree edit distance are not significantly different between the two columns at that level.

% Linguistic diversity ablation: multi-round vs.\ one-round request generation.
\begin{table}[t]
\centering
\small
\begin{tabular}{@{}lccccc@{}}
\toprule
& & \multicolumn{2}{c}{\textbf{Our Method (Multi-Round)}} & \multicolumn{2}{c}{\textbf{Ablation (1-Round)}} \\
\cmidrule(lr){3-4} \cmidrule(lr){5-6}
\textbf{Metric} & \textbf{Type} & \textbf{Score} & \textbf{Std} & \textbf{Score} & \textbf{Std} \\
\midrule
TTR & Lexical & \textbf{0.4699} & 0.0050 & 0.4506 & 0.0061 \\
Compression Ratio Diversity & Lexical & 0.4440 & 0.0031 & 0.4429 & 0.0043 \\
Var(FKGL) & Lex./Syn. & \textbf{23.7097} & 1.5565 & 17.8648 & 1.1375 \\
Parse Tree Entropy & Syntactic & \textbf{6.4919} & 0.0000 & 6.3640 & 0.0357 \\
Var(Length) & Syntactic & \textbf{170.9111} & 16.4555 & 94.5344 & 6.8131 \\
Paraphrase Variety (Avg Sim) & Semantic & \textbf{0.9396} & 0.0023 & 0.9273 & 0.0026 \\
Chamfer Distance Score & Semantic & \textbf{0.6700} & 0.0068 & 0.5898 & 0.0155 \\
Vendi Score & Semantic & \textbf{62.8876} & 0.3994 & 56.1126 & 0.9509 \\
\midrule
Simpson's Index & Lexical & 0.9892 & 0.0003 & 0.9884 & 0.0003 \\
Tree Edit Distance & Syntactic & 22.7710 & 0.5807 & 21.6350 & 0.4741 \\
Cluster Entropy & Semantic & \textbf{6.4919} & 0.0000 & 6.3640 & 0.0357 \\
Semantic Spread & Semantic & \textbf{0.9639} & 0.0012 & 0.9576 & 0.0014 \\
\midrule
Total Queries & -- & 90 & -- & 90 & -- \\
\bottomrule
\end{tabular}
\caption{Ablation study: effect of multi-round vs.\ single-round query generation. Both settings contain 90 queries (stratified subsample from our full dataset vs.\ 1-round generation ablation). Bootstrap standard deviations from 100 subsamples of 80\% of data. \textbf{Bold} scores mark the higher value in that row when it is statistically significant at $\alpha = 0.05$ versus the other column under $|\mu_1 - \mu_2| > 1.96\sqrt{\sigma_1^2 + \sigma_2^2}$. Higher values indicate greater diversity for all metrics.}
\label{tbl:ablation_linguistic_diversity_multi_vs_single_round}
\end{table}

\subsubsection{Manual evaluation}\label{app:Manual evaluation}
\begin{figure}[ht]
\centering
\includegraphics[width=0.8\textwidth]{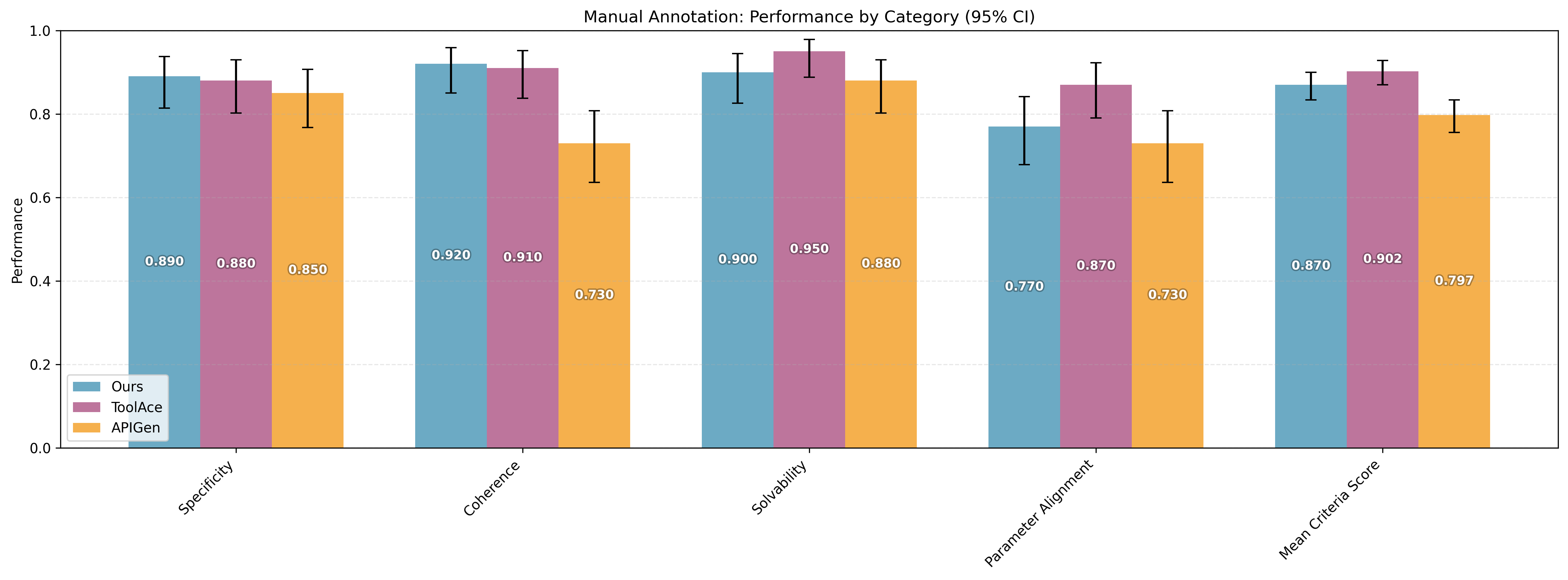}
\caption{Manual quality annotations on four metrics and their mean, for 100 randomly sampled examples from each of three corpora: the base ToolAce and APIGen releases, and our dataset (augmented variants excluded). Error bars show 95\% confidence intervals.}
\label{fig:dataset_quality_metrics}
\end{figure}

To verify that our increased diversity does not introduce noise or compromise data quality, we conducted a manual evaluation following the methodology of \citet{iskander2024quality}. We randomly sampled 100 examples from each of our dataset, ToolAce, and APIGen (the base releases only) and had annotators evaluate each query across four quality dimensions: specificity, coherence, solvability, and parameter alignment.
\textbf{Specificity} measures whether an instruction contains all necessary details and parameter values required for an LLM to fulfill the user's request without requiring additional information.
\textbf{Coherence} assesses whether multiple requests within an instruction are logically connected and relevant to each other, forming a sensible flow that reflects real-world use cases.
\textbf{Solvability} evaluates whether the available API tools possess the functionality needed to address the requests, regardless of whether specific parameter values are explicitly provided in the instruction.
\textbf{Parameter Alignment} determines whether all parameter values in an API-call sequence are correctly extracted or inferred from the instruction, with no missing or hallucinated values.
Figure~\ref{fig:dataset_quality_metrics} presents the four metrics and their mean. Our mean is $0.870$ (95\% CI $[0.833, 0.899]$), compared with $0.902$ ($[0.869, 0.928]$) for ToolAce and $0.797$ ($[0.755, 0.834]$) for APIGen. Our interval overlaps ToolAce's, and the point estimate is higher than APIGen's. The separate metrics follow the same pattern: our dataset stays close to ToolAce and ahead of APIGen. These results indicate that our diversity optimization does not come at the cost of data quality: the generated queries remain correct, useful, and suitable for training while spanning a broader range of linguistic and semantic variations. The consistently high quality scores across these three corpora indicate that the observed differences in diversity metrics are not attributable to malformed or low-quality generations. We did not annotate the LLM-augmented ToolAce/APIGen variants; those corpora are retained for intrinsic diversity only and are not held to the same filtering pipeline as our data, matching the asymmetric comparison discussed in \S\ref{sec:Analysis}.
\onecolumn
{\small
\begin{longtable}{@{}cL{4.5cm}cccL{2cm}@{}}
\toprule
\textbf{\#} & \textbf{Query} & \textbf{Length} & \textbf{Formality} & \textbf{Context} & \textbf{User Persona} \\
\midrule
\endfirsthead
\toprule
\textbf{\#} & \textbf{Query} & \textbf{Length} & \textbf{Formality} & \textbf{Context} & \textbf{User Persona} \\
\midrule
\endhead
1 & What's the average salary for a Software Engineer in Melbourne? & 10 & Neutral & Minimal & Job seeker \\
\midrule
2 & Hey, I'm thinking about moving to Melbourne for a Software Engineering job. What's the general sentiment among current employees about the job market and salary expectations there? & 27 & Casual & Moderate & Relocating professional \\
\midrule
3 & As a career advisor, I'm helping a client transition from a Sales Manager role to a Software Engineer position in Melbourne. Could you provide the salary benchmark for this new role and how it might impact their overall compensation package? & 40 & Professional & High & Career advisor \\
\midrule
4 & As a city planner evaluating the economic impact of tech industry growth on urban development, I need the salary benchmark for Software Engineers in Melbourne to assess the potential for attracting tech talent and the subsequent effects on local real estate markets. & 42 & Professional & High & City planner \\
\midrule
5 & Hey, what's the deal with tech salaries in Melb? Any insights on what Software Engineers typically earn? & 17 & Casual & Minimal & Casual inquirer \\
\bottomrule
\caption{Five queries targeting a single API (getSalaryBenchmark) with same parameters (job\_role: ``Software Engineer'', location: ``Melbourne''). Despite targeting the same function and parameters, the queries exhibit substantial variation across formality, length (10-42 words), contextual depth, and user personas.}
\label{tbl:salary_query_diversity}
\end{longtable}}

\begin{longtable}{@{}lL{3.5cm}L{3.5cm}L{3.5cm}@{}}
\toprule
\textbf{Sample} & \textbf{ToolAce Value} & \textbf{Ours} & \textbf{na\"ive T=1.0} \\
\midrule
\endfirsthead
\toprule
\textbf{Sample} & \textbf{ToolAce Value} & \textbf{Ours} & \textbf{na\"ive T=1.0} \\
\midrule
\endhead
1  & Los Angeles   & Bangkok, Thailand        & New York City       \\
2  & London        & Milan                    & London              \\
3  & Boston        & Mumbai, India            & New York City       \\
4  & New York      & Kuala Lumpur, Malaysia   & New York            \\
5  & Los Angeles   & Nairobi, Kenya           & Los Angeles         \\
6  & Monaco        & Prague, Czech Republic   & New York            \\
7  & Reykjavik     & Vancouver, Canada        & San Francisco, CA   \\
8  & New York City & Moscow, Russia           & New York City, USA  \\
9  & Cleveland     & Lagos, Nigeria           & Los Angeles         \\
10 & Los Angeles   & Cairo, Egypt             & Los Angeles         \\
11 & Los Angeles   & Toronto, Canada          & New York City       \\
12 & New York      & 90210                    & San Francisco, CA   \\
13 & Sydney        & Buenos Aires, Argentina  & New York, USA       \\
14 & Miami         & Reykjavik, Iceland       & New York City       \\
15 & San Francisco & Jakarta, Indonesia       & New York, USA       \\
16 & Paris         & Sofia, Bulgaria          & New York City       \\
17 & Monaco        & Baku, Azerbaijan         & Monaco              \\
18 & Helsinki      & Seoul, South Korea       & San Francisco, CA   \\
19 & New York      & Cape Town, South Africa  & New York, USA       \\
20 & Tokyo         & Tokyo, Japan             & San Francisco, CA, USA \\
\bottomrule
\caption{Parameter values comparison - City}
\label{tbl:toolace city}
\end{longtable}

% currncy word cloud
%\input{word_clouds}

% Currency table - ToolAce
\begin{longtable}{@{}lL{3.5cm}L{3.5cm}L{3.5cm}@{}}
\toprule
\textbf{Sample} & \textbf{ToolAce Value} & \textbf{Ours} & \textbf{na\"ive T=1.0} \\
\midrule
\endfirsthead
\toprule
\textbf{Sample} & \textbf{ToolAce Value} & \textbf{Ours} & \textbf{na\"ive T=1.0} \\
\midrule
\endhead
1  & BTC        & JPY & BTC     \\
2  & USD        & NOK & USD     \\
3  & btc        & ETH & btc     \\
4  & GBP        & HKD & USD     \\
5  & USD        & RUB & USD     \\
6  & EUR        & AUD & USD     \\
7  & JPY        & SGD & USD     \\
8  & USD        & GBP & USD     \\
9  & USD        & THB & USD     \\
10 & AUD to CAD & MXN & USD     \\
11 & USD        & XRP & USD     \\
12 & USD        & PHP & USD     \\
13 & INR to CHF & BCH & USD     \\
14 & USD        & ILS & USD     \\
15 & GBP/USD    & USD/JPY & EUR/USD \\
16 & Bitcoin    & XMR & bitcoin \\
17 & USD        & COP & USD     \\
18 & USD        & CHF & USD     \\
19 & USD        & AED & USD     \\
20 & JPY        & BGN & USD     \\
\bottomrule
\caption{Parameter values comparison - Currency.}
\label{tbl:toolace currency}
\end{longtable}

% year table - ToolAce
\begin{longtable}{@{}lL{3.5cm}L{3.5cm}L{3.5cm}@{}}
\toprule
\textbf{Sample} & \textbf{ToolAce Value} & \textbf{Ours} & \textbf{na\"ive T=1.0} \\
\midrule
\endfirsthead
\toprule
\textbf{Sample} & \textbf{ToolAce Value} & \textbf{Ours} & \textbf{na\"ive T=1.0} \\
\midrule
\endhead
1  & 2019 & 1982 & 2020 \\
2  & 2020 & 2022 & 2020 \\
3  & 1971 & 1971 & 2020 \\
4  & 2020 & 1970 & 2000 \\
5  & 2025 & 2015 & 2023 \\
6  & 2022 & 2013 & 2023 \\
7  & 2025 & 2025 & 2021 \\
8  & 2022 & 2016 & 2020 \\
9  & 2026 & 2031 & 2023 \\
10 & 2018 & 1993 & 2020 \\
11 & 2020 & 2007 & 2020 \\
12 & 2016 & 2035 & 2021 \\
13 & 2023 & 2014 & 2020 \\
14 & 2023 & 1985 & 2020 \\
15 & 2020 & 1975 & 2020 \\
16 & 2020 & 2021 & 2020 \\
17 & 2023 & 2082 & 2020 \\
18 & 2017 & 1945 & 2000 \\
19 & 2024 & 1980 & 2020 \\
20 & 2010 & 2045 & 2020 \\
\bottomrule
\caption{Parameter values comparison - Year.}\label{tbl:toolace year}
\end{longtable}

% email adresses table - ToolAce
\begin{longtable}{@{}lL{3.5cm}L{3.5cm}L{3.5cm}@{}}
\toprule
\textbf{Sample} & \textbf{ToolAce Value} & \textbf{Ours} & \textbf{na\"ive T=1.0} \\
\midrule
\endfirsthead
\toprule
\textbf{Sample} & \textbf{ToolAce Value} & \textbf{Ours} & \textbf{na\"ive T=1.0} \\
\midrule
\endhead
1  & \url{example@email.com} & \url{nat.romanoff@sample.cloud} & \url{user@example.com} \\
2  & \url{Jane.Smith@example.com} & \url{xavier.platinum@insanemail.com} & \url{recipient@example.com} \\
3  & \url{bob@example.com} & \url{tina.bronze@covert.dk} & \url{test@example.com} \\
4  & \url{john.doe@example.com} & \url{emma.wilson@cloud.de} & \url{test@example.com} \\
5  & \url{test@tempmail.com} & \url{mike.davis@site.biz} & \url{test@example.com} \\
6  & \url{johndoe2027@example.com} & \url{david.miller1978@hotmail.com} & \url{user123@example.com} \\
7  & \url{myawesomeblog.com} & \url{quickmail.io} & \url{example.com} \\
8  & \url{info@phishingsite1.com} & \url{emily.wilson@demo.co} & \url{test.user@example.com} \\
9  & \url{jane.doe@mail.com} & \url{jack.frost@winter.land} & \url{test@example.com} \\
10 & \url{john.doe@examplecompany.com} & \url{george.white@info.co.nz} & \url{user@example.com} \\
11 & \url{alice.johnson@example.com} & \url{ursula.brown@dev.land} & \url{newmember@example.com} \\
12 & \url{john@example.com} & \url{oliver.garcia@trial.biz} & \url{user123@example.com} \\
13 & \url{john.doe@example.com} & \url{henry.gray@sample.net} & \url{test@example.com} \\
14 & \url{sales@company.com} & \url{xavier.platinum@protectedmail.info} & \url{john.doe@example.com} \\
15 & \url{john.doe@example.com} & \url{ivan.ivanov123@sample.cloud} & \url{user123@example.com} \\
16 & \url{user@example.com} & \url{alex.johnson@test.net} & \url{user@example.com} \\
17 & A1234 & \url{amelia.hall456@alias.tv} & alias12345 \\
18 & \url{jane.doe@email.com} & \url{john.doe@example.com} & \url{test@example.com} \\
19 & \url{charlie@example.org} & \url{luna.eclipse@stellarvoyage.space} & \url{test@example.com} \\
20 & \url{John.Doe@example.com} & \url{john.constantine@hellblazer.com} & \url{recipient@example.com} \\
\bottomrule
\caption{Parameter values comparison - Email.}\label{tbl:toolace email}
\end{longtable}

% Name (Instagram) table - ToolAce
\begin{longtable}{@{}lL{3.5cm}L{3.5cm}L{3.5cm}@{}}
\toprule
\textbf{Sample} & \textbf{ToolAce Value} & \textbf{Ours} & \textbf{na\"ive T=1.0} \\
\midrule
\endfirsthead
\toprule
\textbf{Sample} & \textbf{ToolAce Value} & \textbf{Ours} & \textbf{na\"ive T=1.0} \\
\midrule
\endhead
1  & travelwithme & DanceDiva\_321 & example\_user \\
2  & exampleuser & NatureNerd\_44 & example\_user \\
3  & therock & NatureEnthusiast\_54 & example\_user \\
4  & lonelyplanet & TechSavvy\_765 & example\_instagram\_prostring \\
5  & insta1234 & WineConnoisseur\_890 & 123456789 \\
6  & financeguru456 & CryptoCrusader\_789 & exampleUser123 \\
7  & wonderful\_places & Explorer\_321 & example\_user \\
8  & traveltheglobe & GamerGal\_987 & example\_user \\
9  & 123456789 & Explorer\_012 & 123456789 \\
10 & financeguru456 & Engineer\_987 & exampleUser123 \\
11 & voyaged & SportsFanatic\_123 & example\_instagram\_prostring \\
12 & FashionForYou & EcoWarrior\_321 & instagram\_handle\_123 \\
13 & awesomecontentcreator & Dancer\_456 & instagram\_user \\
14 & techenthusiast123 & AnimeLover\_456 & exampleUser123 \\
15 & TechEnthusiast & CryptoTrader\_012 & john\_doe \\
16 & adventure\_nick & Adventurer\_123 & john\_doe \\
17 & instaAcc123 & FitnessGuru\_123 & example\_user \\
18 & https://instagram.com/\allowbreak user\_prof\_one/ & https://www.instagram.com/\allowbreak stories/PetLover\_789/\allowbreak 7766554433/ & https://www.instagram.com/\allowbreak stories/exampleuser/\allowbreak 1234567890 \\
19 & ninh.duong.lan.ngoc & MovieNight\_456 & instagram\_user\_123 \\
20 & instausername & NatureLover\_987 & instagram\_official \\
\bottomrule
\caption{Parameter values comparison - Name (Instagram).}
\label{tbl:toolace name_insta}
\end{longtable}

% Name (TikTok) table - ToolAce
\begin{longtable}{@{}lL{3.5cm}L{3.5cm}L{3.5cm}@{}}
\toprule
\textbf{Sample} & \textbf{ToolAce Value} & \textbf{Ours} & \textbf{na\"ive T=1.0} \\
\midrule
\endfirsthead
\toprule
\textbf{Sample} & \textbf{ToolAce Value} & \textbf{Ours} & \textbf{na\"ive T=1.0} \\
\midrule
\endhead
1  & alice\_jones & 9nO0pQ1r & newUser123 \\
2  & @TheGadgetGuy & GamerGal\_99 & exampleUser123 \\
3  & John Doe & QuantumLeap\_2020 & john\_doe \\
4  & JennyDoe & FitnessFreak\_2023 & john\_doe \\
5  & market\_research2023 & PetLoverPat\_2019 & john\_doe\_123 \\
6  & johndoe & CamperCindy\_2045 & john\_doe \\
7  & KitchenVibes2024 & ArtisanAlice\_2018 & john\_doe \\
8  & johndoe & GreenThumb\_2029 & john\_doe \\
9  & johnDoe45 & ScientistSam\_2034 & john\_doe \\
10 & JohnDoe & YogaYogi\_2033 & john\_doe \\
11 & john\_doe & EntrepreneurEmma\_2041 & john\_doe \\
12 & jane\_smith & MusicianMike\_2026 & john\_doe \\
13 & cooldance23 & HikingHank\_2027 & john\_doe \\
14 & johnSmith & GamerGuy\_2022 & john\_doe \\
15 & StyleGuru & HistorianHenry\_2028 & exampleUser123 \\
16 & ArtisticSoul123 & ArchitectAlice\_2034 & john\_doe \\
17 & john\_doe & MountaineerMia\_2037 & john\_doe \\
18 & study\_buddy21 & CoderCarl\_2027 & john\_doe \\
19 & JohnDoe & InventorIvy\_2029 & john\_doe \\
20 & john\_doe & DesignerDana\_2032 & john\_doe \\
\bottomrule
\caption{Parameter values comparison - Name (TikTok).}
\label{tbl:toolace name_tiktok}
\end{longtable}

{\small
\begin{longtable}{@{}llcccc@{}}
\toprule
\textbf{Parameter Group} & \textbf{Metric} & \textbf{ToolAce} & \textbf{ToolAce (Ent.\ swap)} & \textbf{Ours} & \textbf{Temp 1.0} \\
\midrule
\endfirsthead
\toprule
\textbf{Parameter Group} & \textbf{Metric} & \textbf{ToolAce} & \textbf{ToolAce (Ent.\ swap)} & \textbf{Ours} & \textbf{Temp 1.0} \\
\midrule
\endhead
\multirow{2}{*}{\texttt{Currency}}
& NCD Diversity & 0.357 (0.018) & 0.357 (0.017) & 0.292 (0.010) & 0.305 (0.011) \\
& Cluster Entropy & 2.461 (0.217) & 2.319 (0.233) & \textbf{\underline{4.322 (0.000)}}$^{\dagger}$ & 1.022 (0.207) \\
\midrule
\multirow{2}{*}{Username (Instagram)}
& NCD Diversity & 0.648 (0.009) & 0.648 (0.010) & 0.646 (0.014) & 0.560 (0.022) \\
& Cluster Entropy & 4.222 (0.062) & 4.022 (0.086) & \underline{4.322 (0.000)}$^{\dagger}$ & 3.009 (0.160) \\
\midrule
\multirow{2}{*}{Location}
& NCD Diversity & 0.516 (0.008) & 0.535 (0.014) & \textbf{0.679 (0.007)} & 0.537 (0.019) \\
& Cluster Entropy & 3.422 (0.159) & 3.304 (0.175) & \textbf{\underline{4.322 (0.000)}}$^{\dagger}$ & 1.781 (0.155) \\
\midrule
\multirow{2}{*}{Year}
& NCD Diversity & 0.269 (0.013) & 0.305 (0.008) & \textbf{0.333 (0.000)} & 0.127 (0.018) \\
& Cluster Entropy & 3.304 (0.139) & 3.584 (0.111) & \textbf{\underline{4.322 (0.000)}}$^{\dagger}$ & 1.479 (0.155) \\
\midrule
\multirow{2}{*}{Email}
& NCD Diversity & 0.564 (0.015) & 0.602 (0.013) & \textbf{0.724 (0.003)} & 0.428 (0.015) \\
& Cluster Entropy & 3.446 (0.194) & 3.722 (0.129) & \textbf{\underline{4.322 (0.000)}}$^{\dagger}$ & 2.764 (0.169) \\
\midrule
\multirow{2}{*}{Username (TikTok)}
& NCD Diversity & 0.544 (0.020) & 0.544 (0.022) & \textbf{0.616 (0.007)} & 0.272 (0.038) \\
& Cluster Entropy & 3.522 (0.158) & 3.684 (0.149) & \textbf{\underline{4.322 (0.000)}}$^{\dagger}$ & 1.022 (0.235) \\
\midrule
\midrule
\multirow{2}{*}{\textbf{Average}}
& NCD Diversity & 0.483 (0.014) & 0.499 (0.014) & \textbf{0.548 (0.007)} & 0.371 (0.020) \\
& Cluster Entropy & 3.396 (0.155) & 3.439 (0.147) & \textbf{\underline{4.322 (0.000)}}$^{\dagger}$ & 1.846 (0.180) \\
\bottomrule
\caption{Comparison of parameter diversity across four generation methods for six parameter groups identified by \Cref{alg:param_grouping}. For each parameter group and diversity metric, we report the mean and bootstrap standard deviation (in parentheses), computed over 20 parameter values. Bootstrap standard deviations are calculated by repeatedly sampling 80\% of the parameter values uniformly at random without replacement and computing the metric on each sample. \textbf{Bold values} indicate statistically significant superiority over all three other methods at the 95\% confidence level, determined by the criterion $|\mu_1 - \mu_2| > 1.96\sqrt{\sigma_1^2 + \sigma_2^2}$ for all pairwise comparisons. Underlining with a dagger ($^{\dagger}$) marks the theoretical maximum for the metric: for Cluster Entropy with 20 parameters, the maximum value is $\log_2(20) = 4.322$.}
\label{tab:toolace parameter_diversity}
\end{longtable}}

% \subsubsection{Our approach vs. APIGen}\label{app:argument diversity vs. APIGen}

\begin{longtable}{@{}lL{3.5cm}L{3.5cm}L{3.5cm}@{}}
\toprule
\textbf{Sample} & \textbf{APIGen Value} & \textbf{Ours} & \textbf{na\"ive T=1.0} \\
\midrule
\endfirsthead
\toprule
\textbf{Sample} & \textbf{APIGen Value} & \textbf{Ours} & \textbf{na\"ive T=1.0} \\
\midrule
\endhead
1  & \url{example.com},\url{test.net},\url{demo.org} & \url{beauty.salon},\url{cosmetic.shop},\url{groom.studio} & \url{example.com},\url{test.org},\url{sample.net} \\
2  & \url{wikipedia.org} & \url{fashion.style} & \url{example.com} \\
3  & \url{example.com} & \url{pet.care} & \url{example.com} \\
4  & \url{amazon.com} & \url{photography.pics} & \url{example.com} \\
5  & \url{wikipedia.org} & \url{startupsuccess.venture} & \url{amazon.com} \\
6  & \url{example4.com} & \url{softwaredevelopment.dev} & \url{example.com} \\
7  & \url{google.com} & \url{blockchaintechnology.crypto} & \url{example.com} \\
8  & \url{baidu.com} & \url{fooddelivery.app} & \url{example.com} \\
9  & \url{amazon.com} & \url{ethicalfashion.boutique} & \url{amazon.com} \\
10 & com & \url{amazon.imaginary} & com \\
11 & \url{amazon.com} & \url{socialmediaplatform.network} & \url{example.com} \\
12 & \url{twitter.com} & \url{sportsnews.arena} & \url{example.com} \\
13 & \url{twitter.com} & \url{techinnovations.com} & \url{example.com} \\
14 & \url{google.com} & \url{artificialintelligence.lab} & \url{example.com} \\
15 & \url{amazon.com} & \url{smartcity.urban} & \url{example.com} \\
16 & \url{freemail.com} & \url{realty.estate} & \url{example.com} \\
17 & tech & circulareconomy & newbusiness \\
18 & \url{outlook.com} & \url{fleeting.org} & \url{example.com} \\
19 & \url{example.com} & \url{traveladventures.tours} & \url{example.com} \\
20 & \url{domain5.com} & \url{techinnovations.io},\url{healthandwellness.org},\url{edutech.academy},\url{entertainmenthub.tv},\url{greenenergy.eco},\url{digitalmarketing.co},\url{fintechsolutions.net},\url{travelandadventure.tour},\url{fashionforward.style},\url{realestatepros.property},\url{foodandbeverage.gourmet},\url{sportsandfitness.club},\url{musicandarts.culture},\url{gamingworld.play},\url{automotiveinsights.drive},\url{beautyandspa.relax},\url{petlovers.care},\url{kidsandfamily.fun},\url{scienceandtech.lab},\url{agricultureandfarming.grow},\url{legalandlaw.justice},\url{constructionandbuild.craft},\url{photographyandart.capture},\url{socialmediatrends.network},\url{nonprofitandcharity.give} & \url{example.com},\url{test.org},\url{sample.net} \\
\bottomrule
\caption{Parameter values comparison - Domain.}
\label{tbl:apigen domain}
\end{longtable}

% year table - APIGen
\begin{longtable}{@{}lL{3.5cm}L{3.5cm}L{3.5cm}@{}}
\toprule
\textbf{Sample} & \textbf{APIGen Value} & \textbf{Ours} & \textbf{na\"ive T=1.0} \\
\midrule
\endfirsthead
\toprule
\textbf{Sample} & \textbf{APIGen Value} & \textbf{Ours} & \textbf{na\"ive T=1.0} \\
\midrule
\endhead
1  & 2024 & 2040 & 2024 \\
2  & 2023 & 2022 & 2023 \\
3  & 25 & 75 & 5 \\
4  & 100000 & 140 & 10 \\
5  & 25 & 160 & 5 \\
6  & 2020 & 1998 & 2021 \\
7  & 2023 & 2035 & 2023 \\
8  & 1994 & 2015 & 2020 \\
9  & 15 & 60 & 10 \\
10 & 2021 & 1974 & 2019 \\
11 & 2010 & 1970 & 2020 \\
12 & 10 & 100 & 10 \\
13 & 10 & 40 & 10 \\
14 & 1000 & 130 & 10 \\
15 & 2 & 92 & 1 \\
16 & 2017 & 1904 & 2024 \\
17 & 5 & 30 & 5 \\
18 & 15 & 110 & 5 \\
19 & 2023 & 2024 & 2023 \\
20 & 2009 & 1988 & 2024 \\
\bottomrule
\caption{Parameter values comparison - Year.}
\label{tbl:apigen year}
\end{longtable}

% location table - APIGen
\begin{longtable}{@{}lL{3.5cm}L{3.5cm}L{3.5cm}@{}}
\toprule
\textbf{Sample} & \textbf{APIGen Value} & \textbf{Ours} & \textbf{na\"ive T=1.0} \\
\midrule
\endfirsthead
\toprule
\textbf{Sample} & \textbf{APIGen Value} & \textbf{Ours} & \textbf{na\"ive T=1.0} \\
\midrule
\endhead
1  & San Francisco & Moscow, Russia & San Francisco \\
2  & Paris & Jakarta, Indonesia & San Francisco \\
3  & 40.7128 & 45.5017 & 37.7749 \\
4  & 47.6062 & -34.6037 & 40.7128 \\
5  & -33.8651 & -34.5678 & 37.7749 \\
6  & remote & Borås & Stockholm \\
7  & Tokyo & Dubai, UAE & New York \\
8  & Australia & Eurasia & America \\
9  & Tokyo & New York, USA & New York \\
10 & 38.8977 & 7.4321 & 5.6037 \\
11 & Touba & Douala & Central \\
12 & Berlin, Germany & Tokyo, Japan & Paris \\
13 & 9.0820 & 5.2345 & 5.6037 \\
14 & Extrême-Nord & Kribi & Central \\
15 & Tokyo & Lima, Peru & New York \\
16 & 36.1069 & 55.7558 & 37.7749 \\
17 & New York City & Oslo, Norway & New York \\
18 & Abu Dhabi & Fuji, Japan & Monaco \\
19 & 37.369268, -122.038116 & 41.8781,-87.6298 & 37.7749,-122.4194 \\
20 & Chicago & Gothenburg & Stockholm \\
\bottomrule
\caption{Parameter values comparison - Location / Geographic.}
\label{tbl:apigen location}
\end{longtable}

% currency table - APIGen
\begin{longtable}{@{}lL{3.5cm}L{3.5cm}L{3.5cm}@{}}
\toprule
\textbf{Sample} & \textbf{APIGen Value} & \textbf{Ours} & \textbf{na\"ive T=1.0} \\
\midrule
\endfirsthead
\toprule
\textbf{Sample} & \textbf{APIGen Value} & \textbf{Ours} & \textbf{na\"ive T=1.0} \\
\midrule
\endhead
1  & XAU & NOK & USD \\
2  & EUR & CNY & EUR \\
3  & ETH & UNI & BTC \\
4  & USD & NZD & USD \\
5  & 2022-03-15 & 2000-12-31 & 2023-10-01 \\
6  & USD & EUR & USD \\
7  & EUR & TRY & EUR \\
8  & Litecoin & XRP & bitcoin \\
9  & GBP & HKD & USD \\
10 & Binance & OKEx & bitcoin \\
11 & GBP & IDR & USD \\
12 & JPY & COP & EUR \\
13 & h4 & h1 & h1 \\
14 & stable & Aave & Bitcoin \\
15 & USD & ETH & USDT \\
16 & USD & SEK & USD \\
17 & EUR & GBP & USD \\
18 & EUR & BRL & USD \\
19 & JPY & SGD & USD \\
20 & 2023-02-28 & 2030-04-01 & 2023-02-01 \\
\bottomrule
\caption{Parameter values comparison - Currency.}
\label{tbl:apigen currency}
\end{longtable}

% email table -APIGen
\begin{longtable}{@{}lL{3.5cm}L{3.5cm}L{3.5cm}@{}}
\toprule
\textbf{Sample} & \textbf{APIGen Value} & \textbf{Ours} & \textbf{na\"ive T=1.0} \\
\midrule
\endfirsthead
\toprule
\textbf{Sample} & \textbf{APIGen Value} & \textbf{Ours} & \textbf{na\"ive T=1.0} \\
\midrule
\endhead
1  & \url{info@techblog.com} & \url{ruby.bronze@freemail.hu} & \url{test@example.com} \\
2  & \url{john.doe@gmail.com} & \url{victor.frank@mail.se} & \url{test@example.com} \\
3  & \url{john@example.com} & \url{warren.buffett@gmx.com} & \url{example@domain.com} \\
4  & \url{admin@domain} & \url{isabella.lewis@posteo.com} & \url{example@domain.com} \\
5  & \url{contact@company.io.} & \url{joseph.garcia345@mail.ru} & \url{example@domain.com} \\
6  & \url{jane.smith@testmail.com} & \url{mary.white@fastmail.com} & \url{test@example.com} \\
7  & \url{john.doe@example.com} & \url{thomas.lee@gmx.net} & \url{example@example.com} \\
8  & \url{jane.smith@example.org} & \url{george.taylor@privatemail.com} & \url{test.user@example.com} \\
9  & \url{personal3@outlook.com} & \url{nancy.gray@onionmail.org} & \url{test@example.com} \\
10 & \url{https://www.mywebsite.com} & \url{http://profile.example.jp/users} & \url{http://example.com} \\
11 & \url{email@.com} & \url{alex.brown456@test.net} & \url{example@domain.com} \\
12 & \url{user@user.com} & \url{wanda.gonzalez@hushmail.com} & \url{example@example.com} \\
13 & \url{jane_doe@sub.domain.net} & \url{judy.green@disroot.org} & \url{example@example.com} \\
14 & \url{info@company.com} & \url{charles.white@fastmail.com} & \url{example@example.com} \\
15 & \url{jane.doe@temporarymail.com} & \url{karen.blue@mailcatch.com} & \url{test@example.com} \\
16 & \url{info@my-company.io} & \url{tom.lewis@tutanota.com} & \url{test.user@example.com} \\
17 & \url{marketing@example.eu} & \url{kevin.orange@mail.com} & \url{test@example.com} \\
18 & \url{testuser123@hotmail.co.uk} & \url{ron.robinson@rocketmail.com} & \url{example@example.com} \\
19 & \url{jane_doe@domain.co} & \url{roger.flores345@trial.pt} & \url{example@domain.com} \\
20 & \url{user2@temporary.com} & \url{wanda.maximoff@scarletmail.com} & \url{test@example.com} \\
\bottomrule
\caption{Parameter values comparison - Email.}
\label{tbl:apigen email}
\end{longtable}

% username table - APIGen
\begin{longtable}{@{}lL{3.5cm}L{3.5cm}L{3.5cm}@{}}
\toprule
\textbf{Sample} & \textbf{APIGen Value} & \textbf{Ours} & \textbf{na\"ive T=1.0} \\
\midrule
\endfirsthead
\toprule
\textbf{Sample} & \textbf{APIGen Value} & \textbf{Ours} & \textbf{na\"ive T=1.0} \\
\midrule
\endhead
1  & dave & ironMan & john\_doe \\
2  & JohnDoe123 & Daredevil & example\_user\_123 \\
3  & guest & drax2 & john\_doe \\
4  & johnsmith & StarLord852 & john\_doe \\
5  & artlover & falconFlyHigh & example\_user \\
6  & doe & spiderVerse & john\_doe \\
7  & joe\_doe & quarkQuest & john\_doe \\
8  & food\_blogger & vision\_android & instagram\_user \\
9  & securepassword123 & X1yZ2aB3cD4eF5gH\% & SecureP@ss123 \\
10 & jane\_doe & algorithmWizard & exampleUser123 \\
11 & user3 & fashionista890 & example\_user\_123 \\
12 & JohnDoe & algorithmKing\_321 & exampleUser \\
13 & creative\_ideas & mariaHill\_2013 & creative\_user \\
14 & jane\_doe & hawkeye\_archer & john\_doe \\
15 & emmawatson & blackWidow\_agent & example\_user\_123 \\
16 & janedoe & thorHammer2021 & example\_user\_123 \\
17 & cristiano & vintageVibes\_456 & exampleuser \\
18 & AdventureSeeker & antMan\_shrink & example\_user\_123 \\
19 & leetcode\_user5 & heapHierophant\_321 & leetcode\_user \\
20 & NatureLover & gamora\_234 & example\_user\_123 \\
\bottomrule
\caption{Parameter values comparison - Username (social media).}
\label{tbl:apigen username}
\end{longtable}

{\small
\begin{longtable}{@{}llcccc@{}}
\toprule
\textbf{Parameter Group} & \textbf{Metric} & \textbf{APIGen} & \textbf{APIGen (Ent.\ swap)} & \textbf{Ours} & \textbf{Temp 1.0} \\
\midrule
\endfirsthead
\toprule
\textbf{Parameter Group} & \textbf{Metric} & \textbf{APIGen} & \textbf{APIGen (Ent.\ swap)} & \textbf{Ours} & \textbf{Temp 1.0} \\
\midrule
\endhead
\multirow{2}{*}{\texttt{Domain}}
& NCD Diversity & 0.484 (0.021) & 0.501 (0.023) & \textbf{0.726 (0.015)} & 0.365 (0.032) \\
& Cluster Entropy & 3.522 (0.129) & 3.641 (0.166) & \textbf{\underline{4.322 (0.000)}}$^{\dagger}$ & 1.457 (0.193) \\
\midrule
\multirow{2}{*}{Year}
& NCD Diversity & 0.300 (0.005) & 0.303 (0.006) & 0.315 (0.004) & 0.276 (0.007) \\
& Cluster Entropy & 3.784 (0.106) & 3.684 (0.114) & \textbf{\underline{4.322 (0.000)}}$^{\dagger}$ & 2.766 (0.123) \\
\midrule
\multirow{2}{*}{Location / Geographic}
& NCD Diversity & 0.550 (0.017) & 0.530 (0.014) & 0.574 (0.011) & 0.506 (0.014) \\
& Cluster Entropy & 4.084 (0.104) & 3.784 (0.121) & \textbf{\underline{4.322 (0.000)}}$^{\dagger}$ & 3.284 (0.125) \\
\midrule
\multirow{2}{*}{Currency}
& NCD Diversity & 0.377 (0.019) & 0.381 (0.018) & 0.338 (0.017) & 0.375 (0.016) \\
& Cluster Entropy & 3.222 (0.150) & 3.241 (0.172) & \textbf{\underline{4.322 (0.000)}}$^{\dagger}$ & 2.320 (0.178) \\
\midrule
\multirow{2}{*}{Email}
& NCD Diversity & 0.648 (0.007) & 0.568 (0.020) & \textbf{0.697 (0.008)} & 0.483 (0.018) \\
& Cluster Entropy & 4.222 (0.056) & 3.241 (0.183) & \underline{4.322 (0.000)}$^{\dagger}$ & 1.578 (0.125) \\
\midrule
\multirow{2}{*}{Username (social media)}
& NCD Diversity & 0.559 (0.011) & 0.540 (0.017) & \textbf{0.643 (0.005)} & 0.499 (0.010) \\
& Cluster Entropy & 4.222 (0.060) & 3.641 (0.168) & \underline{4.322 (0.000)}$^{\dagger}$ & 2.671 (0.172) \\
\midrule
\midrule
\multirow{2}{*}{\textbf{Average}}
& NCD Diversity & 0.486 (0.013) & 0.471 (0.016) & \textbf{0.549 (0.010)} & 0.417 (0.016) \\
& Cluster Entropy & 3.843 (0.101) & 3.539 (0.154) & \textbf{\underline{4.322 (0.000)}}$^{\dagger}$ & 2.346 (0.153) \\
\bottomrule
\caption{Comparison of parameter diversity across four generation methods for six parameter groups identified by clustering analysis. For each parameter group and diversity metric, we report the mean and bootstrap standard deviation (in parentheses), computed over 20 parameter values. Bootstrap standard deviations are calculated by repeatedly sampling 80\% of the parameter values uniformly at random without replacement and computing the metric on each sample. \textbf{Bold values} indicate statistically significant superiority over all other methods at the 95\% confidence level, determined by the criterion $|\mu_1 - \mu_2| > 1.96\sqrt{\sigma_1^2 + \sigma_2^2}$ for all pairwise comparisons. Underlining with a dagger ($^{\dagger}$) marks the theoretical maximum for the metric: for Cluster Entropy with 20 parameters, the maximum value is $\log_2(20) = 4.322$.}
\label{tbl:apigen parameter_diversity}
\end{longtable}}

\begin{landscape}
\begin{table}[p]
\centering
\begingroup
\footnotesize
\setlength{\tabcolsep}{3.2pt}
\begin{adjustbox}{max width=\linewidth,center}
\begin{tabular}{@{}l*{15}{c}@{}}
\toprule
& & \multicolumn{2}{c}{\textbf{Our Dataset}} & \multicolumn{2}{c}{\textbf{ToolAce Dataset}} & \multicolumn{2}{c}{\textbf{APIGen Dataset}} & \multicolumn{2}{c}{\textbf{ToolAce (Para.)}} & \multicolumn{2}{c}{\textbf{ToolAce (Ent.\ swap)}} & \multicolumn{2}{c}{\textbf{APIGen (Para.)}} & \multicolumn{2}{c}{\textbf{APIGen (Ent.\ swap)}} \\
\cmidrule(lr){3-4} \cmidrule(lr){5-6} \cmidrule(lr){7-8} \cmidrule(lr){9-10} \cmidrule(lr){11-12} \cmidrule(lr){13-14} \cmidrule(lr){15-16}
\textbf{Metric} & \textbf{Type} & \textbf{Score} & \textbf{Std} & \textbf{Score} & \textbf{Std} & \textbf{Score} & \textbf{Std} & \textbf{Score} & \textbf{Std} & \textbf{Score} & \textbf{Std} & \textbf{Score} & \textbf{Std} & \textbf{Score} & \textbf{Std} \\
\midrule
TTR & Lexical & 0.2107 & 0.0010 & 0.1577 & 0.0012 & 0.1934 & 0.0016 & 0.2092 & 0.0013 & 0.1786 & 0.0013 & 0.1925 & 0.0014 & 0.2069 & 0.0015 \\
Compression Ratio Diversity & Lexical & \textbf{0.3808} & 0.0008 & 0.3210 & 0.0016 & 0.3527 & 0.0015 & 0.3735 & 0.0009 & 0.3486 & 0.0014 & 0.3516 & 0.0011 & 0.3608 & 0.0013 \\
Var(FKGL) & Lex./Syn. & \textbf{24.4330} & 0.3063 & 9.5889 & 0.2080 & 11.3507 & 0.2024 & 12.6713 & 0.2979 & 9.7335 & 0.2146 & 10.6301 & 0.1968 & 11.1636 & 0.2099 \\
Parse Tree Entropy & Syntactic & 10.7959 & 0.0025 & 10.7892 & 0.0029 & 10.7547 & 0.0047 & \textbf{10.8037} & 0.0017 & 10.7921 & 0.0025 & 10.7977 & 0.0021 & 10.7652 & 0.0043 \\
Var(Length) & Syntactic & 150.8306 & 2.4315 & \textbf{1592.1097} & 263.7907 & 72.3491 & 2.3585 & 230.7761 & 6.7615 & 783.2886 & 32.8223 & 51.8765 & 1.9011 & 70.6532 & 2.0554 \\
Paraphrase Variety (Avg Sim) & Semantic & \textbf{0.9428} & 0.0006 & 0.9129 & 0.0008 & 0.9063 & 0.0007 & 0.9111 & 0.0009 & 0.9137 & 0.0008 & 0.9031 & 0.0008 & 0.9059 & 0.0007 \\
Chamfer Distance Score & Semantic & \textbf{0.4771} & 0.0022 & 0.4118 & 0.0022 & 0.3220 & 0.0026 & 0.4544 & 0.0021 & 0.4523 & 0.0019 & 0.3347 & 0.0025 & 0.3446 & 0.0021 \\
Vendi Score & Semantic & \textbf{171.2605} & 0.5818 & 132.3699 & 1.0812 & 136.7663 & 0.5626 & 152.3683 & 0.7031 & 149.9300 & 0.7409 & 136.7206 & 0.5823 & 138.8024 & 0.5308 \\
\midrule
Simpson's Index & Lexical & 0.9901 & 0.0001 & 0.9898 & 0.0001 & 0.9859 & 0.0001 & 0.9892 & 0.0001 & 0.9899 & 0.0001 & 0.9847 & 0.0001 & 0.9861 & 0.0001 \\
Tree Edit Distance & Syntactic & \textbf{23.3919} & 0.1368 & 17.2244 & 0.1362 & 19.9884 & 0.1386 & 21.2785 & 0.1603 & 17.3181 & 0.1371 & 21.9569 & 0.1278 & 19.9006 & 0.1273 \\
Cluster Entropy & Semantic & 10.7715 & 0.0047 & 10.4026 & 0.0219 & 10.6748 & 0.0094 & \textbf{10.8037} & 0.0014 & 10.7534 & 0.0075 & 10.7349 & 0.0065 & 10.7382 & 0.0066 \\
Semantic Spread & Semantic & \textbf{0.9707} & 0.0003 & 0.9552 & 0.0004 & 0.9517 & 0.0004 & 0.9542 & 0.0005 & 0.9556 & 0.0004 & 0.9501 & 0.0004 & 0.9515 & 0.0004 \\
\midrule
Total Queries & -- & 1793 & -- & 1793 & -- & 1793 & -- & 1793 & -- & 1793 & -- & 1793 & -- & 1793 & -- \\
\bottomrule
\end{tabular}
\end{adjustbox}
\endgroup
\caption{Linguistic diversity scores comparison between our dataset, baseline datasets (ToolAce, APIGen), and augmented baselines (paraphrasing and entity swapping) with bootstrap standard deviations (from 100 subsamples of 80\% of data). \textbf{Bold} scores mark the highest value in that row when it is statistically significant at $\alpha = 0.05$ versus every other column under $|\mu_1 - \mu_2| > 1.96\sqrt{\sigma_1^2 + \sigma_2^2}$. Higher values indicate greater diversity for all metrics.}
\label{tbl:diversity_scores_toolace_apigen_full}
\end{table}
\end{landscape}

\begin{table}[htbp]
\centering
\small
\begin{tabular}{@{}lccccc@{}}
\toprule
& & \multicolumn{2}{c}{\textbf{Our Dataset (Subsampled)}} & \multicolumn{2}{c}{\textbf{BFCL Benchmark}} \\
\cmidrule(lr){3-4} \cmidrule(lr){5-6}
\textbf{Metric} & \textbf{Type} & \textbf{Score} & \textbf{Std Dev} & \textbf{Score} & \textbf{Std Dev} \\
\midrule
TTR & Lexical & \textbf{0.2389} & 0.0013 & 0.1554 & 0.0016 \\
Compression Ratio Diversity & Lexical & \textbf{0.3833} & 0.0010 & 0.3176 & 0.0014 \\
Var(FKGL) & Lex./Syn. & \textbf{24.4315} & 0.3950 & 17.0292 & 0.6034 \\
Parse Tree Entropy & Syntactic & \textbf{10.2632} & 0.0029 & 10.0481 & 0.0109 \\
Var(Length) & Syntactic & 155.1082 & 3.4905 & \textbf{808.6525} & 32.9232 \\
Paraphrase Variety (Avg Sim) & Semantic & \textbf{0.9426} & 0.0006 & 0.9175 & 0.0008 \\
Chamfer Distance Score & Semantic & \textbf{0.4975} & 0.0028 & 0.2908 & 0.0039 \\
Vendi Score & Semantic & \textbf{165.5398} & 0.5949 & 129.2562 & 0.7663 \\
\midrule
Simpson's Index & Lexical & \textbf{0.9900} & 0.0001 & 0.9871 & 0.0001 \\
Tree Edit Distance & Syntactic & \textbf{23.6807} & 0.1883 & 21.0218 & 0.2193 \\
Cluster Entropy & Semantic & \textbf{10.2465} & 0.0040 & 10.0652 & 0.0104 \\
Semantic Spread & Semantic & \textbf{0.9705} & 0.0003 & 0.9575 & 0.0004 \\
\midrule
Total Queries & -- & 1240 & -- & 1240 & -- \\
\bottomrule
\end{tabular}
\caption{Linguistic diversity scores comparison between our dataset (subsampled to match BFCL size) and the BFCL benchmark with bootstrap standard deviations (from 100 subsamples of 80\% of data). Both datasets contain 1240 queries. Bold values indicate statistically significant differences at $\alpha = 0.05$ level using $|\mu_1 - \mu_2| > 1.96\sqrt{\sigma_1^2 + \sigma_2^2}$. Higher values indicate greater diversity for all metrics.}
\label{tbl:diversity_scores_bfcl}
\end{table}

\twocolumn

\subsection{BFCL category errors}\label{app:bfcl category errors}

In \Cref{tab:bfcl_benchmark}, the model fine-tuned on our data trails both baselines on \texttt{IRRELEVANCE} and trails APIGen on \texttt{PARALLEL\_MULTIPLE}. Every \texttt{IRRELEVANCE} error, for all three models, is a function call on an example where the gold behavior is to call nothing. The offered tool overlaps the query in its name or parameters. For example, on ``Find out about traffic laws in Texas.'', our model calls \texttt{lawsuit\_search}. Out of 240 examples, this occurs 17 times for our model, 11 times for ToolAce, and 6 times for APIGen.

On \texttt{PARALLEL\_MULTIPLE} (200 examples), three error types occur. An output that cannot be parsed as a list of calls never occurs for our model; it occurs for 2 (1\%) APIGen examples and 4 (2\%) ToolAce examples. A wrong number of calls, usually too few, occurs for $10\%$ of examples for our model, $5.5\%$ for APIGen, and $36\%$ for ToolAce. The third type is a call list of the right length that still fails to match, in the function name or, typically, a parameter value; this occurs for $10\%$ of examples for our model and for APIGen, and $7\%$ for ToolAce. A wrong call count can arise only when several functions are required, so ToolAce's weakness on this category is easy to miss in the average over all categories.

\section{Fine-tuning Details}\label{sec:fine_tuning_details}

% As reported in the main text, we fine-tuned three models 

% Llama-3.1-8B-Instruct \citep{grattafiori2024llama3herdmodels} using LoRA \citep{hu2022lora} for each of the three training sets described above, resulting in one model per training set

% We trained 3 models. For ToolAce we used their reported HPs. For Ours we observed the HPs of ToolAce performed well on a validation fold, so we used the same. For APIGen, these HPs did not perform well, and we used HPs closer to those reported by them. The reason that we did not take their exact paramters is that they performed full fine-tuning, which is less suitable for the smaller subsample that we used here, hence we used different HPs and Lora of the same rank as the other models, $16$.

% %%%%%%%%%%%%

As reported in the main text, for our experiments in \S\ref{sec:LLM Fine-tuning} we fine tuned three models based on Llama-3.1-8B-Instruct \citep{grattafiori2024llama3herdmodels} using Low Rank Adaptaion (LoRA) fine-tuning using the \texttt{transformers}, \texttt{datasets}, \texttt{accelerate}, \texttt{peft} and \texttt{trl} libraries from Hugging Face \citep{hu2022lora,wolf2020transformers}.

For all three models, we used the hyperparameters reported by \citet{liu2409toolace} for the ToolAce model. Specifically, we set the LoRA rank to $16$, LoRA $\alpha$ to 32, batch size to $48$, number of training epochs to $3$, warmup ratio to $0.1$, and the mixed precision to `fp16'.
In their paper, they used a cosine decaying learning rate that starts at $10^{-4}$, which we applied in our training for all models. However, the model fine-tuned on APIGen performed very poorly under these hyperparemters, so to enable a fair comparison, we examined a handful of alternative hyperparameter settings to find one that performs well on a held out validation set (random subset of their data). We ended up with the same parameters other than a different learning rate, decreased to $2 \times 10^{-5}$. Note that \citet{liu2024apigen} did report a set of hyperparameters for fine-tuning, but these were not applicable for our setting since they were for full fine-tuning rather than LoRA, and for a much larger training set.

\section{Prompts}\label{sec:prompts_appendix}
Our data generation pipeline employs multiple specialized prompt signatures across four main stages: (1) function selection, (2) argument value generation, (3) query generation, and (4) distractor selection. Each signature is carefully designed to elicit specific structured outputs from the language model. This section presents representative examples from each stage, with complete prompts available in Appendix~\ref{app:full-prompts}. Augmentation baselines used in \S\ref{sec:Analysis} are specified in Appendix \ref{sec:prompts-augmentation-baselines}.

\subsection{Augmentation baseline prompts}\label{sec:prompts-augmentation-baselines}

We construct four intrinsic-diversity baselines by applying two LLM procedures to the released ToolAce and APIGen corpora (same examples and tool schemas as the originals). In both procedures the model receives the original user query, a short summary of tool names and descriptions, and the ground-truth assistant content; it returns four candidates and one is sampled uniformly at random.

\paragraph{Paraphrase baseline.} The model must output four paraphrases of the user query that each preserve the exact same ground-truth tool invocation (same function and arguments). The instruction stresses that every paraphrase should still lead a competent model to the provided target call.

\paragraph{Entity-swapping baseline.} The model outputs four \texttt{(new\_query, new\_ground\_truth)} pairs. If the ground-truth call lists parameters, it should prefer changing named entities tied to those parameters in the query and emit the matching updated call (same tools invoked; only parameter values may change). If the call is empty or has no parameters, it changes other named entities when possible and keeps the ground truth unchanged. If no named entities can be changed, it returns the original pair unchanged.

\subsection{Function Selection}
\label{sec:prompts-function-selection}

The function selection stage validates compatibility between API schemas before parameter generation to ensure sequential chains are feasible.

\subsubsection{Validate Sequential Schema Compatibility}

This signature validates whether API schemas support sequential execution, where later APIs can depend on earlier APIs' return values.

\paragraph{System Prompt:} The signature instructs the model to analyze whether the Kth API's parameters can depend on outputs from previous K-1 calls, and whether this dependency is mandatory. It checks for: (1) return type compatibility with subsequent parameter types, (2) schema structure supporting sequential dependency, and (3) mandatory dependency requirements.

\paragraph{Example Input:} Given two TikTok APIs -- one retrieving video captions and another getting encoded video information -- the model analyzes whether the caption retrieval output can serve as input for the video information API.

\paragraph{Example Output:} The model determined these schemas are incompatible (\texttt{is\_compatible: NO}) because the first API returns caption data that does not provide the video ID required by the second API, breaking the sequential dependency chain.

\subsection{Argument Value Generation}
\label{sec:prompts-argument-generation}

This stage generates diverse, feasible parameter values while maintaining consistency across API invocations.

\subsubsection{Generate Multiple String Parameters}

Generates diverse string parameter values in batch, with explicit diversity requirements.

\paragraph{System Prompt:} The model receives parameter metadata (name, description, type, function context) and must generate multiple diverse candidates that match the parameter specification while differing from existing values.

\paragraph{Example Input:} For a TikTok video URL parameter, the model generates 25 diverse URL candidates.

\paragraph{Example Output:} The model produces 25 structurally valid TikTok URLs with varied user names and video IDs: \texttt{https://www.tiktok.com/@user1/}\allowbreak\texttt{video/}\allowbreak\texttt{1234567890}, \texttt{https://www.tiktok.com/@user2/}\allowbreak\texttt{video/}\allowbreak\texttt{abcdef123}, etc.

\subsubsection{Generate Sequential Cohesive String Parameter}

Generates string parameters for sequential API chains, prioritizing cohesion with return values.

\paragraph{System Prompt:} The model must generate parameters that are logically related to: (1) the return value this API will produce (highest priority), (2) parameters of the next API in the chain, and (3) the API's general purpose. The return value context is emphasized as the most important factor since it will be used by subsequent APIs.

\paragraph{Example Input:} For a TikTok video URL parameter where the API will return a watermark-free URL that feeds into the next API, the model receives both the expected return value structure and the next API's parameter requirements.

\paragraph{Example Output:} The model generates \texttt{https://www.tiktok.com/@user17/}\allowbreak\texttt{video/}\allowbreak\texttt{fedcba543210}, which matches the video ID in the expected return value and subsequent API parameters.

\subsubsection{Generate Sequential Cohesive-\\NumericalParameter}

Similar to string generation but for numerical parameters in sequential chains.

\paragraph{System Prompt:} Uses the same prioritization as string generation (return value $>$ next API parameters $>$ API purpose) but handles numerical constraints and defaults.

\paragraph{Example Input:} For a ``fresh'' flag parameter (forcing fresh vs. cached data) in a video URL retrieval API.

\paragraph{Example Output:} The model generates \texttt{1.0} to force fresh data retrieval, aligning with the need for current video information.

\subsubsection{Generate Return Value}

Generates realistic API return values that will be consumed by subsequent APIs.

\paragraph{System Prompt:} The model generates return values matching the schema that must be: (1) consistent with the API's purpose, (2) useful for the next API's requirements, and (3) cohesive with provided parameter values.

\paragraph{Example Input:} For an API that retrieves watermark-free TikTok video URLs, given that the next API needs a TikTok URL to fetch details.

\paragraph{Example Output:} Returns a CDN URL structure: \texttt{\{"url": "https://v16.tiktok.com/}\allowbreak\texttt{video/tos/useast2a/...\}}, containing the video ID that matches the next API's expected input.

\subsubsection{Generate Cohesive String Parameter}

Generates string parameters for parallel API invocations, emphasizing logical relationships.

\paragraph{System Prompt:} The model must generate values that are logically related to parameters from other APIs executing in parallel, potentially using identical values (e.g., same location, date, or identifier) to ensure coherence.

\paragraph{Example Input:} For a travel mode parameter in a route duration estimator, given parallel context showing a route planner API with Auckland to Boston journey details.

\paragraph{Example Output:} Generates \texttt{transit} as the appropriate mode for the long-distance international journey.

\subsubsection{Parameter Set Validator}

Validates that a complete set of parameter values forms a coherent, conflict-free API call.

\paragraph{System Prompt:} The model checks for: (1) missing required dependencies, (2) conflicting parameters, (3) invalid combinations, and (4) type mismatches.

\paragraph{Example Input:} A TikTok video details API call with a single URL parameter.

\paragraph{Example Output:} Validates as correct (\texttt{is\_valid: YES}) since all required parameters are present with appropriate types and no conflicts exist.

\subsubsection{Validate Return Value}

Validates return values for appropriateness in sequential chains.

\paragraph{System Prompt:} Checks whether return values: (1) match the schema, (2) are logically consistent with API purpose, (3) align with input parameters, and (4) are useful for subsequent APIs.

\paragraph{Example Input:} A watermark-free URL return value for a TikTok video API that feeds into a video details API.

\paragraph{Example Output:} Validates as correct (\texttt{is\_valid: YES}) because the URL structure matches the schema and contains the video ID needed by the next API.

\subsubsection{Validate Sequential Chain}

Validates entire sequential API execution chains for coherence and strict sequential dependency.

\paragraph{System Prompt:} The model performs comprehensive validation: (1) individual parameter validity, (2) individual return value validity, (3) \textbf{strict sequential dependency verification} -- each API (except first) must require information from the previous API's return value that cannot be obtained from the query alone, (4) impossibility of parallel execution, and (5) cohesion between return values and subsequent parameters.

\paragraph{Example Input:} A two-API chain: first retrieving a watermark-free TikTok URL, then getting video details using that URL.

\paragraph{Example Output:} Marked invalid (\texttt{is\_valid: NO}) because the second API can use the original TikTok URL directly from the query without requiring the first API's watermark-free URL, indicating weak sequential dependency.

\subsection{Query Generation}
\label{sec:prompts-query-generation}

This stage generates natural language queries that justify the API invocations while maintaining dataset diversity.

\subsubsection{No API Query Generator}

Generates queries answerable without any API calls.

\paragraph{System Prompt:} The model generates diverse natural language queries that can be answered with general knowledge, without requiring external API calls.

\paragraph{Example Output:} Queries like ``What are the primary causes of climate change and how do they impact the environment?'' that require knowledge synthesis rather than API invocations.

\subsubsection{Sequential Query Generator}

Generates queries for sequential API chains, emphasizing end goals rather than execution steps.

\paragraph{System Prompt:} The model must describe the \textbf{end goal or desired outcome}, not the sequence of steps. It should avoid: (1) explicitly mentioning API call sequences, (2) instructional language revealing the sequence, (3) telling the model what steps to take. Instead, it should make the query natural and let the model infer that sequential calls are needed.

\paragraph{Example Input:} A two-API chain for TikTok video analysis (watermark removal then details retrieval).

\paragraph{Example Output:} ``Provide a detailed summary of the TikTok video located at [URL], including the title, description, upload date, number of views, likes, and comments.'' The query describes the desired information without revealing the intermediate watermark removal step.

\subsubsection{Parallel Query Generator}

Generates queries requiring multiple simultaneous API calls.

\paragraph{System Prompt:} Generates queries where multiple APIs should be called together in parallel, scaling tool calls to query complexity (1 for simple facts, 3-5 for medium tasks, 5-10 for deep research).

\paragraph{Example Output:} ``Plan a route from Sydney to Melbourne departing at 8:00 AM and estimate the travel time by car considering optimistic traffic conditions.'' This naturally requires both route planning and duration estimation APIs in parallel.

\subsubsection{Multi Query Generator}

Generates diverse queries for single API invocations.

\paragraph{System Prompt:} The model generates multiple natural language queries for a specific API call with given parameters, maximizing diversity across semantic, pragmatic, syntactic, lexical, stylistic, contextual, and structural dimensions.

\paragraph{Example Input:} A disaster resilience planning API for Mexico City covering earthquakes, floods, fires, and hurricanes over one year.

\paragraph{Example Output:} Diverse phrasings from direct requests (``Can you suggest some community-based initiatives...'') to hypothetical scenarios (``Imagine you're a city planner...'') to narrative prompts (``Tell me a story about how Mexico City prepared...'').

\subsubsection{Missing Params Query Generator}

Generates queries that require a specific API but intentionally lack required parameters.

\paragraph{System Prompt:} Creates queries where the API is clearly needed but some required parameters cannot be inferred from the query text, testing the model's ability to recognize missing information.

\paragraph{Example Input:} A hotel reviews API requiring a hotel ID, with only language code provided.

\paragraph{Example Output:} ``Can you tell me about the most memorable guest experiences shared in the featured reviews for that hotel?'' The query clearly needs reviews but doesn't specify which hotel.

\subsubsection{Sequential Query Judge}

Validates queries for sequential API invocations.

\paragraph{System Prompt:} Verifies that: (1) all APIs are required and non-redundant, (2) the first API's parameters can be inferred from the query, (3) subsequent APIs' parameters can be inferred from the query and/or previous return values, and (4) the sequential chain is logically appropriate.

\paragraph{Example Input:} Query for TikTok video summary with a two-API chain (watermark removal then details retrieval).

\paragraph{Example Output:} Marked as unreasonable (\texttt{is\_reasonable: NO}) because the first API (watermark removal) is not necessary -- the second API can directly use the original URL from the query.

\subsubsection{Parallel Query Judge}

Validates queries for parallel API invocations.

\paragraph{System Prompt:} In parallel execution, all parameters must be inferable \textbf{directly from the query text alone}, not from other API outputs. Validates: (1) all APIs are required, (2) all parameters can be inferred from query text only, and (3) parallel execution is appropriate (no sequential dependencies).

\paragraph{Example Input:} Route planning query with mismatched locations (query asks for Sydney-Melbourne but parameters specify Auckland-Boston).

\paragraph{Example Output:} Marked invalid (\texttt{is\_reasonable: NO}) due to parameter-query mismatch and incorrect travel mode.

\subsubsection{API Query Judge}

Validates queries for single API invocations.

\paragraph{System Prompt:} Performs step-by-step analysis to determine if the proposed API call with given parameters is reasonable for answering the user's query.

\paragraph{Example Input:} Query about disaster resilience planning for Mexico City with matching parameters.

\paragraph{Example Output:} Validated (\texttt{is\_reasonable: YES}) because the API and parameters directly address the query requirements.

\subsubsection{Missing Params Query Judge}

Validates that queries appropriately lack specified required parameters.

\paragraph{System Prompt:} Determines if a query reasonably requires a specific API but is genuinely missing required parameters that cannot be inferred from the query text.

\paragraph{Example Input:} Hotel reviews query without hotel ID.

\paragraph{Example Output:} Validated (\texttt{is\_reasonable: YES}) because the query clearly needs the reviews API but lacks the essential hotel identifier.

\subsubsection{Diversity Enhancement Signatures}

To combat homogeneity in generated queries, we employ three specialized signatures:

\paragraph{Dataset Pattern Analysis:} Identifies patterns of homogeneity across seven dimensions: semantic clustering, pragmatic patterns, implicit assumptions, discourse structures, persona/voice homogeneity, and contextual patterns. For example, it detected that a sample dataset was dominated by formal, informational requests about specific topics (TikTok summaries, climate change), limiting diversity.

\paragraph{Diversity Guidance Generation:} Converts pattern analysis into actionable guidance. For the homogeneous sample, it recommended: avoiding semantic clustering by introducing varied topics, exploring different speech acts (commands, opinions, hypotheticals), varying syntactic complexity, diversifying tone and voice, and expanding contextual assumptions.

\paragraph{Integration:} These signatures work iteratively -- patterns are analyzed periodically during generation, guidance is produced, and subsequent query generators receive this guidance to steer toward underrepresented dimensions.

\subsection{Distractor Selection}
\label{sec:prompts-distractor-selection}

This stage retrieves and scores candidate APIs for inclusion as distractors in the dataset.

\subsubsection{Batch API Relevance Scorer}

Scores multiple API candidates for relevance to a query on a 1-5 scale.

\paragraph{System Prompt:} The model assigns scores where 5 indicates perfect match for query intent, 4 is very relevant, 3 is somewhat relevant, 2 is weakly relevant, and 1 is not relevant.

\paragraph{Example Input:} Climate change query with 20 API candidates including news retrieval, impact assessment, and prediction APIs.

\paragraph{Example Output:} Highest scores (5) for news retrieval APIs providing comprehensive climate information; moderate scores (3-4) for specific impact assessments; lowest score (1) for unrelated APIs like house plant recommendations.

\subsubsection{Parallel API Relevance Scorer}

Scores API candidates considering parallel execution context with target APIs.

\paragraph{System Prompt:} Similar to batch scoring but evaluates candidates for use alongside other APIs in a single query, considering compatibility and complementary functionality.

\paragraph{Example Input:} TikTok video summary query with target APIs for watermark removal and details retrieval.

\paragraph{Example Output:} High scores (5) for APIs providing comprehensive video metadata and content; moderate scores (4) for supplementary information like music details and comments; low scores (1-2) for unrelated functionality like device registration.

\subsubsection{Sequential and Parallel Invocation Construction}

\paragraph{ConstructSequentialInvocation:} Given a query and previous API calls in a chain, constructs the next API invocation. Uses return values from earlier APIs to determine parameters for the next call. For example, it determined that for a TikTok video summary, the first call should be to \texttt{Video Details by URL} with the URL from the query.

\paragraph{ValidateSequentialInvocation:} Validates complete sequential chains by checking: (1) API relevance, (2) first API parameters inferable from query, (3) subsequent API parameters inferable from query/previous returns, (4) logical sequential execution, and (5) completeness for the query.

\paragraph{ConstructParallelInvocation:} Constructs parallel API invocations from available APIs. For a tournament website design query, it selected both logo retrieval and placeholder image APIs with tournament ID 22022, reasoning they work together to fulfill the request.

\paragraph{ValidateParallelInvocation:} Validates parallel invocations by ensuring: (1) all APIs are relevant, (2) parameters are inferable from query text alone, (3) no sequential dependencies exist, and (4) the combination makes sense.

\subsection{Evaluation}\label{sec:prompts_llm_judge_eval}
The evaluation stage assesses whether predicted tool calls are semantically equivalent to ground truth calls, accounting for variations in parameter representation that would still produce identical results.
\subsubsection{Tool Call Equivalence}
This signature evaluates whether two tool calls are functionally equivalent by comparing tool names and analyzing parameter semantic equivalence rather than strict syntactic matching.
\paragraph{System Prompt:} The signature instructs the model to perform a step-by-step analysis considering: (1) whether both calls use the same tool name, (2) whether parameter values are semantically equivalent (e.g., `NYC' equals `New York'), and (3) whether the calls would produce the same result for the user's query. The model outputs reasoning followed by a binary YES/NO equivalence judgment.
\paragraph{Example Input:} Given a user query to find Italian restaurants in New York, the ground truth call uses \texttt{{``location'': ``New York''}} while the predicted call uses \texttt{{``location'': ``NYC''}}. Both calls use the \texttt{get\_restaurants} tool with identical cuisine parameters.
\paragraph{Example Output:} The model determined these calls are equivalent (\texttt{equivalent: YES}) because ``New York'' and ``NYC'' refer to the same city, and both tool calls would return identical restaurant results despite the location parameter variation.

\section{Complete Prompt Specifications}
\label{app:full-prompts}

This appendix provides complete prompt specifications for all signatures used in our data generation pipeline. Each signature includes the full system prompt, input field definitions, output field definitions, and the structured format used for interaction.

\subsection{Function Selection Prompts}

\subsubsection{Validate Sequential Schema Compatibility}

\paragraph{Objective:} Validate if API schemas are compatible for sequential execution BEFORE generating parameters.

\paragraph{System Prompt:}
\begin{tiny}
\begin{verbatim}
Your input fields are:
1. `api_schemas` (str): JSON array of API schemas for all APIs in the 
   proposed sequential chain, in execution order

Your output fields are:
1. `reasoning` (str): Step-by-step analysis: (1) For each API (except 
   first), check if its parameter schemas can accept information from 
   previous APIs' return value schemas, (2) Verify that return types are 
   compatible with subsequent parameter types (type matching, structure 
   compatibility), (3) Check if the K-th API's parameters CAN depend on 
   outputs from previous K-1 calls, (4) Verify that the dependency on the 
   (K-1)-th API's output is MANDATORY (the K-th API must be able to 
   require it), (5) Assess if schemas support true sequential dependency 
   vs parallel execution, (6) Determine if the schemas are compatible for 
   sequential execution.
2. `is_compatible` (str): 'YES' if the API schemas are compatible for 
   sequential execution (later APIs can and must depend on earlier API 
   outputs). 'NO' if schemas don't support sequential dependency, if 
   return types are incompatible with parameter types, or if APIs can 
   only be called in parallel. Return only 'YES' or 'NO'.

All interactions will be structured in the following way, with the 
appropriate values filled in.

[[ ## api_schemas ## ]]
{api_schemas}

[[ ## reasoning ## ]]
{reasoning}

[[ ## is_compatible ## ]]
{is_compatible}

[[ ## completed ## ]]

In adhering to this structure, your objective is: 
Validate if API schemas are compatible for sequential execution BEFORE 
generating parameters.

This validates that the API schemas themselves allow for a true sequential 
chain where:
1. Each API (except the first) CAN receive information from previous APIs' 
   return values
2. Each API (except the first) MUST be able to depend on the immediately 
   previous API's output
3. The return types of earlier APIs are compatible with parameter types of 
   later APIs
4. The schemas support the possibility that parameters of the K-th API 
   depend on outputs of previous K-1 calls
5. The dependency on the (K-1)-th API's output is mandatory (the K-th API 
   must be able to use it)

CRITICAL: This checks SCHEMA COMPATIBILITY only - it doesn't validate 
actual parameter values. It determines if the schemas ALLOW for sequential 
dependency, not whether specific values work.

The judge should check:
- Return type compatibility: Can return values from API i be used as 
  parameters for API i+1?
- Type matching: Do return value types match or can be converted to 
  parameter types?
- Schema structure: Do the schemas support extracting needed information 
  from return values?
- Mandatory dependency: Can API K's parameters be structured to require 
  API (K-1)'s output?
- Sequential possibility: Is it possible (based on schemas) that later 
  APIs depend on earlier outputs?

REJECT if:
- Return types cannot be used as parameter inputs for subsequent APIs
- Schemas don't support sequential dependency (e.g., all parameters are 
  independent)
- The chain cannot have mandatory dependency on previous API outputs
- APIs are inherently parallel (schemas show no dependency possibility)
\end{verbatim}
\end{tiny}

\subsection{Argument Value Generation Prompts}

\subsubsection{Generate Multiple String Parameters}

\paragraph{Objective:} Generate multiple diverse and feasible string parameter values in one call.

\paragraph{System Prompt:}
\begin{tiny}
\begin{verbatim}
Your input fields are:
1. `parameter_name` (str): Name of the parameter
2. `parameter_description` (str): Description of what this parameter does
3. `parameter_type` (str): Data type of the parameter
4. `function_name` (str): Function this parameter belongs to
5. `function_description` (str): Description of what this function does
6. `other_parameter_values` (str): The value of the other parameters for 
   this function that were already determined
7. `existing_values` (str): Previously generated values for this parameter
8. `parameter_group_context` (str): Information about related parameters 
   in the group
9. `num_candidates` (str): Number of diverse candidates to generate
10. `previous_failures` (str): Information about previous generation 
    attempts that failed, including error messages and validation feedback. 
    Use this to avoid repeating the same mistakes. Format: 'None' if no 
    previous failures, otherwise a description of what went wrong.

Your output fields are:
1. `reasoning` (str): 
2. `generated_values` (str): A JSON list of diverse parameter values. Each 
   should be feasible, match the parameter description, make sense in 
   context of the function, and be as different as possible from existing 
   values and each other. Format: ["value1", "value2", "value3", "value4", 
   "value5"]. ONLY return the JSON list, no explanation.

All interactions will be structured in the following way, with the 
appropriate values filled in.

[[ ## parameter_name ## ]]
{parameter_name}

[[ ## parameter_description ## ]]
{parameter_description}

[[ ## parameter_type ## ]]
{parameter_type}

[[ ## function_name ## ]]
{function_name}

[[ ## function_description ## ]]
{function_description}

[[ ## other_parameter_values ## ]]
{other_parameter_values}

[[ ## existing_values ## ]]
{existing_values}

[[ ## parameter_group_context ## ]]
{parameter_group_context}

[[ ## num_candidates ## ]]
{num_candidates}

[[ ## previous_failures ## ]]
{previous_failures}

[[ ## reasoning ## ]]
{reasoning}

[[ ## generated_values ## ]]
{generated_values}

[[ ## completed ## ]]

In adhering to this structure, your objective is: 
Generate multiple diverse and feasible string parameter values in one call.
\end{verbatim}
\end{tiny}

\subsubsection{Generate Sequential Cohesive String Parameter}

\paragraph{Objective:} Generate a string parameter value for sequential API execution.

\paragraph{System Prompt:}
\begin{tiny}
\begin{verbatim}
Your input fields are:
1. `parameter_name` (str): Name of the parameter
2. `parameter_description` (str): Description of what this parameter does
3. `parameter_type` (str): Data type of the parameter
4. `function_name` (str): Function this parameter belongs to
5. `function_description` (str): Description of what this function does
6. `other_parameter_values` (str): The value of the other parameters for 
   this function that were already determined
7. `return_value` (str): JSON string of the return value that this API 
   will produce. This is the MOST IMPORTANT context - the parameter should 
   be cohesive with this return value. The return value will be used as 
   input for the next API in the sequential chain.
8. `next_api_parameters` (str): JSON string of parameters for the next API 
   in the sequential chain. The return value of this API should provide 
   information needed by the next API's parameters.
9. `later_api_parameters` (str): JSON array of parameter dictionaries from 
   APIs later in the sequential chain. These provide additional context 
   for cohesion.
10. `existing_values` (str): Previously generated values for this parameter 
    (for context only)
11. `parameter_group_context` (str): Information about related parameters 
    in the group
12. `previous_failures` (str): Information about previous generation 
    attempts that failed. Format: 'None' if no previous failures, otherwise 
    a description of what went wrong.

Your output fields are:
1. `reasoning` (str): 
2. `generated_value` (str): A parameter value that is cohesive with the 
   return value and sequential chain. PRIORITY 1: The value should be 
   logically related to the return value this API will produce. PRIORITY 2: 
   The value should be cohesive with parameters of the next API in the 
   chain. PRIORITY 3: The value should be feasible and match the parameter 
   description. Diversity from existing values is encouraged but secondary. 
   ONLY return the parameter value as a string, no quotes, no explanation.

All interactions will be structured in the following way, with the 
appropriate values filled in.

[[ ## parameter_name ## ]]
{parameter_name}

[[ ## parameter_description ## ]]
{parameter_description}

[[ ## parameter_type ## ]]
{parameter_type}

[[ ## function_name ## ]]
{function_name}

[[ ## function_description ## ]]
{function_description}

[[ ## other_parameter_values ## ]]
{other_parameter_values}

[[ ## return_value ## ]]
{return_value}

[[ ## next_api_parameters ## ]]
{next_api_parameters}

[[ ## later_api_parameters ## ]]
{later_api_parameters}

[[ ## existing_values ## ]]
{existing_values}

[[ ## parameter_group_context ## ]]
{parameter_group_context}

[[ ## previous_failures ## ]]
{previous_failures}

[[ ## reasoning ## ]]
{reasoning}

[[ ## generated_value ## ]]
{generated_value}

[[ ## completed ## ]]

In adhering to this structure, your objective is: 
Generate a string parameter value for sequential API execution.

This signature prioritizes logical cohesion with:
1. The return value of this API (MOST IMPORTANT - this API's output will 
   be used by next API)
2. Parameters of the next API in the chain
3. Parameters of APIs later in the chain
\end{verbatim}
\end{tiny}

\subsubsection{Generate Sequential Cohesive Numerical Parameter}

\paragraph{Objective:} Generate a numerical parameter value for sequential API execution.

\paragraph{System Prompt:}
\begin{tiny}
\begin{verbatim}
Your input fields are:
1. `parameter_name` (str): Name of the parameter
2. `parameter_description` (str): Description of what this parameter does
3. `parameter_type` (str): Data type (int, float, etc.)
4. `function_name` (str): Function this parameter belongs to
5. `function_description` (str): Description of what this function does
6. `other_parameter_values` (str): The value of the other parameters for 
   this function that were already determined
7. `return_value` (str): JSON string of the return value that this API 
   will produce. This is the MOST IMPORTANT context - the parameter should 
   be cohesive with this return value. The return value will be used as 
   input for the next API in the sequential chain.
8. `next_api_parameters` (str): JSON string of parameters for the next API 
   in the sequential chain. The return value of this API should provide 
   information needed by the next API's parameters.
9. `later_api_parameters` (str): JSON array of parameter dictionaries from 
   APIs later in the sequential chain. These provide additional context 
   for cohesion.
10. `existing_values` (str): Previously generated numerical values (for 
    context only)
11. `parameter_group_context` (str): Information about related parameters
12. `previous_failures` (str): Information about previous generation 
    attempts that failed. Format: 'None' if no previous failures, otherwise 
    a description of what went wrong.

Your output fields are:
1. `reasoning` (str): 
2. `generated_value` (str): A numerical parameter value that is cohesive 
   with the return value and sequential chain. PRIORITY 1: The value should 
   be logically related to the return value this API will produce. 
   PRIORITY 2: The value should be cohesive with parameters of the next 
   API in the chain. PRIORITY 3: The value should be feasible and match 
   the parameter description. Diversity from existing values is encouraged 
   but secondary. ONLY return the numerical value, no quotes, no 
   explanation.

All interactions will be structured in the following way, with the 
appropriate values filled in.

[[ ## parameter_name ## ]]
{parameter_name}

[[ ## parameter_description ## ]]
{parameter_description}

[[ ## parameter_type ## ]]
{parameter_type}

[[ ## function_name ## ]]
{function_name}

[[ ## function_description ## ]]
{function_description}

[[ ## other_parameter_values ## ]]
{other_parameter_values}

[[ ## return_value ## ]]
{return_value}

[[ ## next_api_parameters ## ]]
{next_api_parameters}

[[ ## later_api_parameters ## ]]
{later_api_parameters}

[[ ## existing_values ## ]]
{existing_values}

[[ ## parameter_group_context ## ]]
{parameter_group_context}

[[ ## previous_failures ## ]]
{previous_failures}

[[ ## reasoning ## ]]
{reasoning}

[[ ## generated_value ## ]]
{generated_value}

[[ ## completed ## ]]

In adhering to this structure, your objective is: 
Generate a numerical parameter value for sequential API execution.

This signature prioritizes logical cohesion with:
1. The return value of this API (MOST IMPORTANT - this API's output will 
   be used by next API)
2. Parameters of the next API in the chain
3. Parameters of APIs later in the chain
\end{verbatim}
\end{tiny}

\subsubsection{Generate Return Value}

\paragraph{Objective:} Generate a realistic return value for an API that will be used as input for a subsequent API call.

\paragraph{System Prompt:}
\begin{tiny}
\begin{verbatim}
Your input fields are:
1. `api_name` (str): Name of the API function
2. `api_description` (str): Description of what the API does
3. `return_type_schema` (str): JSON string of the return type schema 
   showing the expected return structure
4. `next_api_name` (str): Name of the next API in the sequential chain 
   (if any). The return value should be cohesive with what this next API 
   needs.
5. `next_api_description` (str): Description of the next API in the chain 
   (if any). Use this to ensure the return value is useful for the next 
   API.
6. `next_api_parameters_schema` (str): JSON string of the next API's 
   parameter schema (if any). The return value should contain information 
   that can be used as input for the next API.
7. `next_api_parameters_values` (str): JSON string of the next API's 
   actual parameter values (if any). The return value should be cohesive 
   with these specific parameter values that will be used in the next API 
   call.
8. `previous_failures` (str): Information about previous generation 
   attempts that failed. Format: 'None' if no previous failures, otherwise 
   a description of what went wrong.

Your output fields are:
1. `reasoning` (str): 
2. `return_value` (str): JSON string representing a realistic return value 
   that matches the return_type_schema. The return value should be 
   logically consistent with the API's purpose and description, and should 
   be cohesive with what the next API needs as input (if next_api_name is 
   provided). Return ONLY valid JSON, no markdown, no code blocks, no 
   explanation.

All interactions will be structured in the following way, with the 
appropriate values filled in.

[[ ## api_name ## ]]
{api_name}

[[ ## api_description ## ]]
{api_description}

[[ ## return_type_schema ## ]]
{return_type_schema}

[[ ## next_api_name ## ]]
{next_api_name}

[[ ## next_api_description ## ]]
{next_api_description}

[[ ## next_api_parameters_schema ## ]]
{next_api_parameters_schema}

[[ ## next_api_parameters_values ## ]]
{next_api_parameters_values}

[[ ## previous_failures ## ]]
{previous_failures}

[[ ## reasoning ## ]]
{reasoning}

[[ ## return_value ## ]]
{return_value}

[[ ## completed ## ]]

In adhering to this structure, your objective is: 
Generate a realistic return value for an API that will be used as input 
for a subsequent API call.

This generates the return value BEFORE the API's input parameters are 
known. The return value should:
1. Match the API's return type schema
2. Be logically consistent with the API's purpose and description
3. Be cohesive with what the next API in the chain needs as input (if 
   next_api_info is provided)
\end{verbatim}
\end{tiny}

\subsubsection{Generate Cohesive String Parameter}

\paragraph{Objective:} Generate a string parameter value that logically fits with other APIs in a parallel invocation.

\paragraph{System Prompt:}
\begin{tiny}
\begin{verbatim}
Your input fields are:
1. `parameter_name` (str): Name of the parameter
2. `parameter_description` (str): Description of what this parameter does
3. `parameter_type` (str): Data type of the parameter
4. `function_name` (str): Function this parameter belongs to
5. `function_description` (str): Description of what this function does
6. `other_parameter_values` (str): The value of the other parameters for 
   this function that were already determined
7. `parallel_context_parameters` (str): JSON array of parameter 
   dictionaries from other APIs in this parallel invocation. These APIs 
   will be called together, so parameters should be logically related. 
   When appropriate, use identical values (e.g., same location, date, or 
   identifier).
8. `existing_values` (str): Previously generated values for this parameter 
   (for context only)
9. `parameter_group_context` (str): Information about related parameters 
   in the group
10. `previous_failures` (str): Information about previous generation 
    attempts that failed. Format: 'None' if no previous failures, otherwise 
    a description of what went wrong.

Your output fields are:
1. `reasoning` (str): 
2. `generated_value` (str): A parameter value that logically fits with the 
   parallel context parameters. PRIORITY 1: The value should be logically 
   related and coherent with parameters from other APIs (consider using 
   identical values when appropriate, e.g., same location, date, ID). 
   PRIORITY 2: The value should be feasible and match the parameter 
   description. Diversity from existing values is encouraged but secondary. 
   ONLY return the parameter value as a string, no quotes, no explanation.

All interactions will be structured in the following way, with the 
appropriate values filled in.

[[ ## parameter_name ## ]]
{parameter_name}

[[ ## parameter_description ## ]]
{parameter_description}

[[ ## parameter_type ## ]]
{parameter_type}

[[ ## function_name ## ]]
{function_name}

[[ ## function_description ## ]]
{function_description}

[[ ## other_parameter_values ## ]]
{other_parameter_values}

[[ ## parallel_context_parameters ## ]]\n{parallel_context_parameters}

[[ ## existing_values ## ]]
{existing_values}

[[ ## parameter_group_context ## ]]
{parameter_group_context}

[[ ## previous_failures ## ]]
{previous_failures}

[[ ## reasoning ## ]]
{reasoning}

[[ ## generated_value ## ]]
{generated_value}

[[ ## completed ## ]]

In adhering to this structure, your objective is: 
Generate a string parameter value that logically fits with other APIs in 
a parallel invocation.

This signature prioritizes logical cohesion with already-generated 
parameters from other APIs in the same parallel context. Diversity is a 
secondary consideration.
\end{verbatim}
\end{tiny}

\subsubsection{Generate Cohesive Other Parameter}

\paragraph{Objective:} Generate a complex/other parameter value that logically fits with other APIs in a parallel invocation.

\paragraph{System Prompt:}
\begin{tiny}
\begin{verbatim}
Your input fields are:
1. `parameter_name` (str): Name of the parameter
2. `parameter_description` (str): Description of the parameter
3. `parameter_type` (str): The parameter's data type
4. `function_name` (str): Function this parameter belongs to
5. `function_description` (str): Description of what this function does
6. `other_parameter_values` (str): The value of the other parameters for 
   this function that were already determined
7. `parallel_context_parameters` (str): JSON array of parameter 
   dictionaries from other APIs in this parallel invocation. These APIs 
   will be called together, so parameters should be logically related. 
   When appropriate, use identical values (e.g., same location, date, or 
   identifier).
8. `existing_values` (str): Previously generated values (for context only)
9. `parameter_group_context` (str): Related parameters context
10. `previous_failures` (str): Information about previous generation 
    attempts that failed. Format: 'None' if no previous failures, otherwise 
    a description of what went wrong.

Your output fields are:
1. `generated_value` (str): A parameter value that logically fits with the 
   parallel context parameters. PRIORITY 1: The value should be logically 
   related and coherent with parameters from other APIs (consider using 
   identical values when appropriate). PRIORITY 2: The value should be 
   feasible and match the parameter description. Diversity from existing 
   values is encouraged but secondary. ONLY return the parameter value, no 
   quotes, no explanation

All interactions will be structured in the following way, with the 
appropriate values filled in.

[[ ## parameter_name ## ]]
{parameter_name}

[[ ## parameter_description ## ]]
{parameter_description}

[[ ## parameter_type ## ]]
{parameter_type}

[[ ## function_name ## ]]
{function_name}

[[ ## function_description ## ]]
{function_description}

[[ ## other_parameter_values ## ]]
{other_parameter_values}

[[ ## parallel_context_parameters ## ]]
{parallel_context_parameters}

[[ ## existing_values ## ]]
{existing_values}

[[ ## parameter_group_context ## ]]
{parameter_group_context}

[[ ## previous_failures ## ]]
{previous_failures}

[[ ## generated_value ## ]]
{generated_value}

[[ ## completed ## ]]

In adhering to this structure, your objective is: 
Generate a complex/other parameter value that logically fits with other 
APIs in a parallel invocation.

This signature prioritizes logical cohesion with already-generated 
parameters from other APIs in the same parallel context. Diversity is a 
secondary consideration.
\end{verbatim}
\end{tiny}

\subsubsection{Generate Other Parameter}

\paragraph{Objective:} Generate a feasible value for complex/other parameter types.

\paragraph{System Prompt:}
\begin{tiny}
\begin{verbatim}
Your input fields are:
1. `parameter_name` (str): Name of the parameter
2. `parameter_description` (str): Description of the parameter
3. `parameter_type` (str): The parameter's data type
4. `function_name` (str): Function this parameter belongs to
5. `function_description` (str): Description of what this function does
6. `other_parameter_values` (str): The value of the other parameters for 
   this function that were already determined
7. `existing_values` (str): Previously generated values
8. `parameter_group_context` (str): Related parameters context
9. `previous_failures` (str): Information about previous generation 
   attempts that failed, including error messages and validation feedback. 
   Use this to avoid repeating the same mistakes. Format: 'None' if no 
   previous failures, otherwise a description of what went wrong.

Your output fields are:
1. `generated_value` (str): A feasible parameter value. The value should 
   be feasible, match the parameter description, make sense in context of 
   the function, and be as different as possible from existing values. 
   ONLY return the parameter value, no quotes, no explanation

All interactions will be structured in the following way, with the 
appropriate values filled in.

[[ ## parameter_name ## ]]
{parameter_name}

[[ ## parameter_description ## ]]
{parameter_description}

[[ ## parameter_type ## ]]
{parameter_type}

[[ ## function_name ## ]]
{function_name}

[[ ## function_description ## ]]
{function_description}

[[ ## other_parameter_values ## ]]
{other_parameter_values}

[[ ## existing_values ## ]]
{existing_values}

[[ ## parameter_group_context ## ]]
{parameter_group_context}

[[ ## previous_failures ## ]]
{previous_failures}

[[ ## generated_value ## ]]
{generated_value}

[[ ## completed ## ]]

In adhering to this structure, your objective is: 
Generate a feasible value for complex/other parameter types.
\end{verbatim}
\end{tiny}

\subsubsection{Generate Multiple Numerical Parameters}

\paragraph{Objective:} Generate multiple diverse and feasible numerical parameter values in one call.

\paragraph{System Prompt:}
\begin{tiny}
\begin{verbatim}
Your input fields are:
1. `parameter_name` (str): Name of the parameter
2. `parameter_description` (str): Description of what this parameter does
3. `parameter_type` (str): Data type (int, float, etc.)
4. `function_name` (str): Function this parameter belongs to
5. `function_description` (str): Description of what this function does
6. `other_parameter_values` (str): The value of the other parameters for 
   this function that were already determined
7. `existing_values` (str): Previously generated numerical values
8. `parameter_group_context` (str): Information about related parameters
9. `num_candidates` (str): Number of diverse candidates to generate
10. `previous_failures` (str): Information about previous generation 
    attempts that failed, including error messages and validation feedback. 
    Use this to avoid repeating the same mistakes. Format: 'None' if no 
    previous failures, otherwise a description of what went wrong.

Your output fields are:
1. `reasoning` (str): 
2. `generated_values` (str): A JSON list of diverse numerical values. Each 
   should be feasible, match the parameter description, make sense in 
   context of the function, and be as different as possible from existing 
   values and each other. Format: [value1, value2, value3, value4, value5] 
   (numbers without quotes). ONLY return the JSON list, no explanation.

All interactions will be structured in the following way, with the 
appropriate values filled in.

[[ ## parameter_name ## ]]
{parameter_name}

[[ ## parameter_description ## ]]
{parameter_description}

[[ ## parameter_type ## ]]
{parameter_type}

[[ ## function_name ## ]]
{function_name}

[[ ## function_description ## ]]
{function_description}

[[ ## other_parameter_values ## ]]
{other_parameter_values}

[[ ## existing_values ## ]]
{existing_values}

[[ ## parameter_group_context ## ]]
{parameter_group_context}

[[ ## num_candidates ## ]]
{num_candidates}

[[ ## previous_failures ## ]]
{previous_failures}

[[ ## reasoning ## ]]
{reasoning}

[[ ## generated_values ## ]]
{generated_values}

[[ ## completed ## ]]

In adhering to this structure, your objective is: 
Generate multiple diverse and feasible numerical parameter values in one 
call.
\end{verbatim}
\end{tiny}

\subsubsection{Generate Sequential Cohesive Other Parameter}

\paragraph{Objective:} Generate a complex/other parameter value for sequential API execution.

\paragraph{System Prompt:}
\begin{tiny}
\begin{verbatim}
Your input fields are:
1. `parameter_name` (str): Name of the parameter
2. `parameter_description` (str): Description of the parameter
3. `parameter_type` (str): The parameter's data type
4. `function_name` (str): Function this parameter belongs to
5. `function_description` (str): Description of what this function does
6. `other_parameter_values` (str): The value of the other parameters for 
   this function that were already determined
7. `return_value` (str): JSON string of the return value that this API 
   will produce. This is the MOST IMPORTANT context - the parameter should 
   be cohesive with this return value. The return value will be used as 
   input for the next API in the sequential chain.
8. `next_api_parameters` (str): JSON string of parameters for the next API 
   in the sequential chain. The return value of this API should provide 
   information needed by the next API's parameters.
9. `later_api_parameters` (str): JSON array of parameter dictionaries from 
   APIs later in the sequential chain. These provide additional context 
   for cohesion.
10. `existing_values` (str): Previously generated values (for context only)
11. `parameter_group_context` (str): Related parameters context
12. `previous_failures` (str): Information about previous generation 
    attempts that failed. Format: 'None' if no previous failures, otherwise 
    a description of what went wrong.

Your output fields are:
1. `generated_value` (str): A parameter value that is cohesive with the 
   return value and sequential chain. PRIORITY 1: The value should be 
   logically related to the return value this API will produce. PRIORITY 2: 
   The value should be cohesive with parameters of the next API in the 
   chain. PRIORITY 3: The value should be feasible and match the parameter 
   description. Diversity from existing values is encouraged but secondary. 
   ONLY return the parameter value, no quotes, no explanation

All interactions will be structured in the following way, with the 
appropriate values filled in.

[[ ## parameter_name ## ]]
{parameter_name}

[[ ## parameter_description ## ]]
{parameter_description}

[[ ## parameter_type ## ]]
{parameter_type}

[[ ## function_name ## ]]
{function_name}

[[ ## function_description ## ]]
{function_description}

[[ ## other_parameter_values ## ]]
{other_parameter_values}

[[ ## return_value ## ]]
{return_value}

[[ ## next_api_parameters ## ]]
{next_api_parameters}

[[ ## later_api_parameters ## ]]
{later_api_parameters}

[[ ## existing_values ## ]]
{existing_values}

[[ ## parameter_group_context ## ]]
{parameter_group_context}

[[ ## previous_failures ## ]]
{previous_failures}

[[ ## generated_value ## ]]
{generated_value}

[[ ## completed ## ]]

In adhering to this structure, your objective is: 
Generate a complex/other parameter value for sequential API execution.

This signature prioritizes logical cohesion with:
1. The return value of this API (MOST IMPORTANT - this API's output will 
   be used by next API)
2. Parameters of the next API in the chain
3. Parameters of APIs later in the chain
\end{verbatim}
\end{tiny}

\subsubsection{Generate Cohesive Numerical Parameter}

\paragraph{Objective:} Generate a numerical parameter value that logically fits with other APIs in a parallel invocation.

\paragraph{System Prompt:}
\begin{tiny}
\begin{verbatim}
Your input fields are:
1. `parameter_name` (str): Name of the parameter
2. `parameter_description` (str): Description of what this parameter does
3. `parameter_type` (str): Data type (int, float, etc.)
4. `function_name` (str): Function this parameter belongs to
5. function_description (str): Description of what this function does
6. other_parameter_values (str): The value of the other parameters for
this function that were already determined
7. parallel_context_parameters (str): JSON array of parameter
dictionaries from other APIs in this parallel invocation. These APIs
will be called together, so parameters should be logically related.
When appropriate, use identical values (e.g., same location, date, or
identifier).
8. existing_values (str): Previously generated numerical values (for
context only)
9. parameter_group_context (str): Information about related parameters
10. previous_failures (str): Information about previous generation
attempts that failed. Format: 'None' if no previous failures, otherwise
a description of what went wrong.
Your output fields are:

reasoning (str):
generated_value (str): A numerical parameter value that logically
fits with the parallel context parameters. PRIORITY 1: The value should
be logically related and coherent with parameters from other APIs
(consider using identical values when appropriate, e.g., same location,
date, ID). PRIORITY 2: The value should be feasible and match the
parameter description. Diversity from existing values is encouraged but
secondary. ONLY return the numerical value, no quotes, no explanation.

All interactions will be structured in the following way, with the
appropriate values filled in.
[[ ## parameter_name ## ]]
{parameter_name}
[[ ## parameter_description ## ]]
{parameter_description}
[[ ## parameter_type ## ]]
{parameter_type}
[[ ## function_name ## ]]
{function_name}
[[ ## function_description ## ]]
{function_description}
[[ ## other_parameter_values ## ]]
{other_parameter_values}
[[ ## parallel_context_parameters ## ]]
{parallel_context_parameters}
[[ ## existing_values ## ]]
{existing_values}
[[ ## parameter_group_context ## ]]
{parameter_group_context}
[[ ## previous_failures ## ]]
{previous_failures}
[[ ## reasoning ## ]]
{reasoning}
[[ ## generated_value ## ]]
{generated_value}
[[ ## completed ## ]]
In adhering to this structure, your objective is:
Generate a numerical parameter value that logically fits with other APIs
in a parallel invocation.
This signature prioritizes logical cohesion with already-generated
parameters from other APIs in the same parallel context. Diversity is a
secondary consideration.
\end{verbatim}
\end{tiny}

\subsubsection{Parameter Set Validator}
\paragraph{Objective:} Validate that a set of parameter values forms a coherent, conflict-free API call.
\paragraph{System Prompt:}
\begin{tiny}
\begin{verbatim}
Your input fields are:

api_name (str): Name of the API function
api_description (str): Description of what this API does
full_parameter_schema (str): Complete parameter schema including all
parameters (required and optional) with their descriptions
selected_parameters (str): The specific parameters that were selected
for this API call (JSON dict with param names and values)

Your output fields are:

reasoning (str): Step-by-step analysis of whether this parameter
combination is valid. Check for: (1) missing required dependencies,
(2) conflicting parameters, (3) parameters that don't make sense
together, (4) invalid combinations, (5) parameter types that don't
match the schema.
is_valid (str): YES if this parameter set is valid and conflict-free,
NO if there are issues

All interactions will be structured in the following way, with the
appropriate values filled in.
[[ ## api_name ## ]]
{api_name}
[[ ## api_description ## ]]
{api_description}
[[ ## full_parameter_schema ## ]]
{full_parameter_schema}
[[ ## selected_parameters ## ]]
{selected_parameters}
[[ ## reasoning ## ]]
{reasoning}
[[ ## is_valid ## ]]
{is_valid}
[[ ## completed ## ]]
In adhering to this structure, your objective is:
Validate that a set of parameter values forms a coherent, conflict-free
API call.
\end{verbatim}
\end{tiny}

\subsubsection{Partial Parameter Set Validator}

\paragraph{Objective:} Validate that a partial set of parameter values forms a coherent, conflict-free combination.

\paragraph{System Prompt:}
\begin{tiny}
\begin{verbatim}
Your input fields are:
1. `api_name` (str): Name of the API function
2. `api_description` (str): Description of what this API does
3. `full_parameter_schema` (str): Complete parameter schema including all 
   parameters (required and optional) with their descriptions
4. `provided_parameters` (str): The specific parameters that are provided 
   for this API call (JSON dict with param names and values). Note: Some 
   required parameters may be intentionally missing.

Your output fields are:
1. `reasoning` (str): Step-by-step analysis of whether the provided 
   parameter combination is valid. Check for: (1) conflicting parameters, 
   (2) parameters that don't make sense together, (3) invalid combinations, 
   (4) parameter types that don't match the schema. DO NOT reject the 
   parameter set just because required parameters are missing - that is 
   expected and intentional. Only validate that the provided parameters 
   are coherent together.
2. `is_valid` (str): YES if the provided parameters form a valid, 
   conflict-free combination (ignoring missing required params), NO if 
   there are issues with the provided parameters themselves

All interactions will be structured in the following way, with the 
appropriate values filled in.

[[ ## api_name ## ]]
{api_name}

[[ ## api_description ## ]]
{api_description}

[[ ## full_parameter_schema ## ]]
{full_parameter_schema}

[[ ## provided_parameters ## ]]
{provided_parameters}

[[ ## reasoning ## ]]
{reasoning}

[[ ## is_valid ## ]]
{is_valid}

[[ ## completed ## ]]

In adhering to this structure, your objective is: 
Validate that a partial set of parameter values forms a coherent, 
conflict-free combination.

This validator is used when some required parameters are intentionally 
missing. It validates that the provided parameters are valid together, 
without checking for missing required parameters.
\end{verbatim}
\end{tiny}

\subsubsection{Validate Return Value}

\paragraph{Objective:} Validate if a return value is appropriate for an API in a sequential chain.

\paragraph{System Prompt:}
\begin{tiny}
\begin{verbatim}
Your input fields are:
1. `api_name` (str): Name of the API function
2. `api_description` (str): Description of what the API does
3. `return_type_schema` (str): JSON string of the return type schema
4. `return_value` (str): JSON string of the return value to validate
5. `api_parameters` (str): JSON string of the API's parameters (if 
   available). The return value should be logically consistent with these 
   parameters.
6. `next_api_name` (str): Name of the next API in the sequential chain 
   (if any). The return value should be useful as input for this next API.
7. `next_api_description` (str): Description of the next API in the chain 
   (if any)
8. `next_api_parameters` (str): JSON string of the next API's parameter 
   schema (if any). The return value should contain information usable by 
   the next API.

Your output fields are:
1. `reasoning` (str): Step-by-step analysis: (1) Check if return value 
   matches return type schema, (2) Verify logical consistency with API 
   description and parameters, (3) Assess if return value is useful for 
   the next API (if provided), (4) Determine if this is a valid return 
   value.
2. `is_valid` (str): 'YES' if the return value is valid, 'NO' otherwise. 
   Return only 'YES' or 'NO'.

All interactions will be structured in the following way, with the 
appropriate values filled in.

[[ ## api_name ## ]]
{api_name}

[[ ## api_description ## ]]
{api_description}

[[ ## return_type_schema ## ]]
{return_type_schema}

[[ ## return_value ## ]]
{return_value}

[[ ## api_parameters ## ]]
{api_parameters}

[[ ## next_api_name ## ]]
{next_api_name}

[[ ## next_api_description ## ]]
{next_api_description}

[[ ## next_api_parameters ## ]]
{next_api_parameters}

[[ ## reasoning ## ]]
{reasoning}

[[ ## is_valid ## ]]
{is_valid}

[[ ## completed ## ]]

In adhering to this structure, your objective is: 
Validate if a return value is appropriate for an API in a sequential chain.

The return value should:
1. Match the API's return type schema
2. Be logically consistent with the API's purpose and description
3. Be cohesive with the API's parameters (if provided)
4. Be useful as input for the next API in the chain (if next_api_info is 
   provided)
\end{verbatim}
\end{tiny}

\subsubsection{Validate Sequential Chain}

\paragraph{Objective:} Validate an entire sequential API execution chain for coherence, consistency, and true sequential dependency.

\paragraph{System Prompt:}
\begin{tiny}
\begin{verbatim}
Your input fields are:
1. `api_schemas` (str): JSON array of API schemas for all APIs in the 
   sequential chain, in execution order
2. `parameters_list` (str): JSON array of parameter dictionaries, one per 
   API, in execution order
3. `return_values_list` (str): JSON array of return value dictionaries, 
   one per API (except last), in execution order

Your output fields are:
1. `reasoning` (str): Step-by-step analysis: (1) Check each API's 
   parameters for individual validity and consistency, (2) Check each 
   API's return value for individual validity and consistency, (3) 
   CRITICAL: Verify STRICT sequential dependency - for each API (except 
   first), determine if it REQUIRES information from the previous API's 
   return value that cannot be obtained from the query alone. If any API 
   can be called without the previous API's output, the chain is INVALID. 
   (4) Verify that APIs cannot be called in parallel - if all parameters 
   can be inferred from the query, reject. (5) Verify cohesion between 
   return values and subsequent API parameters (return values must provide 
   necessary inputs that subsequent APIs actually need), (6) Check for 
   contradictions across the chain (conflicting values, impossible 
   sequences), (7) Assess overall logical coherence and true sequential 
   dependency of the chain, (8) Determine if the entire chain is valid 
   (must be INVALID if sequential dependency is weak or missing).
2. `is_valid` (str): 'YES' if the entire sequential chain is valid, 
   coherent, and has STRICT sequential dependencies (each API requires the 
   previous API's output). 'NO' if APIs can be called in parallel, if any 
   API doesn't require the previous output, or if the sequential dependency 
   is weak. Return only 'YES' or 'NO'.

All interactions will be structured in the following way, with the 
appropriate values filled in.

[[ ## api_schemas ## ]]
{api_schemas}

[[ ## parameters_list ## ]]
{parameters_list}

[[ ## return_values_list ## ]]
{return_values_list}

[[ ## reasoning ## ]]
{reasoning}

[[ ## is_valid ## ]]
{is_valid}

[[ ## completed ## ]]

In adhering to this structure, your objective is: 
Validate an entire sequential API execution chain for coherence, 
consistency, and true sequential dependency.

CRITICAL REQUIREMENT: This validates a TRUE sequential chain where each 
API REQUIRES the output from the previous API. The chain is INVALID if 
APIs can be called in parallel or if any API doesn't actually need the 
previous API's output.

This validates that:
1. All parameters are sensible on their own (logically consistent, match 
   schemas)
2. All return values are sensible on their own (match return type schemas, 
   logically consistent)
3. STRICT SEQUENTIAL DEPENDENCY: Each API (except the first) REQUIRES 
   information from the previous API's return value that cannot be obtained 
   from the query alone. If an API can be called without the previous API's 
   output, the chain is INVALID.
4. Parameters and return values are cohesive together (return values 
   provide necessary inputs for later APIs)
5. There are no contradictions across the chain (e.g., conflicting values, 
   impossible sequences)
6. The chain makes logical sense as a whole (later APIs genuinely depend 
   on earlier API outputs)

The judge MUST REJECT chains where:
- APIs can be called in parallel (if all APIs' parameters can be inferred 
  from the query alone, reject)
- An API doesn't actually need the previous API's output (if the API can 
  work without the return value, reject)
- The sequential dependency is weak or optional (the dependency must be 
  STRONG and NECESSARY)
- Return values don't provide information that is actually needed by 
  subsequent APIs

The judge should check for:
- Parameter contradictions (e.g., same parameter with conflicting values 
  across APIs)
- Return value contradictions (e.g., return values that don't make sense 
  together)
- Missing sequential dependency (APIs that don't require previous outputs 
  - REJECT)
- Parallel-executable APIs (APIs that can be called together - REJECT)
- Cohesion issues (e.g., return values that can't be used as inputs for 
  later APIs)
- Logical inconsistencies (e.g., impossible sequences, values that don't 
  match expected types)
\end{verbatim}
\end{tiny}

\subsection{Query Generation Prompts}

\subsubsection{No API Query Generator}

\paragraph{Objective:} Generate natural language queries that don't require any API calls.

\paragraph{System Prompt:}
\begin{tiny}
\begin{verbatim}
Your input fields are:
1. `dataset_guidance` (str): guidance on generating diverse queries based 
   on dataset patterns and gaps
2. `previous_attempts` (str): previous attempts with scores/judgements for 
   improvement

Your output fields are:
1. `reasoning` (str): 
2. `query_1` (str): first diverse natural language query answerable 
   without any API calls
3. `query_2` (str): second diverse natural language query answerable 
   without any API calls
4. `query_3` (str): third diverse natural language query answerable 
   without any API calls
5. `query_4` (str): fourth diverse natural language query answerable 
   without any API calls
6. `query_5` (str): fifth diverse natural language query answerable 
   without any API calls

All interactions will be structured in the following way, with the 
appropriate values filled in.

[[ ## dataset_guidance ## ]]
{dataset_guidance}

[[ ## previous_attempts ## ]]
{previous_attempts}

[[ ## reasoning ## ]]
{reasoning}

[[ ## query_1 ## ]]
{query_1}

[[ ## query_2 ## ]]
{query_2}

[[ ## query_3 ## ]]
{query_3}

[[ ## query_4 ## ]]
{query_4}

[[ ## query_5 ## ]]
{query_5}

[[ ## completed ## ]]

In adhering to this structure, your objective is: 
Generate natural language queries that don't require any API calls.
\end{verbatim}
\end{tiny}

\subsubsection{Sequential Query Generator}

\paragraph{Objective:} Generate natural language queries for sequential API calls.

\paragraph{System Prompt:}
\begin{tiny}
\begin{verbatim}
Your input fields are:
1. `api_schemas` (str): JSON array of API schemas (name, description, 
   parameters, returns) in execution order. Later APIs may use return 
   values from earlier APIs as input parameters. Use this to understand 
   what information is available, but DO NOT mention the sequence in the 
   query.
2. `target_parameters_list` (str): JSON array of parameter dictionaries, 
   one per API, in same order as api_schemas. These are the parameter 
   values that will be used. The query should provide information for 
   parameters that come from the query itself (not from previous API 
   outputs).
3. `return_values_list` (str): JSON array of return value dictionaries, 
   one per API (except last), in same order as api_schemas. Each return 
   value represents what the API would return. Use this to understand what 
   information becomes available, but DO NOT mention these intermediate 
   results in the query.
4. `dataset_guidance` (str): sample queries from dataset to guide diversity
5. `previous_attempts` (str): previous attempts with scores/judgements for 
   improvement

Your output fields are:
1. `reasoning` (str): 
2. `query_1` (str): First natural language query that describes the END 
   GOAL, not the sequence. The query should be moderately challenging - 
   the model needs to figure out the sequence, but the goal should be 
   clear. Provide information for parameters that come from the query.
3. `query_2` (str): Second natural language query that describes the END 
   GOAL, not the sequence. The query should be moderately challenging - 
   the model needs to figure out the sequence, but the goal should be 
   clear. Provide information for parameters that come from the query.
4. `query_3` (str): Third natural language query that describes the END 
   GOAL, not the sequence. The query should be moderately challenging - 
   the model needs to figure out the sequence, but the goal should be 
   clear. Provide information for parameters that come from the query.
5. `query_4` (str): Fourth natural language query that describes the END 
   GOAL, not the sequence. The query should be moderately challenging - 
   the model needs to figure out the sequence, but the goal should be 
   clear. Provide information for parameters that come from the query.
6. `query_5` (str): Fifth natural language query that describes the END 
   GOAL, not the sequence. The query should be moderately challenging - 
   the model needs to figure out the sequence, but the goal should be 
   clear. Provide information for parameters that come from the query.

All interactions will be structured in the following way, with the 
appropriate values filled in.

[[ ## api_schemas ## ]]
{api_schemas}

[[ ## target_parameters_list ## ]]
{target_parameters_list}

[[ ## return_values_list ## ]]
{return_values_list}

[[ ## dataset_guidance ## ]]
{dataset_guidance}

[[ ## previous_attempts ## ]]
{previous_attempts}

[[ ## reasoning ## ]]
{reasoning}

[[ ## query_1 ## ]]
{query_1}

[[ ## query_2 ## ]]
{query_2}

[[ ## query_3 ## ]]
{query_3}

[[ ## query_4 ## ]]
{query_4}

[[ ## query_5 ## ]]
{query_5}

[[ ## completed ## ]]

In adhering to this structure, your objective is: 
Generate natural language queries for sequential API calls.

CRITICAL: Generate queries that describe the END GOAL or desired outcome, 
NOT the sequence of steps. The query should be moderately challenging - 
the model should need to figure out the sequence of API calls to achieve 
the goal, but the goal itself should be clear and inferrable.

DO NOT:
- Explicitly mention the sequence of API calls (e.g., "first do X, then Y, 
  then Z")
- Tell the model what steps to take (e.g., "call API A, then use its 
  output for API B")
- Spoonfeed the execution plan (e.g., "I want you to do step A, then step 
  B, and finally step C")
- Use instructional language that reveals the sequence (e.g., "start by 
  doing X, then do Y")

DO:
- Describe what the user wants to know or achieve (the end goal)
- Provide sufficient information for parameters that come from the query 
  (not from previous API outputs)
- Make the query natural and conversational
- Let the model infer that sequential API calls are needed to achieve the 
  goal
- Make it clear what the final desired information is

Example of BAD query: "I want information on X, so do step A, then B, and 
finally C to derive info D on X"
Example of GOOD query: "What is the weather in the capital of France?" 
(model infers: get capital, then get weather)
\end{verbatim}
\end{tiny}

\subsubsection{Parallel Query Generator}

\paragraph{Objective:} Generate multiple diverse queries for parallel API calls (multiple APIs called together).

\paragraph{System Prompt:}
\begin{tiny}
\begin{verbatim}
Your input fields are:
1. `api_schemas` (str): JSON array of API schemas (name, description, 
   parameters) for all APIs to call in parallel
2. `target_parameters_list` (str): JSON array of parameter dictionaries, 
   one per API, in same order as api_schemas
3. `dataset_guidance` (str): sample queries from dataset to guide diversity
4. `previous_attempts` (str): previous attempts with scores/judgements for 
   improvement

Your output fields are:
1. `reasoning` (str): 
2. `query_1` (str): first diverse natural language query requiring 
   parallel API calls
3. `query_2` (str): second diverse natural language query requiring 
   parallel API calls
4. `query_3` (str): third diverse natural language query requiring 
   parallel API calls
5. `query_4` (str): fourth diverse natural language query requiring 
   parallel API calls
6. `query_5` (str): fifth diverse natural language query requiring 
   parallel API calls

All interactions will be structured in the following way, with the 
appropriate values filled in.

[[ ## api_schemas ## ]]
{api_schemas}

[[ ## target_parameters_list ## ]]
{target_parameters_list}

[[ ## dataset_guidance ## ]]
{dataset_guidance}

[[ ## previous_attempts ## ]]
{previous_attempts}

[[ ## reasoning ## ]]
{reasoning}

[[ ## query_1 ## ]]
{query_1}

[[ ## query_2 ## ]]
{query_2}

[[ ## query_3 ## ]]
{query_3}

[[ ## query_4 ## ]]
{query_4}

[[ ## query_5 ## ]]
{query_5}

[[ ## completed ## ]]

In adhering to this structure, your objective is: 
Generate multiple diverse queries for parallel API calls (multiple APIs 
called together).
\end{verbatim}
\end{tiny}

\subsubsection{Multi Query Generator}

\paragraph{Objective:} Generate multiple diverse queries for a specific API call.

\paragraph{System Prompt:}
\begin{tiny}
\begin{verbatim}
Your input fields are:
1. `api_schema` (str): API name, description, and parameter schema
2. `target_parameters` (str): specific parameter values to target
3. `dataset_guidance` (str): sample queries from dataset to guide diversity
4. `previous_attempts` (str): previous attempts with scores/judgements for 
   improvement

Your output fields are:
1. `reasoning` (str): 
2. `query_1` (str): first diverse natural language query
3. `query_2` (str): second diverse natural language query
4. `query_3` (str): third diverse natural language query
5. `query_4` (str): fourth diverse natural language query
6. `query_5` (str): fifth diverse natural language query

All interactions will be structured in the following way, with the 
appropriate values filled in.

[[ ## api_schema ## ]]
{api_schema}

[[ ## target_parameters ## ]]
{target_parameters}

[[ ## dataset_guidance ## ]]
{dataset_guidance}

[[ ## previous_attempts ## ]]
{previous_attempts}

[[ ## reasoning ## ]]
{reasoning}

[[ ## query_1 ## ]]
{query_1}

[[ ## query_2 ## ]]
{query_2}

[[ ## query_3 ## ]]
{query_3}

[[ ## query_4 ## ]]
{query_4}

[[ ## query_5 ## ]]
{query_5}

[[ ## completed ## ]]

In adhering to this structure, your objective is: 
Generate multiple diverse queries for a specific API call.
\end{verbatim}
\end{tiny}

\subsubsection{Missing Params Query Generator}

\paragraph{Objective:} Generate multiple diverse queries for a specific API call where some required parameters are missing.

\paragraph{System Prompt:}
\begin{tiny}
\begin{verbatim}
Your input fields are:
1. `api_schema` (str): API name, description, and parameter schema
2. `provided_parameters` (str): parameter values that are provided in the 
   query
3. `missing_parameters` (str): list of required parameter names that are 
   intentionally missing
4. `dataset_guidance` (str): sample queries from dataset to guide diversity
5. `previous_attempts` (str): previous attempts with scores/judgements for 
   improvement

Your output fields are:
1. `reasoning` (str): 
2. `query_1` (str): first diverse natural language query that requires 
   this API but lacks the specified required parameters
3. `query_2` (str): second diverse natural language query that requires 
   this API but lacks the specified required parameters
4. `query_3` (str): third diverse natural language query that requires 
   this API but lacks the specified required parameters
5. `query_4` (str): fourth diverse natural language query that requires 
   this API but lacks the specified required parameters
6. `query_5` (str): fifth diverse natural language query that requires 
   this API but lacks the specified required parameters

All interactions will be structured in the following way, with the 
appropriate values filled in.

[[ ## api_schema ## ]]
{api_schema}

[[ ## provided_parameters ## ]]
{provided_parameters}

[[ ## missing_parameters ## ]]
{missing_parameters}

[[ ## dataset_guidance ## ]]
{dataset_guidance}

[[ ## previous_attempts ## ]]
{previous_attempts}

[[ ## reasoning ## ]]
{reasoning}

[[ ## query_1 ## ]]
{query_1}

[[ ## query_2 ## ]]
{query_2}

[[ ## query_3 ## ]]
{query_3}

[[ ## query_4 ## ]]
{query_4}

[[ ## query_5 ## ]]
{query_5}

[[ ## completed ## ]]

In adhering to this structure, your objective is: 
Generate multiple diverse queries for a specific API call where some 
required parameters are missing.
\end{verbatim}
\end{tiny}

\subsubsection{Sequential Query Judge}

\paragraph{Objective:} Judge if a query reasonably justifies calling multiple APIs sequentially with given parameters and return values.

\paragraph{System Prompt:}
\begin{tiny}
\begin{verbatim}
Your input fields are:
1. `query` (str): user query to evaluate
2. `api_schemas` (str): JSON array of API schemas for all target APIs in 
   execution order
3. `target_parameters_list` (str): JSON array of parameter dictionaries, 
   one per API
4. `return_values_list` (str): JSON array of return value dictionaries, 
   one per API (except last), representing what each API returns

Your output fields are:
1. `reasoning` (str): Step-by-step analysis that MUST include: (1) For 
   each API in the chain, verify that it is REQUIRED and not redundant. 
   (2) For the FIRST API, check if the query contains enough information 
   to infer all parameters. (3) For each SUBSEQUENT API, check if 
   parameters can be inferred from either the query OR return values of 
   earlier APIs in the chain. (4) Assess whether the sequential execution 
   chain is logically appropriate - verify that later APIs logically use 
   outputs from earlier APIs. The query is NOT reasonable if ANY API is 
   redundant, if the first API's parameters cannot be inferred from the 
   query, if later APIs' parameters cannot be inferred from query/earlier 
   returns, or if the chain is illogical.
2. `is_reasonable` (str): YES only if: (1) ALL APIs are required (no 
   redundant APIs), (2) The query contains enough information to infer ALL 
   parameters for the FIRST API, (3) Parameters for SUBSEQUENT APIs can be 
   inferred from the query and/or return values of earlier APIs, AND (4) 
   The sequential execution chain is logically appropriate. NO if any API 
   is redundant, if parameters cannot be inferred appropriately, or if the 
   chain is illogical. Return YES or NO.

All interactions will be structured in the following way, with the 
appropriate values filled in.

[[ ## query ## ]]
{query}

[[ ## api_schemas ## ]]
{api_schemas}

[[ ## target_parameters_list ## ]]
{target_parameters_list}

[[ ## return_values_list ## ]]
{return_values_list}

[[ ## reasoning ## ]]
{reasoning}

[[ ## is_reasonable ## ]]
{is_reasonable}

[[ ## completed ## ]]

In adhering to this structure, your objective is: 
Judge if a query reasonably justifies calling multiple APIs sequentially 
with given parameters and return values.

IMPORTANT: Sequential execution means APIs are called IN ORDER, where 
later APIs can use return values from earlier APIs as input parameters. 
The first API's parameters come from the query, and subsequent APIs may 
have parameters inferred from previous API outputs.

The judge must verify that:
1. ALL APIs are required - no API is redundant or unnecessary
2. The query contains enough information to infer parameters for the FIRST 
   API directly from the query
3. For subsequent APIs, parameters can be inferred from either the query 
   OR return values of earlier APIs
4. The sequential execution chain is logically appropriate (later APIs 
   logically use earlier API outputs)

A query is reasonable if the chain makes sense and parameters can be 
inferred from the query and/or previous API return values.
\end{verbatim}
\end{tiny}

\subsubsection{Parallel Query Judge}

\paragraph{Objective:} Judge if a query reasonably justifies calling multiple APIs in parallel with given parameters.

\paragraph{System Prompt:}
\begin{tiny}
\begin{verbatim}
Your input fields are:
1. `query` (str): user query to evaluate
2. `api_schemas` (str): JSON array of API schemas for all target APIs
3. `target_parameters_list` (str): JSON array of parameter dictionaries, 
   one per API

Your output fields are:
1. `reasoning` (str): Step-by-step analysis that MUST include: (1) For 
   each API, verify that it is REQUIRED and not redundant - check if 
   removing it would make the query incomplete or unanswerable. (2) For 
   each API, check if the query contains enough information to infer each 
   parameter value DIRECTLY from the query text. Remember: in parallel 
   execution, APIs execute simultaneously, so NO parameter can be inferred 
   from another API's output. All parameters must come from the query 
   alone. Identify any parameters that cannot be inferred from the query. 
   (3) Assess whether the parallel execution is logically appropriate - 
   verify that the APIs can work together without dependencies (no API 
   needs another API's output). The query is NOT reasonable if ANY API is 
   redundant, if ANY parameter cannot be inferred from the query text 
   alone, or if APIs have sequential dependencies.
2. `is_reasonable` (str): YES only if: (1) ALL APIs are required (no 
   redundant APIs), (2) The query contains enough information to infer ALL 
   parameters in target_parameters_list for ALL APIs DIRECTLY from the 
   query text (not from other API outputs, since they execute in parallel), 
   AND (3) The parallel execution is logically appropriate (no sequential 
   dependencies). NO if any API is redundant, if any parameter value 
   cannot be inferred from the query text alone, if APIs have sequential 
   dependencies, or if parallel execution doesn't make sense. Return YES 
   or NO.

All interactions will be structured in the following way, with the 
appropriate values filled in.

[[ ## query ## ]]
{query}

[[ ## api_schemas ## ]]
{api_schemas}

[[ ## target_parameters_list ## ]]
{target_parameters_list}

[[ ## reasoning ## ]]
{reasoning}

[[ ## is_reasonable ## ]]
{is_reasonable}

[[ ## completed ## ]]

In adhering to this structure, your objective is: 
Judge if a query reasonably justifies calling multiple APIs in parallel 
with given parameters.

IMPORTANT: Parallel execution means all APIs are called SIMULTANEOUSLY 
with the same query. In parallel calls, APIs execute at the same time and 
cannot depend on each other's outputs. No parameter value can be inferred 
from the output of another API call - all parameters must be inferable 
directly from the query text alone.

The judge must verify that:
1. ALL APIs are required - no API is redundant or unnecessary
2. The query contains enough information to infer ALL parameters for each 
   API directly from the query (NOT from other API outputs, since they 
   execute simultaneously)
3. The parallel execution is logically appropriate (APIs can work together 
   without dependencies)

A query is only reasonable if all APIs are necessary and all parameter 
values can be inferred from the query text alone, without needing outputs 
from other APIs.
\end{verbatim}
\end{tiny}

\subsubsection{API Query Judge}

\paragraph{Objective:} Judge if a query reasonably justifies calling a specific API with given parameters.

\paragraph{System Prompt:}
\begin{tiny}
\begin{verbatim}
Your input fields are:
1. `query` (str): user query to evaluate
2. `api_schema` (str): API specification
3. `target_parameters` (str): proposed API parameters

Your output fields are:
1. `reasoning` (str): step-by-step analysis of appropriateness
2. `is_reasonable` (str): YES or NO - is this API call reasonable for the 
   query

All interactions will be structured in the following way, with the 
appropriate values filled in.

[[ ## query ## ]]
{query}

[[ ## api_schema ## ]]
{api_schema}

[[ ## target_parameters ## ]]
{target_parameters}

[[ ## reasoning ## ]]
{reasoning}

[[ ## is_reasonable ## ]]
{is_reasonable}

[[ ## completed ## ]]

In adhering to this structure, your objective is: 
Judge if a query reasonably justifies calling a specific API with given 
parameters.
\end{verbatim}
\end{tiny}

\subsubsection{Missing Params Query Judge}

\paragraph{Objective:} Judge if a query reasonably requires a specific API but is missing some required parameters.

\paragraph{System Prompt:}
\begin{tiny}
\begin{verbatim}
Your input fields are:
1. `query` (str): user query to evaluate
2. `api_schema` (str): API specification
3. `provided_parameters` (str): parameter values that are provided in the 
   query
4. `missing_parameters` (str): list of required parameter names that 
   should be missing

Your output fields are:
1. `reasoning` (str): step-by-step analysis of appropriateness and missing 
   parameters
2. `is_reasonable` (str): YES or NO - does this query require this API but 
   lack the specified required parameters

All interactions will be structured in the following way, with the 
appropriate values filled in.

[[ ## query ## ]]
{query}

[[ ## api_schema ## ]]
{api_schema}

[[ ## provided_parameters ## ]]
{provided_parameters}

[[ ## missing_parameters ## ]]
{missing_parameters}

[[ ## reasoning ## ]]
{reasoning}

[[ ## is_reasonable ## ]]
{is_reasonable}

[[ ## completed ## ]]

In adhering to this structure, your objective is: 
Judge if a query reasonably requires a specific API but is missing some 
required parameters.
\end{verbatim}

\end{tiny}

\subsubsection{Dataset Pattern Analysis}

\paragraph{Objective:} Analyze a sample of queries to identify patterns of homogeneity.

\paragraph{System Prompt:}
\begin{tiny}
\begin{verbatim}
Your input fields are:
1. `dataset_sample` (str): A sample of queries from the current dataset 
   (newline-separated)
2. `diversity_context` (str): Information about diversity metrics and what 
   dimensions matter for this dataset

Your output fields are:
1. `reasoning` (str): 
2. `pattern_analysis` (str): Detailed identification of homogeneity 
   patterns. For each pattern: (1) describe it concretely with examples 
   from the sample, (2) explain why it reduces diversity, (3) estimate how 
   prevalent it is in the sample. Focus on: semantic clustering, pragmatic 
   patterns, implicit assumptions, discourse structures, persona/voice 
   homogeneity, and contextual patterns. Be specific and actionable.

All interactions will be structured in the following way, with the 
appropriate values filled in.

[[ ## dataset_sample ## ]]
{dataset_sample}

[[ ## diversity_context ## ]]
{diversity_context}

[[ ## reasoning ## ]]
{reasoning}

[[ ## pattern_analysis ## ]]
{pattern_analysis}

[[ ## completed ## ]]

In adhering to this structure, your objective is: 
Analyze a sample of queries to identify patterns of homogeneity.
\end{verbatim}
\end{tiny}

\subsubsection{Diversity Guidance Generation}

\paragraph{Objective:} Generate actionable diversity guidance based on pattern analysis.

\paragraph{System Prompt:}
\begin{tiny}
\begin{verbatim}
Your input fields are:
1. `dataset_sample` (str): Sample queries from current dataset
2. `pattern_analysis` (str): Analysis of homogeneity patterns identified 
   in the sample

Your output fields are:
1. `reasoning` (str): 
2. `diversity_guidance` (str): Concrete, actionable guidance for 
   generating diverse queries. Should include: (1) specific patterns to 
   avoid (from analysis), (2) concrete contrasts to explore (opposite of 
   identified patterns), (3) dimensions to vary (semantic, syntactic, 
   pragmatic, stylistic), (4) example directions that would increase 
   diversity. Be specific and directive, not vague.

All interactions will be structured in the following way, with the 
appropriate values filled in.

[[ ## dataset_sample ## ]]
{dataset_sample}

[[ ## pattern_analysis ## ]]
{pattern_analysis}

[[ ## reasoning ## ]]
{reasoning}

[[ ## diversity_guidance ## ]]
{diversity_guidance}

[[ ## completed ## ]]

In adhering to this structure, your objective is: 
Generate actionable diversity guidance based on pattern analysis.
\end{verbatim}
\end{tiny}

\subsection{Distractor Selection Prompts}

\subsubsection{Batch API Relevance Scorer}

\paragraph{Objective:} Score multiple API candidates for relevance to a query.

\paragraph{System Prompt:}
\begin{tiny}
\begin{verbatim}
Your input fields are:
1. `query` (str): User's natural language query
2. `apis` (str): JSON array of API candidates with name, description, 
   parameters
3. `target_api` (str): Name of the ground truth target API (for reference)

Your output fields are:
1. `reasoning` (str): 
2. `scores` (str): JSON array of scores. For each API provide: 
   {"api_name": str, "score": int (1-5), "reasoning": str}. Maintain same 
   order as input APIs.

All interactions will be structured in the following way, with the 
appropriate values filled in.

[[ ## query ## ]]
{query}

[[ ## apis ## ]]
{apis}

[[ ## target_api ## ]]
{target_api}

[[ ## reasoning ## ]]
{reasoning}

[[ ## scores ## ]]
{scores}

[[ ## completed ## ]]

In adhering to this structure, your objective is: 
Score multiple API candidates for relevance to a query.

Scoring guide:
5 - Perfectly matches query intent, would definitely use this API
4 - Very relevant, highly plausible choice for this query
3 - Somewhat relevant, could work but not ideal
2 - Weakly relevant, missing key functionality
1 - Not relevant, wrong purpose entirely
\end{verbatim}
\end{tiny}

\subsubsection{Parallel API Relevance Scorer}

\paragraph{Objective:} Score multiple API candidates for relevance to a query in a parallel execution context.

\paragraph{System Prompt:}
\begin{tiny}
\begin{verbatim}
Your input fields are:
1. `query` (str): User's natural language query
2. `apis` (str): JSON array of API candidates with name, description, 
   parameters
3. `target_apis` (str): JSON array of target API names that should be 
   called in parallel. Score candidates considering they may be used 
   alongside these target APIs.

Your output fields are:
1. `reasoning` (str): 
2. `scores` (str): JSON array of scores. For each API provide: 
   {"api_name": str, "score": int (1-5), "reasoning": str}. Maintain same 
   order as input APIs. Consider parallel execution context in scoring.

All interactions will be structured in the following way, with the 
appropriate values filled in.

[[ ## query ## ]]
{query}

[[ ## apis ## ]]
{apis}

[[ ## target_apis ## ]]
{target_apis}

[[ ## reasoning ## ]]
{reasoning}

[[ ## scores ## ]]
{scores}

[[ ## completed ## ]]

In adhering to this structure, your objective is: 
Score multiple API candidates for relevance to a query in a parallel 
execution context.

This scorer evaluates candidates when multiple target APIs should be 
called together in parallel. Score candidates considering that they will 
be used alongside other APIs in a single query.

Scoring guide:
5 - Perfectly matches query intent, would definitely use this API in 
    parallel with target APIs
4 - Very relevant, highly plausible choice for this parallel execution
3 - Somewhat relevant, could work but not ideal for parallel execution
2 - Weakly relevant, missing key functionality for parallel context
1 - Not relevant, wrong purpose entirely or incompatible with parallel 
    execution
\end{verbatim}
\end{tiny}

\subsubsection{Construct Sequential Invocation}

\paragraph{Objective:} Construct the next API invocation in a sequential chain.

\paragraph{System Prompt:}
\begin{tiny}
\begin{verbatim}
Your input fields are:
1. `query` (str): User query
2. `available_apis` (str): JSON array of all available API schemas that 
   can be used to construct the next invocation.
3. `invocations_up_to_this_point` (str): JSON array of previous API 
   invocations in the chain, in order. Each invocation is a string in the 
   format: 'func_name(param_name1=param_value1, ...)'. Empty array for the 
   first API call.
4. `return_values_up_to_this_point` (str): JSON array of return values 
   from previous API calls in the chain, in order. Each return value is a 
   dictionary representing what a previous API returned. Empty array for 
   the first API call.

Your output fields are:
1. `reasoning` (str): Step-by-step reasoning for constructing the next 
   invocation. Explain: (1) Why this API should be called next, (2) How it 
   uses return values from earlier APIs (if any), (3) What parameter 
   values are needed and how they are inferred.
2. `next_api` (str): The next API invocation in the format: 
   'func_name(param_name1=param_value1, param_name2=param_value2, ...)'. 
   The function call should include all required parameters with 
   appropriate values inferred from the query and/or previous API return 
   values. CRITICAL: Use the exact function name as provided in the 
   available_apis, including any spaces, special characters, or formatting 
   - do not modify or 'fix' the function name. Return empty string '' if 
   no valid next invocation can be constructed. Return ONLY the invocation 
   string, no markdown, no code blocks, no explanation.

All interactions will be structured in the following way, with the 
appropriate values filled in.

[[ ## query ## ]]
{query}

[[ ## available_apis ## ]]
{available_apis}

[[ ## invocations_up_to_this_point ## ]]
{invocations_up_to_this_point}

[[ ## return_values_up_to_this_point ## ]]
{return_values_up_to_this_point}

[[ ## reasoning ## ]]
{reasoning}

[[ ## next_api ## ]]
{next_api}

[[ ## completed ## ]]

In adhering to this structure, your objective is: 
Construct the next API invocation in a sequential chain.

A sequential invocation means API functions are called IN ORDER, where 
later APIs can use return values from earlier APIs as input parameters. 
The first API's parameters come from the query, and subsequent APIs may 
have parameters inferred from previous API outputs.

Given the query, available APIs, and the invocations/return values up to 
this point, construct the NEXT single API invocation in the chain.
\end{verbatim}
\end{tiny}

\subsubsection{Validate Sequential Invocation}

\paragraph{Objective:} Validate if a sequential API invocation chain is correct for a given query.

\paragraph{System Prompt:}
\begin{tiny}
\begin{verbatim}
Your input fields are:
1. `query` (str): User query
2. `invocation_apis` (str): A list of full API invocations with parameters 
   in the format: "[func_name1(param_name1=param_value1, ...), 
   func_name2(param_name=param_value, ...)]". This represents the 
   sequential invocation chain to validate.
3. `api_schemas` (str): JSON array of API schemas for all APIs in the 
   invocation chain
4. `return_values_list` (str): JSON array of return value dictionaries, 
   one per API (except last), representing what each API returns in the 
   chain

Your output fields are:
1. `reasoning` (str): Step-by-step analysis: (1) Check if each API is 
   relevant to the query, (2) Verify that the first API's parameters can 
   be inferred from the query, (3) Verify that subsequent APIs' parameters 
   can be inferred from the query and/or return values of earlier APIs, 
   (4) Assess if the sequential execution chain is logically appropriate 
   (later APIs use earlier API outputs), (5) Determine if this is a 
   correct and complete solution for the query.
2. `is_valid` (str): YES if the invocation chain is correct (all APIs are 
   relevant, parameters are valid, and they form a logical sequential 
   chain). NO otherwise. Return YES or NO.

All interactions will be structured in the following way, with the 
appropriate values filled in.

[[ ## query ## ]]
{query}

[[ ## invocation_apis ## ]]
{invocation_apis}

[[ ## api_schemas ## ]]
{api_schemas}

[[ ## return_values_list ## ]]
{return_values_list}

[[ ## reasoning ## ]]
{reasoning}

[[ ## is_valid ## ]]
{is_valid}

[[ ## completed ## ]]

In adhering to this structure, your objective is: 
Validate if a sequential API invocation chain is correct for a given query.

A sequential invocation means APIs are called IN ORDER, where later APIs 
can use return values from earlier APIs as input parameters. The first 
API's parameters come from the query, and subsequent APIs may have 
parameters inferred from previous API outputs.

Validate that: (1) All APIs are relevant to the query, (2) The chain order 
is logical, (3) Parameter values can be inferred from the query and/or 
previous API return values, (4) The sequential execution chain makes sense 
(later APIs logically use earlier API outputs).
\end{verbatim}
\end{tiny}

\subsubsection{Construct Parallel Invocation}

\paragraph{Objective:} Construct a valid parallel API invocation from available APIs.

\paragraph{System Prompt:}
\begin{tiny}
\begin{verbatim}
Your input fields are:
1. `query` (str): User query
2. `available_apis` (str): JSON array of all available API schemas that 
   can be used to construct the invocation. Select a subset of these APIs 
   that together form a valid parallel invocation for the query.

Your output fields are:
1. `reasoning` (str): Step-by-step reasoning for constructing the 
   invocation. Explain: (1) Why these APIs should be called together in 
   parallel, (2) How they work together to answer the query, (3) What 
   parameter values would be needed for each API.
2. `invocation_apis` (str): A list of full API invocations with parameters 
   in the format: "[func_name1(param_name1=param_value1, 
   param_name2=param_value2, ...), func_name2(param_name=param_value, 
   ...)]". Each function call should include all required parameters with 
   appropriate values inferred from the query. CRITICAL: Use the exact 
   function names as provided in the available_apis, including any spaces, 
   special characters, or formatting - do not modify or 'fix' the function 
   names. Return empty array '[]' if no valid invocation can be 
   constructed. Return ONLY the invocation list, no markdown, no code 
   blocks, no explanation.

All interactions will be structured in the following way, with the 
appropriate values filled in.

[[ ## query ## ]]
{query}

[[ ## available_apis ## ]]
{available_apis}

[[ ## reasoning ## ]]
{reasoning}

[[ ## invocation_apis ## ]]
{invocation_apis}

[[ ## completed ## ]]

In adhering to this structure, your objective is: 
Construct a valid parallel API invocation from available APIs.

A parallel invocation means multiple API functions are called 
simultaneously with the same query. All APIs in a parallel invocation 
execute at the same time, sharing the same context and parameters. The 
invocation should be logically coherent - all APIs should work together to 
answer the query.

Construct a valid parallel invocation using any combination of the 
available APIs.
\end{verbatim}
\end{tiny}

\subsubsection{Validate Parallel Invocation}

\paragraph{Objective:} Validate if a parallel API invocation is correct for a given query.

\paragraph{System Prompt:}
\begin{tiny}
\begin{verbatim}
Your input fields are:
1. `query` (str): User query
2. `invocation_apis` (str): A list of full API invocations with parameters 
   in the format: "[func_name1(param_name1=param_value1, ...), 
   func_name2(param_name=param_value, ...)]". This represents the parallel 
   invocation to validate.
3. `api_schemas` (str): JSON array of API schemas for all APIs in the 
   invocation

Your output fields are:
1. `reasoning` (str): Step-by-step analysis: (1) Check if each API is 
   relevant to the query, (2) Verify that parameter values can be inferred 
   from the query, (3) Assess if the APIs can work together in parallel 
   (no sequential dependencies), (4) Determine if this is a correct and 
   complete solution for the query.
2. `is_valid` (str): YES if the invocation is correct (all APIs are 
   relevant, parameters are valid, and they work together coherently in 
   parallel). NO otherwise. Return YES or NO.

All interactions will be structured in the following way, with the 
appropriate values filled in.

[[ ## query ## ]]
{query}

[[ ## invocation_apis ## ]]
{invocation_apis}

[[ ## api_schemas ## ]]
{api_schemas}

[[ ## reasoning ## ]]
{reasoning}

[[ ## is_valid ## ]]
{is_valid}

[[ ## completed ## ]]

In adhering to this structure, your objective is: 
Validate if a parallel API invocation is correct for a given query.

A parallel invocation means multiple API functions are called 
simultaneously with the same query. All APIs execute at the same time and 
should work together coherently to answer the query. Validate that: (1) 
All APIs are relevant to the query, (2) The APIs can work together 
logically, (3) The parameter values are appropriate and can be inferred 
from the query, (4) The invocation makes sense as a parallel execution 
(not sequential dependencies).
\end{verbatim}
\end{tiny}

\subsubsection{Tool Call Equivalence}
\paragraph{Objective:} Evaluate if two tool calls are semantically equivalent.
\paragraph{System Prompt:}
\begin{tiny}
\begin{verbatim}
Your input fields are:
1. `user_query` (str): The user's original query/request
2. `tool_schema` (str): Complete tool schema including name, description, 
   and parameters
3. `ground_truth_call` (str): The ground truth tool call (JSON format)
4. `predicted_call` (str): The predicted tool call to evaluate (JSON 
   format)
Your output fields are:
1. `reasoning` (str): Step-by-step analysis of whether these calls are 
   equivalent. Consider: (1) same tool name, (2) semantically equivalent 
   parameter values (e.g., 'NYC' = 'New York'), (3) would produce same 
   result for the user
2. `equivalent` (str): YES if the calls are semantically equivalent, NO if 
   they differ meaningfully
All interactions will be structured in the following way, with the 
appropriate values filled in.
[[ ## user_query ## ]]
{user_query}
[[ ## tool_schema ## ]]
{tool_schema}
[[ ## ground_truth_call ## ]]
{ground_truth_call}
[[ ## predicted_call ## ]]
{predicted_call}
[[ ## reasoning ## ]]
{reasoning}
[[ ## equivalent ## ]]
{equivalent}
[[ ## completed ## ]]
In adhering to this structure, your objective is: 
Evaluate if two tool calls are semantically equivalent.
\end{verbatim}
\end{tiny}
\section{Scientific Artifacts and Licenses}\label{sec:scientific_artifacts}

This work uses external scientific artifacts and creates an in-house synthetic function-calling dataset (${\sim}$1.8K examples; \S\ref{sec:Analysis}).
We fine-tune \textbf{Llama-3.1-8B-Instruct}~\citep{grattafiori2024llama3herdmodels} under the \textbf{Llama~3.1 Community License} and implement training with Hugging Face \texttt{transformers}, \texttt{datasets}, \texttt{accelerate}, \texttt{peft}, and \texttt{trl} (Apache~2.0; Appendix~\ref{sec:fine_tuning_details}).
For preprocessing embeddings we use \textbf{all-MiniLM-L6-v2} (Apache~2.0).
For synthetic-data generation, quality control, augmentation baselines, distractor filtering, and LLM-as-a-judge evaluation (\S\ref{sec:Our Methods}--\S\ref{sec:LLM Fine-tuning}), we access \textbf{Amazon Nova} models via \textbf{AWS Bedrock} under Amazon's API terms (commercial, metered access; not open-weight).
We evaluate on publicly released \textbf{ToolAce} and \textbf{APIGen} corpora~\citep{liu2409toolace,liu2024apigen} and the \textbf{BFCL} benchmark~\citep{patilberkeley}, following each project's stated license and intended research use.
Our use of these artifacts is limited to academic study of synthetic data diversity and function-calling evaluation, consistent with their research-oriented intended uses.

\end{document}